\documentclass[superscriptaddress, preprint, amsmath, amssymb, aps, floatfix]{revtex4-1}

\usepackage{graphicx, makecell, multirow, xcolor}
\usepackage{bm, xurl, setspace, booktabs}

\renewcommand{\thetable}{\arabic{table}}

\begin{document}

\title{A semantic embedding space based on large language models for modelling human beliefs}

\textcolor{blue}{Note: For citation purposes, please refer to the version published in \textit{Nature Human Behaviour} (\url{https://doi.org/10.1038/s41562-025-02228-z}).}

\author{Byunghwee Lee}
\affiliation{Center for Complex Networks and Systems Research, Luddy School of Informatics, Computing, and Engineering, Indiana University, Bloomington, Indiana, USA, 47408}

\author{Rachith Aiyappa}
\affiliation{Center for Complex Networks and Systems Research, Luddy School of Informatics, Computing, and Engineering, Indiana University, Bloomington, Indiana, USA, 47408}

\author{Yong-Yeol Ahn}
\affiliation{Center for Complex Networks and Systems Research, Luddy School of Informatics, Computing, and Engineering, Indiana University, Bloomington, Indiana, USA, 47408}

\author{Haewoon Kwak}
\email{hwkwak@iu.edu}
\affiliation{Center for Complex Networks and Systems Research, Luddy School of Informatics, Computing, and Engineering, Indiana University, Bloomington, Indiana, USA, 47408}

\author{Jisun An}
\email{jisunan@iu.edu}
\affiliation{Center for Complex Networks and Systems Research, Luddy School of Informatics, Computing, and Engineering, Indiana University, Bloomington, Indiana, USA, 47408}

\begin{abstract}
Beliefs form the foundation of human cognition and decision-making, guiding our actions and social connections. A model encapsulating beliefs and their interrelationships is crucial for understanding their influence on our actions. However, research on belief interplay has often been limited to beliefs related to specific issues and relied heavily on surveys. We propose a method to study the nuanced interplay between thousands of beliefs by leveraging an online user debate data and mapping beliefs onto a neural embedding space constructed using a fine-tuned large language model (LLM). This belief space captures the interconnectedness and polarization of diverse beliefs across social issues. Our findings show that positions within this belief space predict new beliefs of individuals and estimate cognitive dissonance based on the distance between existing and new beliefs. This study demonstrates how LLMs, combined with collective online records of human beliefs, can offer insights into the fundamental principles that govern human belief formation.  
\end{abstract}

\maketitle
\section*{Introduction \label{sec:Intro}}

Beliefs are foundational for human cognition and decision making. The term `belief' refers to a conviction that something is true or exists~\cite{dictionary2008cambridge}, or a confidence in the rightness of something or someone~\cite{dictionary2000oxford}. Beliefs guide how individuals derive meaning, shape behavior, filter information, and form social connections that define their communities~\cite{dellaposta2015liberals, goldberg2018beyond, gonzalez2023asymmetric, cinelli2021echo}.

Research across disciplines has advanced our quantitative understanding of human beliefs. The growth of digital behavioral data and new analytical tools has enabled novel approaches to studying belief systems, ranging from mapping individual belief structures to analyzing how beliefs spread through societies. Notable approaches include modeling of belief dynamics using social network diffusion models~\cite{castellano2009statistical, degroot1974reaching, sen2014sociophysics, deffuant2001mixing, friedkin1990social, watts2002simple} and frameworks that integrate both individual belief systems and social influence mechanism, along with their empirical applications~\cite{dellaposta2015liberals, goldberg2018beyond, macy2019opinion, galesic2021integrating, aiyappa2023weighted, dalege2016toward, dalege2022using, rodriguez2016collective, schweighofer2020weighted}. In parallel, researchers in natural language processing (NLP) and social media analytics have developed methods to detect and predict individuals' beliefs through their digital footprints and textual expressions~\cite{durmus2019exploring, longpre2019persuasion, agarwal2022graphnli, mohammad2016semeval, introne2023measuring, darwish2020unsupervised, rashed2021embeddings}.

Studies have revealed that human beliefs are interconnected, and understanding these relationships is crucial for comprehending how beliefs form, update, and propagate alongside associated behaviors. 
For instance, individuals sharing similar beliefs can influence each other's lifestyle choices, leading to clustered behaviors and preferences within social groups~\cite{dellaposta2015liberals}. This explains why seemingly unrelated beliefs (or preferences), such as being liberal and drinking lattes, can become associated.
The associative diffusion model~\cite{goldberg2018beyond} also demonstrates how relationships between beliefs shape cultural differentiation. This model suggests that cultural differences emerge primarily through the transmission of perceived compatibility between beliefs and behaviors, rather than through the direct transmission of the beliefs themselves. This process can occur even in communities lacking pre-existing social clusters.
Another experiment suggests that a small number of early movers can initiate belief-ideology associations that subsequently develop into strong partisan alignments, even for beliefs fundamentally unrelated to political ideology~\cite{macy2019opinion}. Recent theoretical frameworks have integrated individual belief systems with social network influences, modeling belief dynamics through the lens of dissonance theory and network imbalance~\cite{galesic2021integrating, dalege2016toward, dalege2022using, aiyappa2023weighted, rodriguez2016collective, schweighofer2020weighted}. These models illuminate how discrepancies and imbalances between beliefs shape the ultimate distribution of social beliefs. Collectively, these studies highlight that understanding the interconnected nature of beliefs is crucial for explaining societal fragmentation and polarization.

Yet, the relational landscape of human beliefs remains incompletely mapped, and our understanding of how belief interactions form is still limited. Despite the advances in belief quantification and modeling, significant challenges still persist. A primary challenge in modeling belief systems lies in representing the nuanced \emph{relationships} between beliefs. Network-based approaches to modeling human attitudes typically rely on survey data on specific issues to explore belief interrelationships (e.g., through partial correlation between questionnaire responses~\cite{dellaposta2015liberals, galesic2021integrating, dalege2022using}). However, survey-based methods face inherent scalability limitations when considering the entire ``space'' of beliefs. Capturing relationships among vast numbers of important beliefs and incorporating new beliefs into existing systems---a process known as inductive reasoning---pose significant challenges. 

Here, we construct a robust and general \emph{representation space for beliefs} that enables both continuous and inductive reasoning about beliefs and their relationships. Drawing inspiration from vector space models that encode semantic and contextual relationships between words into geometric relations~\cite{mikolov2013efficient, le2014distributed, an2018semaxis}, our approach leverages large language models (LLMs), combined with revealed belief trajectories extracted from online debates. This methodology creates a continuous, high-dimensional representation space. Our framework uses an empirical dataset of multiple beliefs held by individuals from an online debate forum to fine-tune a pre-trained LLM---a model that initially trained on extensive text corpora to capture broad linguistic patterns. The resulting belief-LLM translates belief statements into embedding vectors, creating a space where spatial distances reflect both semantic relationships and socially perceived relevance between beliefs. 
It also enables representing individuals as vectors in belief space, allowing inference of their implicit beliefs and exploration of belief adoption processes.

In this study, we aim to propose a novel framework for constructing a robust belief embedding space using LLMs integrated with online user activity data. We investigate the emergent structural characteristics in this belief embedding space, focusing on clustering and polarization patterns around social issues. Through evaluation of the embedding space's effectiveness in inferring individuals' stances on new debate topics from their existing beliefs, we address a fundamental question: What mechanisms underlie human belief selection? We explore this by analyzing the factors that enable accurate belief prediction.

\begin{figure*}[t!]
    \centering
    \includegraphics[width=\textwidth]{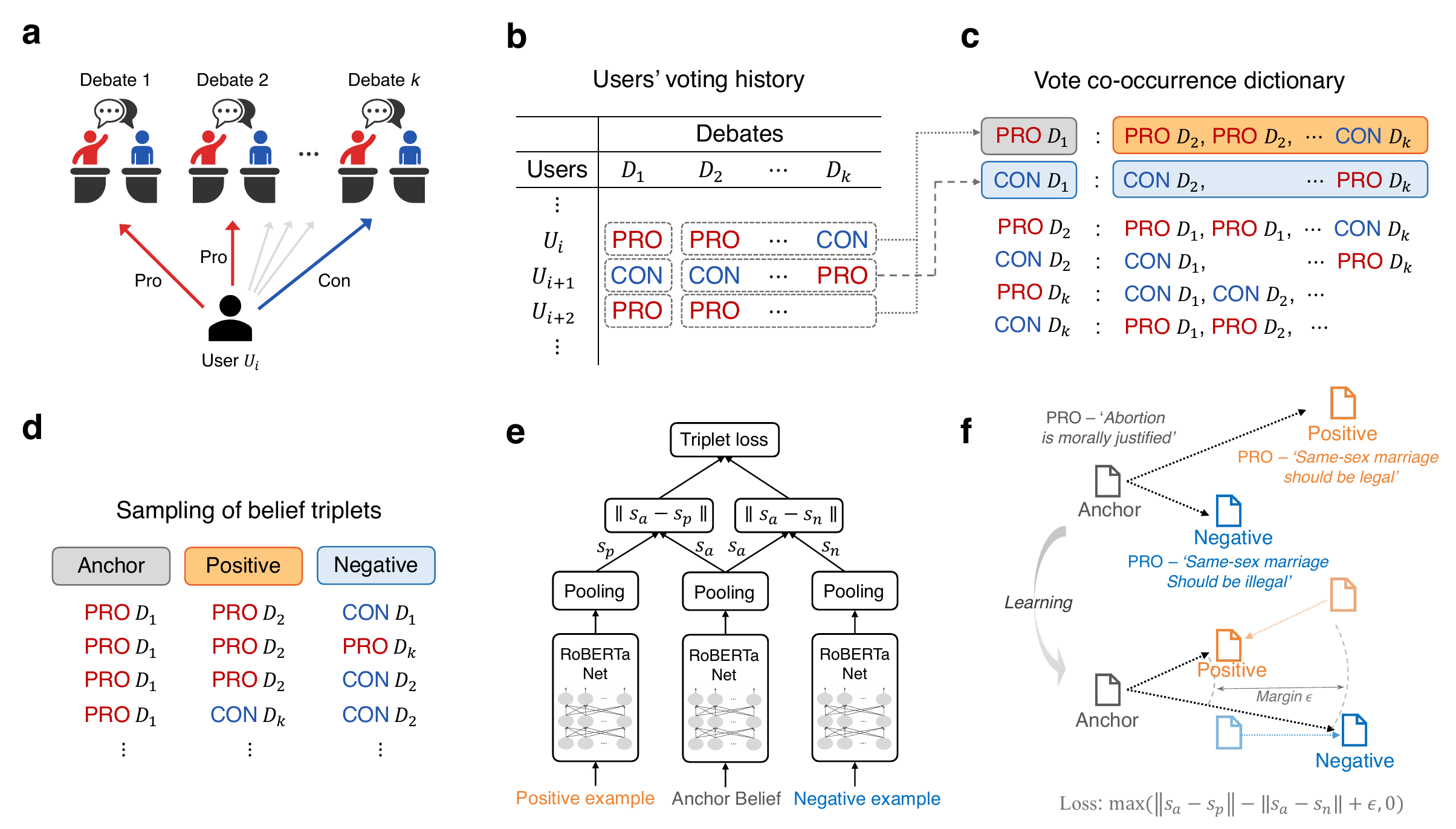}
	\caption{\textbf{Fine-tuning S-BERT with belief triplets.} (a) An illustration of a user's expressed positions in multiple debates. For each debate topic, users can vote for the PRO or CON side. (b) Voting histories of users represented as a matrix. (c) The vote co-occurrence dictionary captures how users voted on other beliefs, given their PRO/CON vote on a certain belief. (d) From the vote co-occurrence dictionary, belief triplets are sampled. Each belief triplet is composed of an anchor belief in addition to a positive and a negative belief in relation to the anchor. (e) A pre-trained S-BERT model is fine-tuned with the belief triplets in the form of triplet network. (f) An illustration of the learning process happening within S-BERT. To minimize the triplet loss, an anchor belief is drawn closer to beliefs with positive relationships.}
	\label{fig:Finetuning}
\end{figure*}

\section*{Results}\label{sec:Results}

\subsection*{Generating and validating the belief embedding space}
Our first goal is to construct a representation space that captures the interdependencies between diverse beliefs. We achieve this by fine-tuning pre-trained LLMs using contrastive learning~\cite{Schroff2015}. This approach enables models to learn a representation space by attracting similar (positive) belief pairs while repelling dissimilar (negative) ones, allowing us to distinguish commonly shared belief pairs from those that are in opposition.

We leverage user participation records from an online debate forum, Debate.org (DDO)~\cite{durmus2019exploring, durmus2019corpus, longpre2019persuasion}. This dataset consists of online debates and corresponding voting records of the users; users can express their position by directly participating as debaters or voting for the PRO, CON or TIE position in the debates. We consider both debaters and voters simply as voters since they support a particular position in the debate. After pre-processing, we obtained a dataset of 59,986 unique debate titles voted on 197,306 times by 40,280 unique users, retaining only PRO and CON votes, which was used for fine-tuning LLMs (see Methods). 

We operationalize each individual's belief as their expressed agreement or disagreement with a certain debate title. We transform voting records (PRO/CON) of users into belief statements by using predefined templates. For example, if a user voted PRO (CON) to a debate titled ``Abortion is morally justified,'' we create a belief statement for the user as ``I agree (disagree) with the following: Abortion is morally justified.'' (See SI Section~3, for template variability on belief embeddings).

To encode these belief statements, we employ a pre-trained Sentence-BERT (S-BERT) model with RoBERTa~\cite{reimers2019sentence, liu2019roberta}. Unlike the original BERT model~\cite{devlin2018bert}, which is focused on token-level tasks, S-BERT is designed for generating semantically meaningful sentence-level embeddings and allows for efficient fine-tuning using sentence-level pairs or triplets. We fine-tune S-BERT model with a triplet-based contrastive learning approach. As illustrated in Figs.~\ref{fig:Finetuning}a-d, we create belief triplets from user voting activities, treating them as positive belief pairs, which are contrasted against negative examples. Specifically, we first create a vote co-occurrence dictionary, where each voted belief serves as a key, and the values consist of other beliefs that were also voted on by users who voted the key beliefs, allowing for duplication. From this dictionary, belief triplets are sampled (See Methods). Notably, the more frequently two beliefs are shared by users, the more likely they are to be sampled as positive examples. Conversely, negative pairs are derived from beliefs that represent the opposing stance of the anchor belief or from beliefs that are frequently co-voted with the opposing belief.

These triplets are then utilized to fine-tune the LLMs using a triplet loss function as depicted in Figs.~\ref{fig:Finetuning}e and f. The resulting model offers a 768-dimensional latent \emph{belief space}, where an individual belief is mapped into a vector within the space, and the distance between two vectors capture their semantic and contextual similarity. We show that the distance between beliefs in the belief space is proportional to the likelihood that an individual has one belief given the other.

\begin{table*}
\centering
\small
\begin{tabular}{lcccc} 
&  & \multicolumn{2}{c}{Triplet evaluator} & GLUE-STSB \\ 
\hline 
Model type & Pre-trained model & Training set & Test set & Spearman corr. \\ 
\hline 
\makecell[l]{S-BERT\\(Fine-tuned)} & \makecell{roberta-base-nli\\-stsb-mean-tokens} & \makecell{0.946\\(0.001)} & \makecell{0.674\\(0.002)} & \makecell{0.718\\(0.005)}\\
\makecell[l]{S-BERT\\(Before fine-tuning)} & \makecell{roberta-base-nli\\-stsb-mean-tokens} & \makecell{0.397\\(0.001)} & \makecell{0.376\\(0.003)} & 0.877 \\ 
\makecell[l]{BERT\\(Fine-tuned)} & bert-base-uncased & \makecell{0.933\\(0.003)} & \makecell{0.669\\(0.004)} & \makecell{0.476\\(0.045)}\\ 
\makecell[l]{BERT\\(Before fine-tuning)} & bert-base-uncased & \makecell{0.376\\(0.001)} & \makecell{0.356\\(0.002)} & 0.615\\ 
\hline 
\end{tabular}
\caption{\label{tab:performance} Performance of various LLMs in the belief triplet evaluation task for both the training and test sets. Scores represent the average accuracy obtained from a 5-fold validation task. A higher accuracy indicates that the model more accurately distinguishes positive examples from negative ones for a given anchor belief. Numbers in parentheses denote standard deviations. The last column demonstrates performance in the GLUE-STSB task~\cite{wang2019glue}, where the goal is to estimate semantic textual similarity between two texts. Performance is assessed through the Spearman correlation between the human-annotated benchmark score (rated on a scale of 1 to 5) and the cosine similarity between the vector representations of the two texts, as generated by LLMs.}
\end{table*}

We use two different approaches, a triplet evaluator and a semantic similarity evaluation task, to evaluate the quality of the belief embeddings generated by the LLMs. 
We initially assessed the performance of various LLMs by employing a triplet evaluator for classifying belief pairs as either positive or negative relations. Table~\ref{tab:performance} compares triplet evaluation results from different models. The fine-tuned S-BERT model showed the highest performance with an average accuracy of about 0.95 for the train dataset and about 0.67 for the test sets (See Table~\ref{tab:performance}). 

Our model also shows good performance in capturing the general semantic meaning of various texts beyond the range of our training dataset, which is directly related to how accurate a vector representation of a new, unseen belief would be. Even after proceeding with the fine-tuning process, the S-BERT still kept a relatively high performance score on the semantic textual similarity benchmark on general language understanding evaluation datasets (GLUE-STSB)~\cite{wang2019glue} compared to other models. The Spearman rank correlation coefficient score of the S-BERT model is $r=0.718\pm0.005$, while the fine-tuned BERT model shows a relatively low correlation score ($r=0.476\pm0.045$) (Table~\ref{tab:performance}).

\subsection*{Belief landscape revealed by belief embeddings}
\subsubsection*{PCA results of the belief space}

\begin{figure*}[t!]
    \centering
	\includegraphics[width=\textwidth]{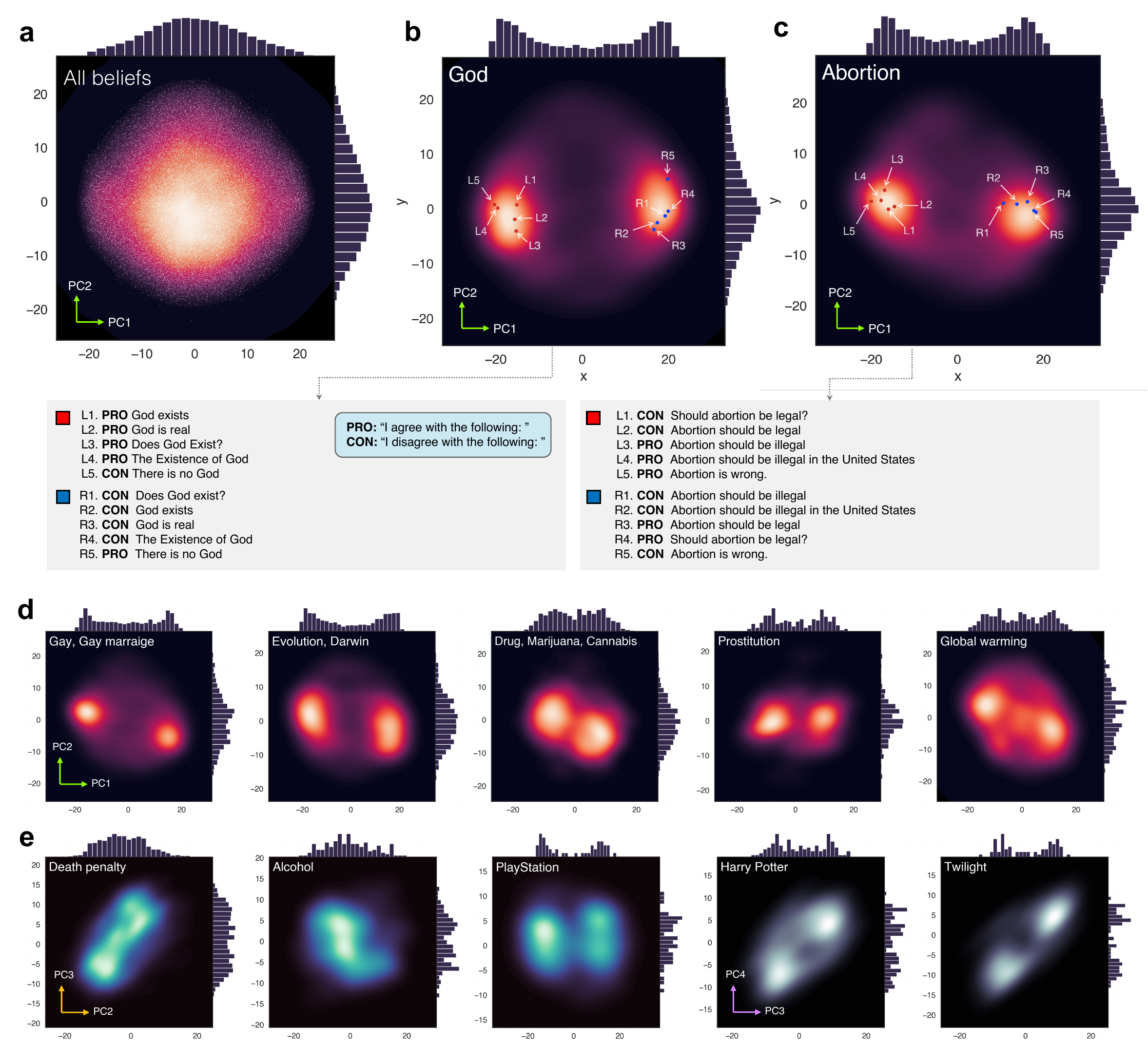}
	\caption{\textbf{Structure of belief space via PCA.} (a) The entire distribution of belief embeddings is represented in the first two principal components (PC1-PC2) space. The background heatmap reflects the density of beliefs, and the overlaid white dots indicate individual beliefs. (b) and (c) depict the distribution of beliefs related to the topics `God' and `Abortion,' respectively. Both belief distributions exhibit two highly clustered regions in the PC space, signifying the presence of two distinct groups of beliefs regarding these topics. Five example beliefs from each cluster, presented in the gray boxes, are highlighted as red and blue points. (d) Distributions of beliefs corresponding to five different topics, displaying two bimodal distributions in the first two principal components (PC1-PC2) space, similar to (a) and (b), are illustrated. (e) Additional belief distributions corresponding to five topics, revealing unique structures in the higher-order principal components, are displayed.}
    \label{fig:beliefspace}
\end{figure*}

To examine the structure of the belief space, we perform Principal Component Analysis (PCA) on the entire belief vectors generated from the fine-tuned S-BERT model. For analysis of the overall distribution of beliefs on various topics, we compiled twelve example sets of beliefs chosen from various fields that exhibit distinct patterns in the PC space, each consisting of a unique set of keywords relevant to their belief statements. For example, 5,000 distinct beliefs relate to the topic of `God' and 1,470 beliefs relate to the topic of `Gay marriage.'

Figure~\ref{fig:beliefspace} presents the density of beliefs along the first and second principal component axes (PC1 and PC2) across belief subgroups, each related to distinct topics. The entire belief set (Fig.~\ref{fig:beliefspace}a) exhibits a smooth, uni-modal, bell-shaped distribution along both the first and second principal component axes (PC1 and PC2). However, the density plots for beliefs related to specific topics, such as ‘God’ and ‘Abortion,’ reveal markedly different, polarized patterns (Figs.~\ref{fig:beliefspace}b and c), suggesting that beliefs regarding these topics are grouped into two clusters with contrasting opinions. These bimodal patterns of beliefs are also observed in belief spaces using other types of dimensionality reduction methods (see Fig.~S9 for the UMAP~\cite{mcinnes2018umap} results).

Belief embeddings also reveal which beliefs are more closely associated with each other. For instance, beliefs favoring the existence of God (god) or opposing abortion are predominantly found on the negative side of the PC1 axis. The positive side of this axis is associated with disbelief in God and support for abortion rights. Additionally, beliefs related to topics such as `Gay and gay marriage,' `Evolution and Darwin,' and `Drug, Marijuana, and Cannabis' also exhibit two dense clusters in the PC1 and PC2 space (Fig.~\ref{fig:beliefspace}d), which aligns with the broader trend of political polarization observed across diverse social issues~\cite{brenan2019birth, gentzkow2016measuring, campbell2018polarized}. 

Examining the other PC axes reveals another dimension of belief separation. For instance, beliefs about `PlayStation' exhibit bimodal distributions along the PC2 axis, while beliefs related to `Alcohol' display a weakly bimodal distribution along the PC3 axis (Fig.~\ref{fig:beliefspace}d). Similarly, beliefs about ‘Harry Potter’ and `Twilight' cluster into two distinct groups on the PC3-PC4 plane, whereas they do not display a noticeable pattern on the PC1-PC2 plane. This indicates that the contextual relationships among beliefs concerning these topics are encoded in the PC3 and PC4 axes. However, not all topics show such polarized distributions; for example, beliefs related to `Society,' `Education,' and `USA' tend to spread around in the PC space. This is likely because these topics encompass a wide array of subtopics, leading to greater variation in belief representation.

Overall, our results demonstrate that the distributions of beliefs related to various topics show unique patterns in the belief space, often forming polarized clusters along specific axes. Furthermore, the arrangement of belief positions of various polarizing issues in the PC space is generally aligned with the partisan polarization observed in public surveys. For instance, according to Gallup's beliefs poll in 2019~\cite{brenan2019birth}, American liberals were more likely to consider `Abortion' (73\%), and `Gay/lesbian relations' (81\%) morally acceptable. In contrast, only 23\% and 45\% of conservatives believed these issues to be morally acceptable, respectively, demonstrating the interconnected nature of these beliefs.

\subsubsection*{Embedding individuals in belief space reveals group polarization}

\begin{figure*}[t!]
    \centering
	\includegraphics[width=\textwidth]{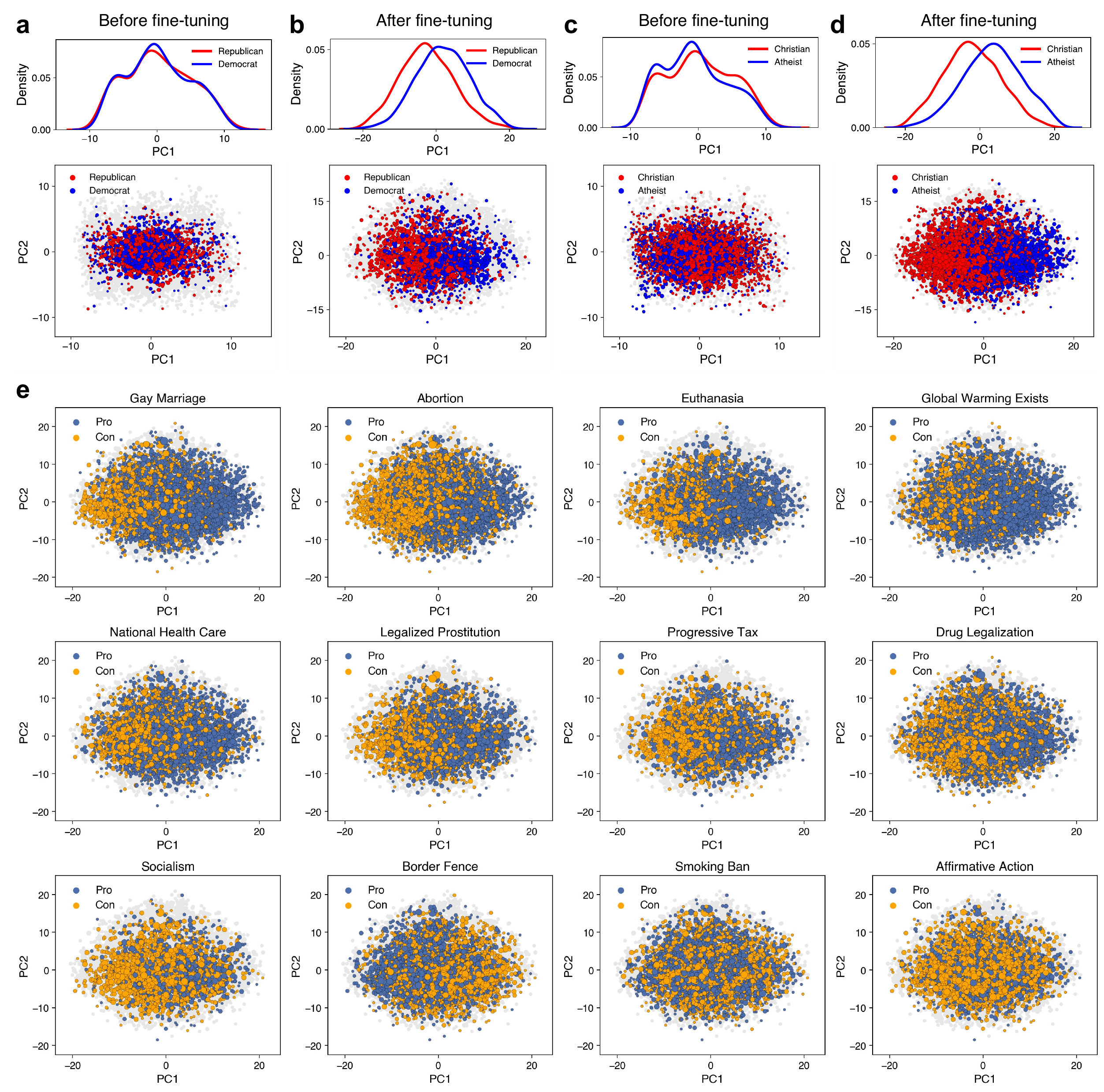}
	\caption{\textbf{User embeddings in belief space.} (a)-(d) depict the positions of every user in the belief space along the first two principal component (PC) axes, colored based on their self-reported supporting political parties (Republican vs. Democrat) and religious ideologies (Christian vs. Atheist). 
    While (a) and (c) illustrate user embeddings obtained by averaging their belief embeddings from the base S-BERT model without fine-tuning, (b) and (d) present embeddings from the fine-tuned S-BERT model. Compared to the non-fine-tuned model, the fine-tuned model exhibits a clearer separation of users from different groups in the belief space. (e) displays users grouped by their self-reported stances on various controversial social issues.}
	\label{fig:userembedding}
\end{figure*}

The presence of topic-specific bimodal belief distributions leads to a question: Do individuals with distinct ideologies also exhibit meaningful clusters within the belief space? To investigate this, we represent each user by their average belief vector, defined as $\vec{u}=\sum_{i=1}^{N_u}{\vec{b_i^u}}/{N_u}$, where $\vec{b}_i^u$ denotes $i$-th belief vector of the user $u$, and $N_u$ is the total number of beliefs associated with user $u$. This allows us to measure how closely users are positioned in the belief space. 

We then visually map users in the belief space according to their self-reported survey responses to assess whether the resulting user representations properly locate them. In DDO, users can self-report their positions on major social issues via pre-survey participation, independently from their debate participation. This includes specifying their supporting political parties, religious beliefs, and positions on 48 key social issues, referred to as \textit{big issues}. The big issues encompass a range of controversial social issues such as `Abortion,' `Drug legalization,' `Gun control', and others. 

Figs.~\ref{fig:userembedding}a-d illustrate the positions of users in the belief space along the first two PC axes. These positions are obtained by averaging their belief embeddings from S-BERT models before and after fine-tuning. The color coding in these figures reflects the users' self-reported political ideologies (i.e., Democrat vs. Republican) and religious ideologies (i.e., Christian vs. Atheist). Remarkably, users represented by their average beliefs derived solely from voting records form two distinct clusters corresponding to their political and religious ideologies.

Figs.~\ref{fig:userembedding}b and d, which depict results from the fine-tuned S-BERT model, show a significant separation of user groups along the PC1 axis, suggesting that this axis primarily captures the alignment of users' beliefs with political and religious ideologies. By contrast, the base S-BERT model without fine-tuning does not exhibit such clear ideological group separations (Figs.~\ref{fig:userembedding}a and c, and Fig.~S10). This demonstrates that the fine-tuned S-BERT model more effectively captures the contextual relationships between beliefs, positioning related beliefs closer within the
space. 

The fine-tuned model also effectively reveals the alignment of partisan identity with beliefs on other disparate social issues. As shown in Fig.~\ref{fig:userembedding}e, distinct user groups on various issues, such as `Gay marriage,' `Abortion,' `Euthanasia,' and `Global warming exists,' exhibit separation along the same PC1 axis, which represents partisan polarization. However, user groups are less clearly separated on other issues, such as `Smoking ban' and `Affirmative action.' It may be because these issues do not align neatly with prominent political or social dichotomies, such as the liberal-conservative spectrum, and because complexities beyond such dichotomies may not be fully captured by the PC1 axis. For example, individuals from both liberal or conservative backgrounds might either support or oppose smoking ban. Similarly, perspectives on affirmative action may be influenced more by ethnicity than by political affiliation.

We further examine how the Euclidean distance between PRO and CON user group centroids correlates with self-reported partisan polarization across the 48 \textit{big issues} (See SI Section 4E and Fig.S11). The analysis reveals a significant positive correlation ($r=0.627$, $P<0.001$), indicating that distances in the belief space accurately reflect the intensity of ideological polarization across these issues.

We note that our results primarily reflect user behaviors within the US-centric DDO dataset, and the observed clustering and polarization patterns may differ in other social and cultural contexts. Nevertheless, our findings demonstrate that the belief embedding framework effectively captures meaningful contextual relationships across a diverse societal beliefs.

\subsection*{Belief embedding predicts user beliefs on unseen debates}

Our results show that like-minded individuals with similar beliefs on specific social issues tend to cluster together in the belief space. This observation raises two further questions: Can we utilize the user embeddings to predict an individual's belief on unseen debates? Can we uncover any underlying mechanisms of human decision-making by analyzing large-scale data on how users select their beliefs? According to the literature on dissonance theory of human attitudes, people tend to experience cognitive dissonance when they are exposed to information that is not in alignment with their existing beliefs~\cite{festinger57cognitive, galesic2021integrating}. Moreover, the feeling of personal discomfort created by the conflict between the new information and one's own beliefs can possibly lead to selective exposure to belief-confirming information~\cite{frimer2017liberals}. Similarly, in our study, we consider a user's prior beliefs about various debate issues to constitute their existing belief system. Choosing a new belief towards a new debate is akin to adding a new belief to a user's existing belief system.

To quantitatively model the belief selection process, we design a binary belief classification task to predict a user's voting position (PRO or CON) in new debates. For this, we split the entire set of debates into an 8:2 ratio and evaluate the model's performance using 5-fold cross-validation, considering users who appear at least once in both the training and test sets (Table~S4). We leverage user embeddings, learned from the training set, to predict a user's positions on previously unseen debates from the test set. We compare our results with multiple baselines and existing models.

\begin{figure*}[ht!]
    \centering
	\includegraphics[width=0.95\textwidth]{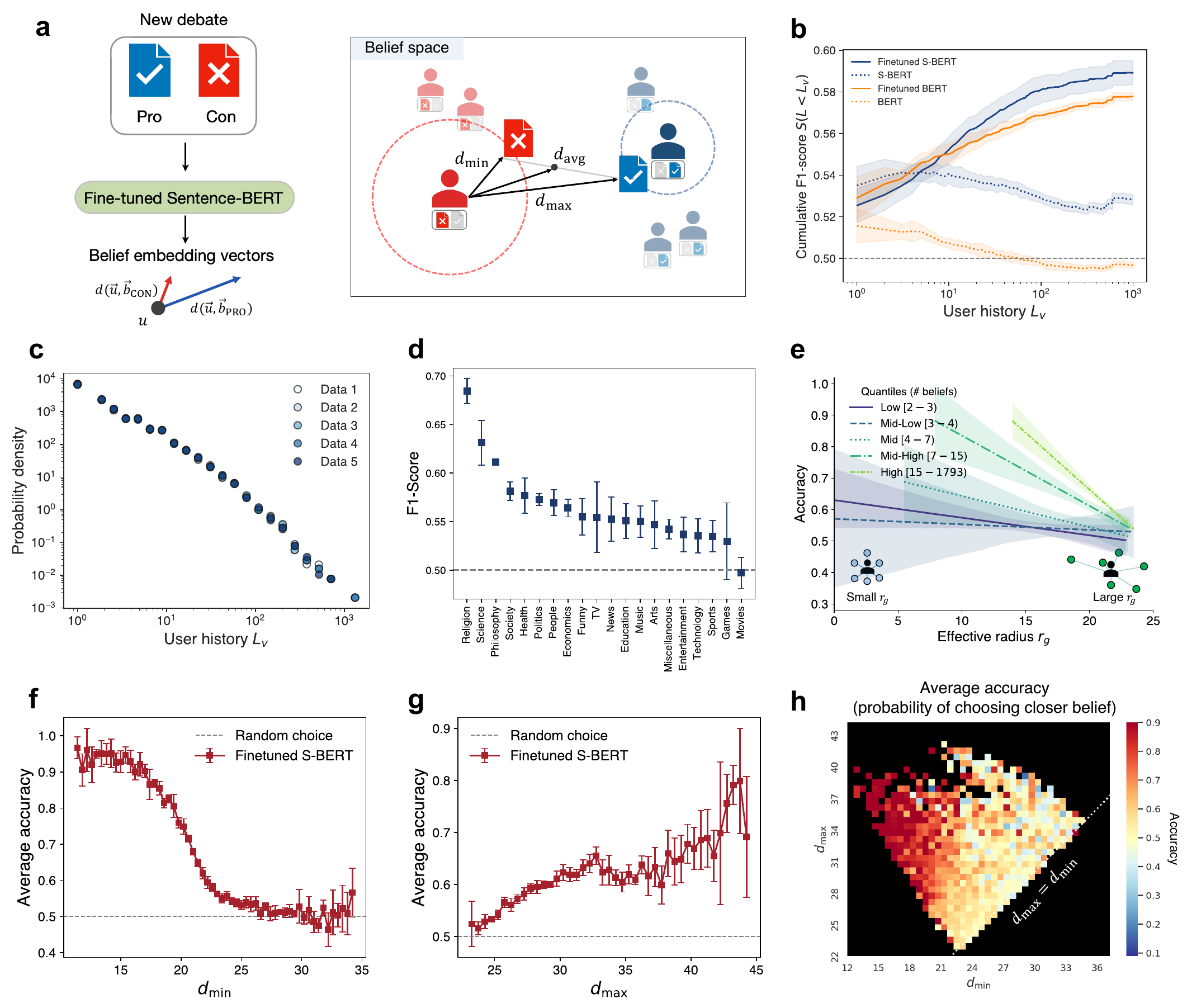}
	\caption{\textbf{Predicting users' beliefs about new debates.} (a) For the two competing debate positions, PRO and CON, in an unseen debate, we define their belief vectors as \( b_{\text{PRO}} \) and \( b_{\text{CON}} \), respectively. Given a user's prior belief \( u \), we compute the distances \( d(u, b_{\text{PRO}}) \) and \( d(u, b_{\text{CON}}) \). A model predicts the user's choice as the belief vector that minimizes the distance, given by: $\hat{b} = \arg\min_{b \in \{b_{\text{PRO}}, b_{\text{CON}}\}} d(u, b)$. The minimum and maximum distances are denoted as $d_\text{min}$ and $d_\text{max}$, respectively, with their average labeled as $d_\text{avg}$. (b) Relationship between the F1-score $S(L < L_v)$ and the length of user history $L_v$, where $S(L < L_v)$ represents the F1-score for users with voting records shorter than $L_v$. Data are presented as mean $\pm$ SD across 5-fold cross-validation. (c) Distribution of user history $L_v$ (number of votes) in a 5-fold dataset. (d) Variation in belief prediction accuracy, quantified using the F1-score, across diverse debate categories. The graph depicts top 20 most frequent categories, each appearing over 100 times in the prediction task. Data are presented as mean $\pm$ SD. (e) Accuracy trend as a function of users' effective radius. The lines represent user groups divided into five quantiles based on the number of prior beliefs, with each group containing a similar number of users. The shaded areas indicate the 95\% confidence interval around the fitted mean regression line. (f) and (g) Average accuracy trends across $d_\text{min}$ and $d_\text{max}$ in the belief prediction task. Error bars represent mean $\pm$ SD. (h) The heatmap illustrates the average prediction accuracy across various ranges of $d_\text{min}$ and $d_\text{max}$.}
\label{fig:downstream}
\end{figure*}

\begin{table*}
\centering
\small
\begin{tabular}{lcc} 
Model type & Accuracy & Macro F1-Score  \\ \hline 
\makecell[l]{S-BERT\\(Fine-tuned)} & \makecell{0.590\\(0.006)} & \makecell{0.590\\(0.005)} \\ 
\makecell[l]{S-BERT\\(Before fine-tuning)} & \makecell{0.565\\(0.002)} & \makecell{0.527\\(0.002)} \\ 
\makecell[l]{BERT\\(Fine-tuned)} & \makecell{0.579\\(0.002)} & \makecell{0.578\\(0.002)} \\ 
\makecell[l]{BERT\\(Before fine-tuning)} & \makecell{0.541\\(0.001)} &\makecell{0.496\\(0.001)}\\ 
\makecell[l]{Baseline 1\\(Random choice)} & \makecell{0.499\\(0.002)} & \makecell{0.499\\(0.002)} \\
\makecell[l]{Baseline 2\\(Majority selection)} & \makecell{0.532\\(0.001)} & \makecell{0.347\\(0.001)} \\
\makecell[l]{Llama2-13b-chat} & \makecell{0.537\\(0.002)} & \makecell{0.371\\(0.002)} \\
\hline
\end{tabular}
\caption{\label{tab:downstream} Performance of various LLMs in a downstream task on predicting users' beliefs for unseen debates. Numbers in parentheses denote standard deviations obtained from 5-fold validation results.}
\end{table*}

Our model employs a straightforward approach; it predicts a user's choice based on the Euclidean distance between the user's position and two opposing belief vectors from a new debate. 
For each debate in the test set, we construct two opposing belief statements from the debate title and generate their corresponding belief vectors, $\vec{b}_{\text{PRO}}$ and $\vec{b}_{\text{CON}}$, using the fine-tuned S-BERT model (Fig.~\ref{fig:downstream}a).
Given a user embedding $\vec{u}$, representing the average of their prior beliefs, we compute the distances $d(\vec{u}, \vec{b}_{\text{PRO}})$ and $d(\vec{u}, \vec{b}_{\text{CON}})$, and predict the user's choice as the belief vector minimizing the distance: $\vec{b'} = \arg\min_{\vec{b} \in \{\vec{b}_{\text{PRO}}, \vec{b}_{\text{CON}}\}} d(\vec{u}, \vec{b})$.

Comparative evaluation with other LLMs reveals that the fine-tuned S-BERT model exhibits the highest performance, with an F1-score of 0.59 ($\sigma=0.01$) and an accuracy of 0.59 ($\sigma=0.01$). We use the macro F1-score to ensure balanced performance evaluation across all classes. This performance is notably superior compared to other models, including the base S-BERT model, the base and fine-tuned BERT models, and other baseline models, as detailed in Table~\ref{tab:downstream}. 
 
We also benchmark our model against two baseline models: the random choice model (baseline 1) and the majority selection model (baseline 2). The random choice model randomly predicts a user's belief between two given belief options. By its random nature, it is expected to achieve an F1-score and accuracy of 0.5. The majority selection model accounts for the asymmetric ratio of PRO and CON beliefs in the training set by predicting that all users will consistently choose the more prevalent side. The majority model registers higher accuracy (0.53) but a lower F1-score (0.35). The fine-tuned S-BERT model outperforms both of these baselines. Additionally, we evaluate Llama2 (Llama2-13b-chat)~\cite{touvron2023llama} in a few-shot setup (see Methods), achieving an accuracy of 0.54 and an F1-score of 0.37, slightly outperforming the majority baseline (Table~\ref{tab:downstream}).

Although the fine-tuned S-BERT model shows superior performance in belief prediction compared to other models, its overall F1-score is not particularly high (0.59). To understand the intricacies affecting the performance of the belief prediction, we explore various factors and identified four critical ones: the length of individuals' voting history, debate category, individual's effective radius, and the relative distance between a user and two beliefs being considered.

First, our findings indicate that the prediction accuracy largely depends on the extent of a user's voting history in the training set. Figures~\ref{fig:downstream}b and S13 show that as we progressively include users with longer voting history, the average F1-score of users almost monotonically increased. This result shows that the users' beliefs are more accurately predicted when we know more about their prior beliefs. However, a fundamental obstacle to accurate prediction is that the degree of user participation activities in DDO follows a highly skewed distribution (Fig.~\ref{fig:downstream}c), which is commonly found in many online human activities~\cite{muchnik2013origins}.

Second, the diverse nature of debate topics also poses a significant challenge. While debates related to politics and religion are common, many debates in the DDO dataset focus on issues closely tied to pop culture and recreational topics, such as `Batman could beat Spiderman in a fight' or `Soccer as the best sport.' Beliefs on such topics are often highly distinct from those on other issues, complicating predictions unless the user has previously engaged in similar topics. We utilize the topic categories provided in the DDO dataset and measured prediction performance across different debate categories. As highlighted in Fig.~\ref{fig:downstream}d, the prediction performance varies considerably over debate topics. For instance, the user beliefs related to `religion' and `philosophy' are more predictable than those under `sports,' `funny,' and `games.' This discrepancy remains consistent even after downsampling the dominant categories, resulting in a training dataset with a relatively more balanced distribution of debates across categories (see SI Section 6 and Fig.~S20).

Third, users can exhibit varying distributions of beliefs within the belief space. Despite having the same number of prior beliefs (voting history), some users display a broader distribution of beliefs, while others show more concentrated beliefs in a smaller region. To investigate how belief selection patterns differ between these groups, we quantify the dispersion of a user's prior beliefs using a metric called the \emph{effective radius} ($r_g$). This measure, analogous to the radius of gyration, is defined for a user $u$ as:
\begin{equation}
    r_g^u = \sqrt{\sum_{i=1}^{N_u}\|\vec{b}_{i}^{u}-\vec{u}\|^2 / N_u},
\end{equation} where $\vec{b}_i^u$ denotes $i$-th prior belief vector of the user $u$, and $\vec{u}$ represents the centroid of the user's prior beliefs. Here, $N_u$ is the total number of prior beliefs of user $u$.

Figure~\ref{fig:downstream}e illustrates the relationship between effective radius and average belief prediction accuracy, where users are grouped into five quantiles based on their number of prior beliefs. Across all user groups, belief prediction accuracy decreases as $r_g$ increases, indicating that individuals with more concentrated beliefs (smaller $r_g$) are more likely to select a belief closer to their prior ones between two opposing beliefs in an unseen debate. By contrast, users with more dispersed beliefs are  more likely to select the belief farther from their prior beliefs. It is important to note that, since our model always assumes that users will choose a belief closer to their prior belief, prediction accuracy can be directly interpreted as the probability of a user choosing the closer belief. Therefore, for the remaining analyses, we use the accuracy score instead of the F1-score for clearer interpretation, a decision further supported by the fact that the two scores are highly correlated.

Fourth, the proximity of a user to the beliefs under consideration in the belief space significantly impacts prediction accuracy (Figs.~\ref{fig:downstream}f-h). When two opposing beliefs are introduced in a new debate during the prediction task, we measure their distances from a user: $d_\text{min}$ for the closer belief and $d_\text{max}$ for the farther belief. Our analysis shows that the prediction accuracy is inversely correlated with $d_\text{min}$ and positively with $d_\text{max}$ (Figs.~\ref{fig:downstream}f and g). 
The heatmap in Fig.~\ref{fig:downstream}h illustrates how the average accuracy varies across the two-dimensional space defined by $d_\text{min}$ and $d_\text{max}$. This suggests that the probability of choosing the closer belief decreases as $d_\text{min}$ increases, and increases as $d_\text{max}$ grows. Consequently, predictions are more accurate when the closer belief is much closer to the user and the farther belief is much farther away.

We further assess the impact of the average distance $d_\text{avg}$ between the user and two opposing beliefs on the prediction accuracy (Fig.~S14). As $d_\text{avg}$ increases, the accuracy converges to 0.5, equivalent to random guessing. This suggests that when both beliefs are sufficiently distant from the user's position (when $d_\text{avg}\approx 33$), predicting the user's choice becomes extremely difficult. A large $d_\text{avg}$ indicates that the debate introduces viewpoints that are distant from or weakly associated with the user's prior beliefs, consequently reducing the predictive power of past voting behavior. We conducted comprehensive robustness checks under various data-splitting and testing scenarios, confirming the consistency of our findings across different conditions (see SI Section 6).

In addition to analyzing these factors, we examined whether beliefs varied in predictability across user groups based on political party, religion, or gender. However, we found no significant differences in predictive accuracy across these groups (Fig.S15).

\subsection*{Role of relative dissonance in belief prediction}

\begin{figure*}[t!]
    \centering
	\includegraphics[width=0.65\textwidth]{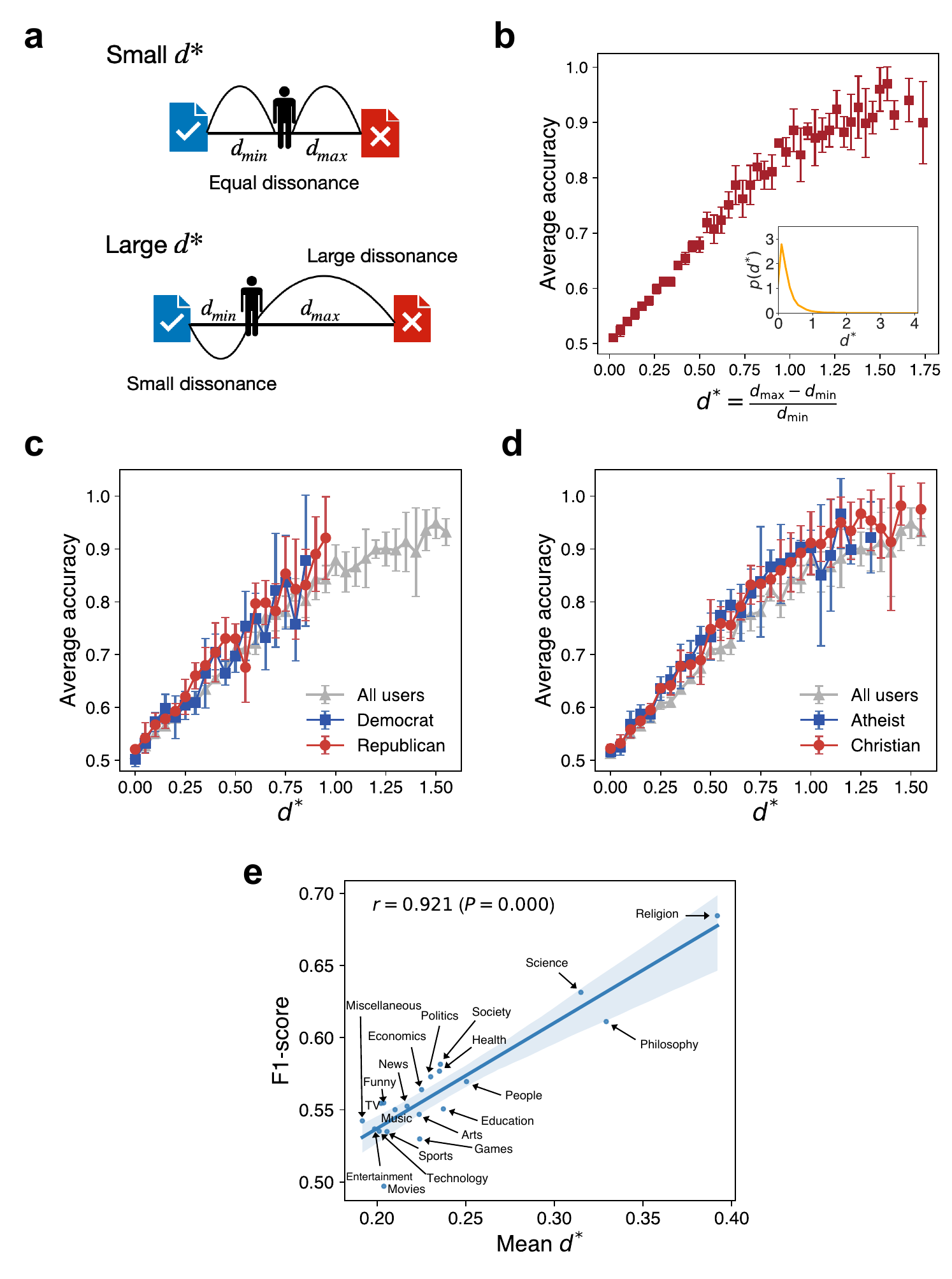}
	\caption{\textbf{Effect of relative dissonance on belief selection.} (a) Illustration of two scenarios in which a user selects a belief for a debate, contrasting cases of small and large relative dissonance. When the two beliefs under consideration are equally distant from the user, selecting either belief may result in an equal level of dissonance. By contrast, when one belief is significantly farther than the other, the potential dissonance a user may experience varies depending on their selection. (b) The likelihood of a user choosing a closer belief linearly increases with relative dissonance $d^{*}$. The inset shows the distribution of $d^{*}$. (c) and (d) Linear relationships between average accuracy in the belief prediction task and $d^{*}$, shown separately for two user groups: Democrats vs. Republicans (c), and Christians vs. Atheists (d). The results show a similar linear relationship regardless of users' political or religious ideologies. Error bars in (b) to (d) represent the standard deviation across the results of the 5-fold validation tasks. (e) Average $d^{*}$ for different debate topics within the prediction task and the corresponding average prediction score across topic areas. Topics with higher mean $d^{*}$ tend to have more accurate predictions. The solid regression line represents this trend, with the shaded area indicating the 95\% confidence interval.}
	\label{fig:belief_dissonance}
\end{figure*} 

Our findings from the belief prediction task reveal that the distances between a user and two opposing beliefs under consideration ($d_\text{min}$ and $d_\text{max}$) significantly impact prediction accuracy. The underlying patterns of belief selection provide important insights into human decision-making mechanisms.

Based on these empirical observations, we propose a new metric, termed `relative dissonance' and denoted as $d^{*}$, to better quantify this decision-making process:
\begin{equation}
d^{*}=\frac{d_\text{max}-d_\text{min}}{d_\text{min}}.
\end{equation} 
$d^{*}$ represents the absolute difference between a user's distances to two opposing beliefs, normalized by the shorter distance, $d_\text{min}$.

From the perspective of cognitive dissonance, a belief closer to a user's existing beliefs (i.e., smaller $d_\text{min}$) is expected to result in less dissonance, while a belief that is farther away would induce greater dissonance (Fig.~\ref{fig:belief_dissonance}a). Thus, $d^{*}$ serves as a relative measure of dissonance reduction when a user opts for the belief closer to their prior beliefs rather than a more distant belief.

We note that we use the term `dissonance' in a broad sense to represent the distance between a user and a belief. In our framework, a user vector is computed by averaging their belief vectors, capturing the average position of the beliefs held by the user. The distance between a user vector and a new belief vector reflects, on average, how much the new belief deviates from (or is dissonant with) the user's prior beliefs. In this context, dissonance is treated as a continuous measure, analogous to distance. Smaller dissonance values indicate alignment or favorability toward the belief, whereas larger dissonance values reflect greater deviation from the user's prior beliefs.

Figs.~\ref{fig:belief_dissonance}a and b illustrate a compelling relationship between $d^{*}$ and the average prediction accuracy of users' beliefs on new debates: the average accuracy shows a linear increase with the rise of $d^{*}$. As we noted previously, the prediction accuracy can be interpreted as the probability that a user selects a belief closer to their prior beliefs. Thus, the observed linear increase suggests that this probability depends on relative dissonance. In other words, when the potential dissonance from one belief outweighs that from another, users are more likely to choose the belief that is closer to them.

When $d^{*}$ is near 0, the probability of a user choosing a closer belief is around 50\%, suggesting that their decisions are largely independent of their prior beliefs (Fig.~\ref{fig:belief_dissonance}b). Conversely, for debates when $d^{*}$ is larger (e.g., $d^{*}\approx1.5$), users exhibit a strong preference for beliefs closer to their position, with probability close to 1. In this scenario, users are strongly inclined to select beliefs closely aligned with their prior ones, indicating a significant reduction in potential dissonance when avoiding an alternative belief.

We further investigate whether user groups with distinct political or religious ideologies---specifically, comparisons between Democrats and Republicans, as well as Christians and Atheists---exhibit different decision-making patterns with respect to relative dissonance $d^{*}$. As shown in Figs.~\ref{fig:belief_dissonance}c and d, we find no significant difference in how relative dissonance influences belief selection between two groups ($p > 0.05$ in two-sample t-tests across all $d^{*}$ ranges). These results suggest that the impact of relative dissonance on belief selection is remarkably consistent across different political and religious groups.

The concept of relative dissonance also helps to explain why users' beliefs on certain debate topics are more predictable than others.
We found a strong correlation ($r=0.921$) between the average $d^{*}$ of a debate category and its prediction F1-score (Fig.~\ref{fig:belief_dissonance}e). This high correlation partly explains why users' belief choices on certain topics (e.g., `religion' or `philosophy') are more predictable than in others (e.g., `funny' or `entertainment'). For example, the difference in distance from a user to two opposing beliefs tends to be much larger in the belief space for debates on `religious' topics than for those on `funny' or `entertainment.'

To assess whether the correlation between a category's average $d^{*}$ and prediction accuracy is merely a byproduct of category size, measured by the number of training examples per category, we analyze the relationships among mean $d^{*}$, average F1-score, and category size (see SI Section 6E). Our robustness checks indicate that while $d^{*}$ and category size are partially correlated, $d^{*}$ consistently exhibits a stronger and more robust correlation with prediction accuracy than category size. This trend becomes even more pronounced when frequently occurring debate categories are downsampled.

\section*{Discussion}\label{sec:Discussion}

Here, we demonstrate that neural embedding approaches based on LLMs offer a powerful and scalable solution for understanding the complex and nuanced relationships among human beliefs. While previous approaches provide insightful theoretical bases for modeling belief systems that incorporate belief relationships, there has been a lack of robust frameworks to comprehensively represent the space of beliefs encompassing a wide range of topics~\cite{galesic2021integrating, introne2023measuring, aiyappa2023weighted}. Existing methods, which often rely on surveys and small, topic-specific datasets, lack scalability and face challenges in capturing the full spectrum of beliefs individuals hold.

In this perspective, LLMs integrated with user activity data can open a new avenue for modeling human beliefs. Pretrained language models, which already possess a strong understanding of complex language patterns and contextual information, can be fine-tuned using extensive belief records to create a comprehensive ``embedding space of human beliefs.'' This embedding space maps a wide range of topics and enables inductive reasoning about new beliefs. Furthermore, this approach efficiently represents an individual's belief system and supports various downstream tasks such as quantifying polarization or predicting beliefs.

The key findings from our study offer several insights into the characterization of human beliefs. First, our study introduces a representative learning framework for constructing a belief embedding space in a continuous high-dimensional vector space using online user activities and LLM. This space effectively reveals the interconnected structure of various human beliefs and the polarization of beliefs related to representative social issues. The continuous belief space created using the fine-tuned LLM facilitates inductive reasoning, enabling the addition of new beliefs.

Second, the vector representation of individuals allows us to identify how people with different opinions are clustered and polarized. The fine-tuned S-BERT model reveals a clearer separation among individuals with similar political or religious ideologies, whereas the base S-BERT model without fine-tuning does not exhibit such patterns. Our results demonstrate the usefulness of the belief space in measuring the polarization of certain social concepts. The distance between groups of individuals with opposing beliefs on a given issue within the belief space is highly correlated with the degree of political polarization associated with that issue.

Third, the downstream task for belief prediction shows that the proposed belief space is useful for predicting individuals' beliefs on new debates based on their pre-existing beliefs. We uncover four critical factors that influence the prediction outcome of an individual's choice of a new belief: the length of individuals' voting records, debate categories, effective radius of individuals, and the distances between the individual and the two beliefs under consideration in the belief space.

Most importantly, our empirical observations highlight that the relative distance between an individual and two opposing beliefs in a new debate is a reliable predictor of their decision. This insight lead us to develop a novel metric called `relative dissonance' $d^{*}$, which quantifies the relative inconsistency a person may experience when adopting a belief into their pre-existing belief system compared to its opposite belief (Fig.~\ref{fig:belief_dissonance}). Our analysis reveals that as the relative dissonance ($d^{*}$) increases, the likelihood of a person choosing a belief closer to their current position in the belief space increases. In other words, the greater the difference in dissonance a person experiences between two beliefs, the more likely they are to choose the belief that causes less dissonance. This finding aligns with conventional cognitive dissonance theory and offers a quantitative measure of cognitive dissonance by linking it to distances within the belief space.

While our model captures many aspects of human belief dynamics, our study does have limitations that will guide future research. 
First, the reliance on a single online debate platform for data collection may limit the generalizability of our findings. Incorporating broader datasets from diverse platforms will help understanding the universal properties of belief systems and their cultural and social variations. Additionally, the dataset used in this study is primarily based on U.S. data, which may not fully represent global perspectives and cultural diversity in human beliefs. Future research should include data from various societies to improve broader relevance of the findings across different cultural contexts.

Second, the DDO dataset used in this study, where users' preferences are easily inferred from explicit voting records, represents a specific data type. Developing methods for extracting human beliefs from more general texts on diverse platforms, such as social media postings, news interviews, and movie scripts, would provide a deeper understanding of human beliefs and significantly increase the applicability of our framework.

Third, our study does not investigate the temporal and dynamical properties of the belief space. Although our study indirectly assumes the stability of the belief space, in reality, a society's beliefs on social issues can continuously change. Investigating how the shape of the entire belief space, which reflects the interconnections of collective societal beliefs, transforms over time would be an interesting avenue for future research.

Fourth, there is a concern regarding the inherent biases present in the pre-trained LLMs used in our study~\cite{gallegos2024bias}. For example, LLMs trained predominantly on English-language internet data may inadvertently reflect Western-centric viewpoints, underrepresenting or misrepresenting beliefs prevalent in non-Western cultures. These models might exhibit biases related to gender, race, and socioeconomic status, which could skew the analysis of the belief relationships. Ongoing efforts to improve fairness and reduce biases in LLMs are crucial for future research to ensure more equitable and accurate representations of human beliefs.

Looking ahead, while our primary goal in this study is to create a comprehensive map of beliefs and uncover the mechanisms behind human belief selection, our contrastive learning approach also shares certain core principles with recommendation systems~\cite{hu2008collaborative, koren2009matrix, kim2024large}. We anticipate that our contrastive learning methods---extracting both positive and negative relationships from user activities as well as utilizing the semantic understanding of LLMs---could be effectively applied in recommendation algorithms. Moreover, integrating the cognitive patterns and belief dynamics revealed in this study may enable recommendation systems to better reflect how human beliefs evolve and interact, ultimately leading to more personalized and context-aware suggestions.

In essence, our research establishes a foundational framework for an advanced, data-driven analysis of human beliefs using LLM. We anticipate that this work on the complex landscape of human beliefs would provide both theoretical insights and practical applications in understanding and modeling human behavior in the fields of cognitive science, social psychology, political science, and beyond.

\section*{Methods}\label{sec:Methods} 

\subsection*{Debate.org dataset and extraction of belief statements}
\label{sec:ddo_dataset}
The Debate.org (DDO) dataset used in this study contains a corpus of 78,376 debates (68,900 unique debate titles excluding duplicates) by 42,906 debaters from October 15, 2007, to September 19, 2018 (Figs.~S1 and S2). In DDO, each debate features two debaters, one supporting the proposition (PRO) and the other opposing it (CON). In each debate, other users can engage by voting on seven different items. Notably, the option ``Agree with after the debate'' enables users to express their position on the debate topic as either PRO, CON, or TIE, reflecting their belief on the issue. To extract belief pairs that reveal clear positive and negative relations, we only considered the PRO and CON votes and excluded TIE votes. We also treated debaters and voters equally as voters, as our study utilizes users' positions on various debate topics as their beliefs.  

Most debate titles in DDO represent beliefs on various topics (e.g., ``Abortion should be legal,'' ``God exists,'' ``All morals are relative''). Thus, users' votes on these titles as PRO or CON can be considered as revealing their beliefs on these topics. To generate a complete belief statement for a user, we appended a template phrase that explicitly describes the user's stance. For example, a PRO (or CON) vote on a debate title leads to the belief statement, ``I agree (disagree) with the following: [DEBATE TITLE].'' For instance, a PRO vote on `Abortion is morally justified' results in the belief statement, ``I agree with the following: Abortion is morally justified.'' These belief statements are then fed into LLMs.

We performed data filtering on the DDO dataset to make it suitable for our analyses. While most debate topics in DDO can be considered in the form of beliefs that allow for support or opposition, there are also incomplete or unsuitable titles that cannot be regarded as beliefs. We filtered these unsuitable debate titles using GPT-4~\cite{openai2023gpt4, achiam2023gpt}, one of the most advanced and reliable artificial intelligence language models at the time of our study. We asked GPT-4 to determine whether a given statement (debate title) can be considered a human belief (see SI Section~2, Table~S1, and Figs.~S3 and S4).

Among 68,900 unique debate titles, GPT-4 classified 8,914 as unsuitable for consideration as belief statements. The unsuitable debate titles include titles that use \textit{versus} or \textit{vs.}, such as ``Batman vs. Spiderman'' and ``atheism vs. agnosticism'', titles denoting battle content like ``Rap battle,'' ``music battle,'' and ``Video Rap battle,'' titles with single words without meaningful context or incomplete sentences, for instance, ``fox news,'' ``useless,'' ``Media are ...,'' titles posing how questions like ``How many donuts are too many donuts,'' ``How can you be an atheist?'' as well as titles expressing personal resolutions or suggestions like ``I will not contradict myself'' and ``I will lose this debate.'' Removing 8,914 inadequate debate titles resulted in 59,986 unique debate titles (from 65,861 debates) that were voted on a total of 192,307 times by a total of 40,280 users.

To assess the consistency of classification results using GPT-4 with human annotations, we compared its classifications of 50 randomly sampled debate titles against those determined by three human annotators (three of the authors on this study). We equally sampled 25 titles from each category of `True' and `False,' as classified by GPT-4, to ensure balanced representation. The annotators were requested to indicate whether or not the debate titles qualify as belief statements. The inter-annotator reliability, measured using Fleiss' Kappa---which quantifies agreement beyond chance---was 0.866, indicating a high level of agreement among the human annotators. GPT-4's classifications showed an 88\% agreement rate with the majority vote of the human annotators. This high agreement rate suggests the strong consistency between GPT-4's classifications and the consensus among human annotators in identifying belief statements.

\subsection*{Training LLMs with belief triplets to build belief space}\label{sec:llm_models}

We employed a pre-trained S-BERT model (roberta-base-nli-stsb-mean-tokens)~\cite{reimers2019sentence}, based on the RoBERTa model~\cite{liu2019roberta}, to learn relationships between beliefs across multiple topics. Using belief triplets, we applied a contrastive learning technique to fine-tune the model. We explored various LLMs, from the original BERT~\cite{devlin2018bert} to other S-BERT models pre-trained with different sources. The RoBERTa-based model exhibited superior performance in diverse tasks and was thus selected for our study.

For the fine-tuning process, we created belief triplets using the voting records of users. A user's voting records on various debates create a sequence of beliefs. Using these belief sequences, we produced a set of belief triplets. Each of the belief triplets comprises three distinct beliefs: an anchor belief statement $B_a$, a positive example belief $B_p$, and a negative example belief $B_n$. We went through all belief statements as anchor beliefs and found corresponding positive and negative examples. The positive example beliefs for a given anchor were sampled from the beliefs that were voted on together with the anchor belief, weighted by their frequency (the more often two beliefs are voted on by the same users, the more likely they are to be sampled as positive examples). Conversely, the negative example beliefs of an anchor belief were selected from either the directly opposing belief statement (expressing an opposite opinion towards the anchor belief) or from the beliefs that were co-voted with the opposite belief statement.

For example, assume that many users frequently voted as PRO to the debates titled ``Abortion is morally justified'' and ``Same-sex marriage should be legal.'' Then, for the anchor belief, ``I agree with the following: Abortion is morally justified,'' a possible positive example could be ``I agree with the following: Same-sex marriage should be legal,'' and a negative example could be ``I disagree with the following: Abortion is morally justified.'' In this way, we sampled at most five positive examples and five negative examples for a given anchor belief statement, and generated all possible combinations of belief triplets based on these examples; A maximum of 25 triplets can be created for one anchor belief.

We note that the same pair of beliefs may appear both as a positive and negative example in different proportions. For example, a belief pair could be co-voted together (positive) by a majority of users yet be opposed (negative) by a minority. By including all such variations, our model learns a weighted, continuous measure of similarity, enabling us to move beyond a simplistic binary determination of ``similar'' versus ``dissimilar.''

The belief triplets were fed into the pre-trained S-BERT model. We divided debates into training and test data in an 8:2 ratio, repeating this process 5 times for 5-fold validation datasets. On average, 1,354,123 triplets were used for the fine-tuning process as training sets. The model was fine-tuned to minimize the triplet loss function $L$,
\begin{equation}
    L = \max(\lVert s_a - s_p \rVert - \lVert s_a - s_n \rVert + \epsilon , 0),    
\end{equation}
where $s_a$, $s_p$, and $s_n$ are the 768-dimensional output vectors of S-BERT corresponding to the sentence embedding of an anchor belief $B_a$, a positive belief $B_p$, and a negative belief $B_n$, respectively. $\epsilon$ is the triplet margin term, which guarantees that the negative belief vector $s_n$ must be farther away from the anchor $s_a$ than the positive belief vector $s_p$. We used the default parameter $\epsilon=5$.

During training, the weight parameters of the S-BERT model are updated in order to minimize the Euclidean distance between $s_a$ and $s_p$, while simultaneously maximizing the gap between $s_a$ and $s_n$. The fine-tuned model thus provides a comprehensive 768-dimensional latent representation of human beliefs, termed the \textit{belief space}. When belief statements are inputted into the LLM, it outputs their vector representations that form this belief space, where the positions and distances between beliefs reveal interdependencies between them.

\subsection*{Belief prediction with a larger LLM in a few-shot setting}\label{sec:DT-zeroshot}

During the downstream task, which involves predicting user beliefs on unseen debates, we benchmarked our results against the performance of Llama2 (Llama2-13b-chat)~\cite{touvron2023llama}, a recent LLM with much larger parameters, for few-shot tasks. We chose Llama2 as it exhibits strong zero/few-shot performance across a variety of tasks such as question answering and natural language reasoning. For our task, Llama2 was prompted with a user's existing beliefs from the training set and tasked with predicting the user's stance on new, unseen debates. After testing several prompts, we chose a prompt for Llama2 that includes a user's prior belief statements, followed by a query: ``Based on these statements, do you think you might agree or disagree with the following: \{DEBATE TITLE\}? Please choose from one of these options: agree or disagree. Do not explain your choice.'' This approach required the model to make a binary decision, answering either `agree' or `disagree.' We were able to test only on approximately 85\% of the dataset due to the context-size limitations of Llama2.

\subsection*{Ethics}
Our study does not involve any human subjects or experiments and is not subject to Institutional Review Board (IRB) approval. Consequently, there were no ethical regulations to comply with, no informed consent was required, and no participant compensation was involved. Additionally, our study was not preregistered.

\section*{Data availability}
The original DDO dataset~\cite{durmus2019exploring, durmus2019corpus} is publicly accessible and can be downloaded from \url{https://esdurmus.github.io/ddo.html}. For the replication of our study, a processed version of this dataset, including pre-processed user-level debate records and the fine-tuned models used in our analyses, is available at \url{https://github.com/ByunghweeLee-IU/Belief-Embedding}.

\section*{Code availability}
We developed custom code using Python 3.9.10 for data analysis. The replication code is available at \url{https://github.com/ByunghweeLee-IU/Belief-Embedding}.

\section*{Acknowledgments}
B.L. and Y.Y.A. are supported in part by the Air Force Office of Scientific Research under award number FA9550-19-1-0391. Y.Y.A. were supported in part by DARPA under contract HR001121C0168. B.L., R.A., J.A., H.K., and Y.Y.A. are in part supported by the Air Force Office of Scientific Research under award number FA9550-25-1-0087. H.K. is supported by the Luddy Faculty Fellow Research Grant Program of the Luddy School of Informatics, Computing, and Engineering at Indiana University Bloomington. 

\section*{Author Contributions}
B.L., H.K., and J.A. conceived the research. B.L. and R.A. performed the empirical analyses. B.L., R.A., Y.A., H.K., and J.A. discussed and interpreted the results, and wrote the manuscript.

\section*{Competing interests}
The authors declare that they have no competing interests.

\clearpage

\clearpage

\newpage
\appendix
\renewcommand{\appendixname}{}

\renewcommand{\thefigure}{S\arabic{figure}}
\renewcommand{\thesection}{S\arabic{section}} 
\renewcommand{\thesubsection}{\Alph{subsection}}
\renewcommand{\thesubsubsection}{\arabic{subsubsection}} 
\renewcommand{\thetable}{S\arabic{table}}

\setcounter{figure}{0}
\setcounter{table}{0}
\setcounter{section}{0}

\section*{Supplementary Information}
\section{Basic statistics}

\begin{figure*}[h!]
\centering
\includegraphics[width=0.9\textwidth]{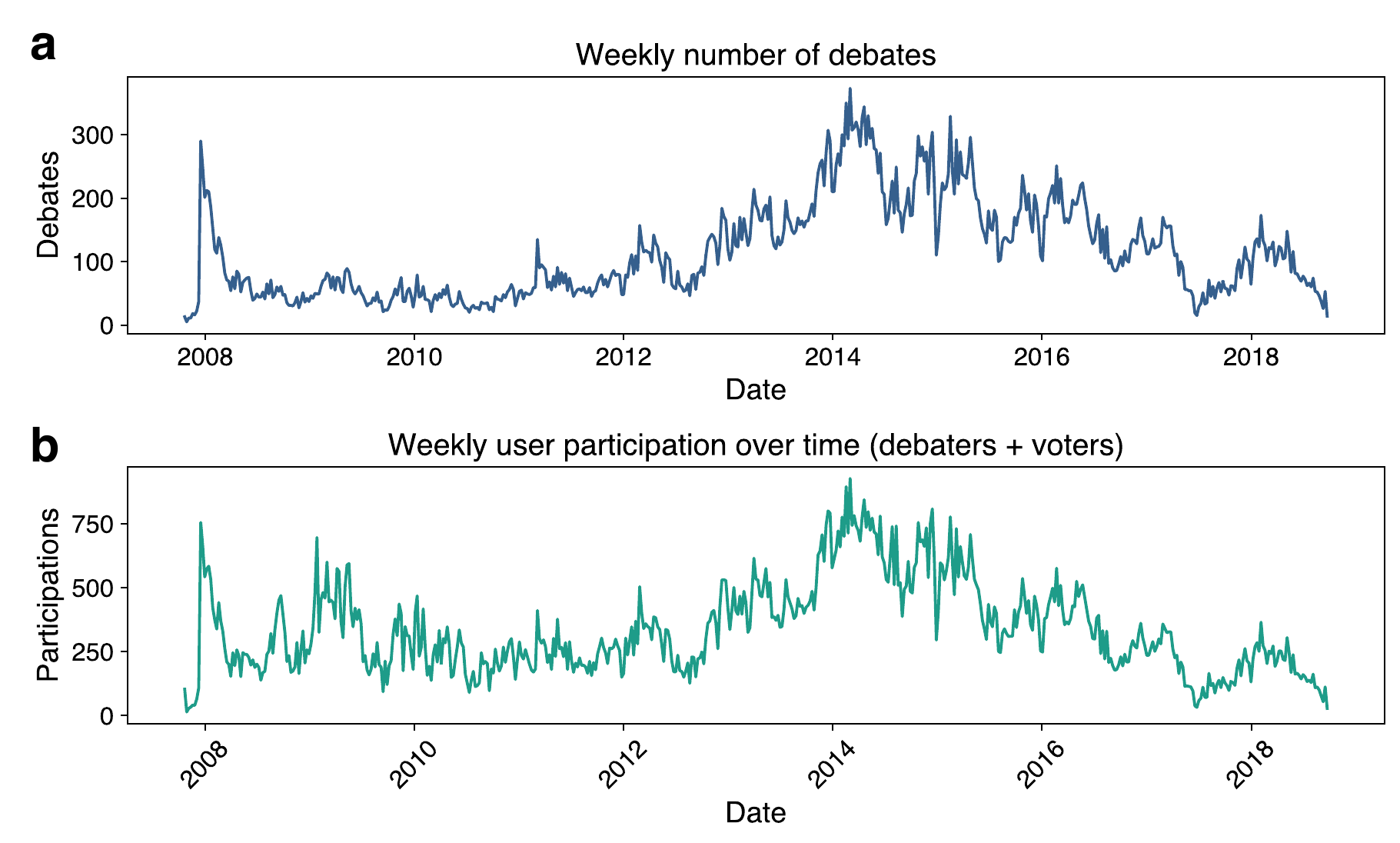}
	\caption{(a) Weekly number of debates over time. The curated DDO dataset utilized in this study comprises a corpus of 65,861 debates, spanning from October 15, 2007, to September 19, 2018. (b) Weekly number of participants (both debaters and voters) in DDO dataset.}
	\label{fig:timespan}
\end{figure*}

The Debate.org (DDO) dataset~\cite{durmus2019corpus, durmus2019exploring} utilized in this study comprises a corpus of 78,376 debates, spanning from October 15, 2007, to September 19, 2018. The pre-processing described in the Methods section of the main manuscript resulted in a curated subset of 65,861 debates with 59,986 unique debate titles. Fig.~\ref{fig:timespan} shows the number of debates and user participation, in roles such as debaters and voters, over time during the period under consideration.

\begin{figure*}[t!]
\centering
\includegraphics[width=0.8\textwidth]{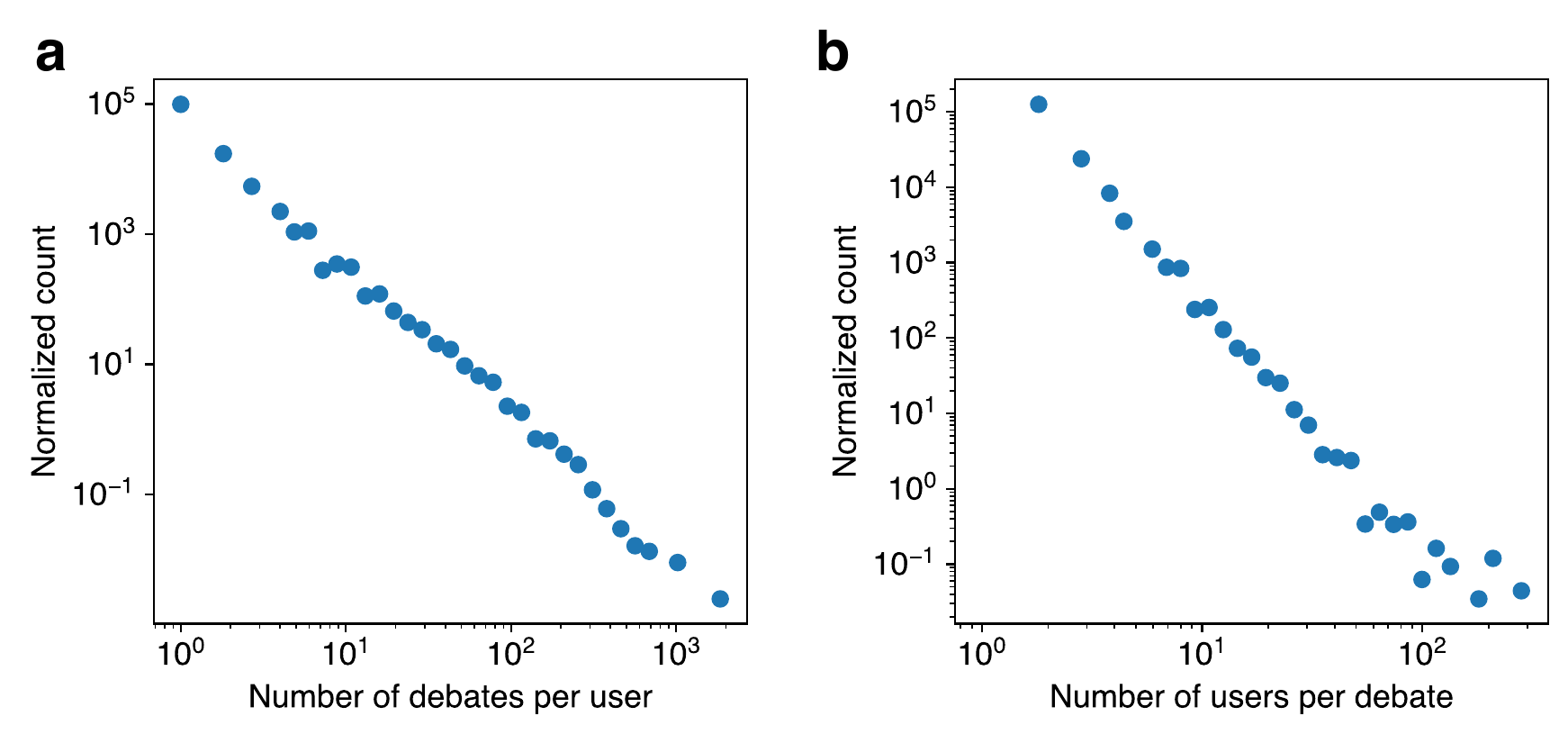}
	\caption{Distribution of the number of debates participated in per user (a) and user participation per debate (b), where the Y-axes represent the log-binned distributions of normalized counts. Both distributions exhibit heavy-tailed characteristics, implying that most users participate in a small number of debates, while a smaller but significant number of users are involved in many debates. Similarly, there are a smaller but significant number of debates in which many users participated. } 
	\label{fig:debate_voter_distribution}
\end{figure*}

Figure~\ref{fig:debate_voter_distribution} demonstrates the basic statistics of user participation. The distribution of the number of debates in which users participated by voting follows a heavy-tailed distribution (Fig.~\ref{fig:debate_voter_distribution}a). The heterogeneity in user participation indicates that most users participate in a small number of debates, while a smaller but significant number of active users are involved in many debates. The number of voters per debate also follows a heavy-tailed distribution (Fig.~\ref{fig:debate_voter_distribution}b), demonstrating that most debates have very few voters, but there are still some significant number of debates where hundreds of people participated into voting.

\section{Data pre-processing}

\begin{figure*}[t!]
\centering
\includegraphics[width=\textwidth]{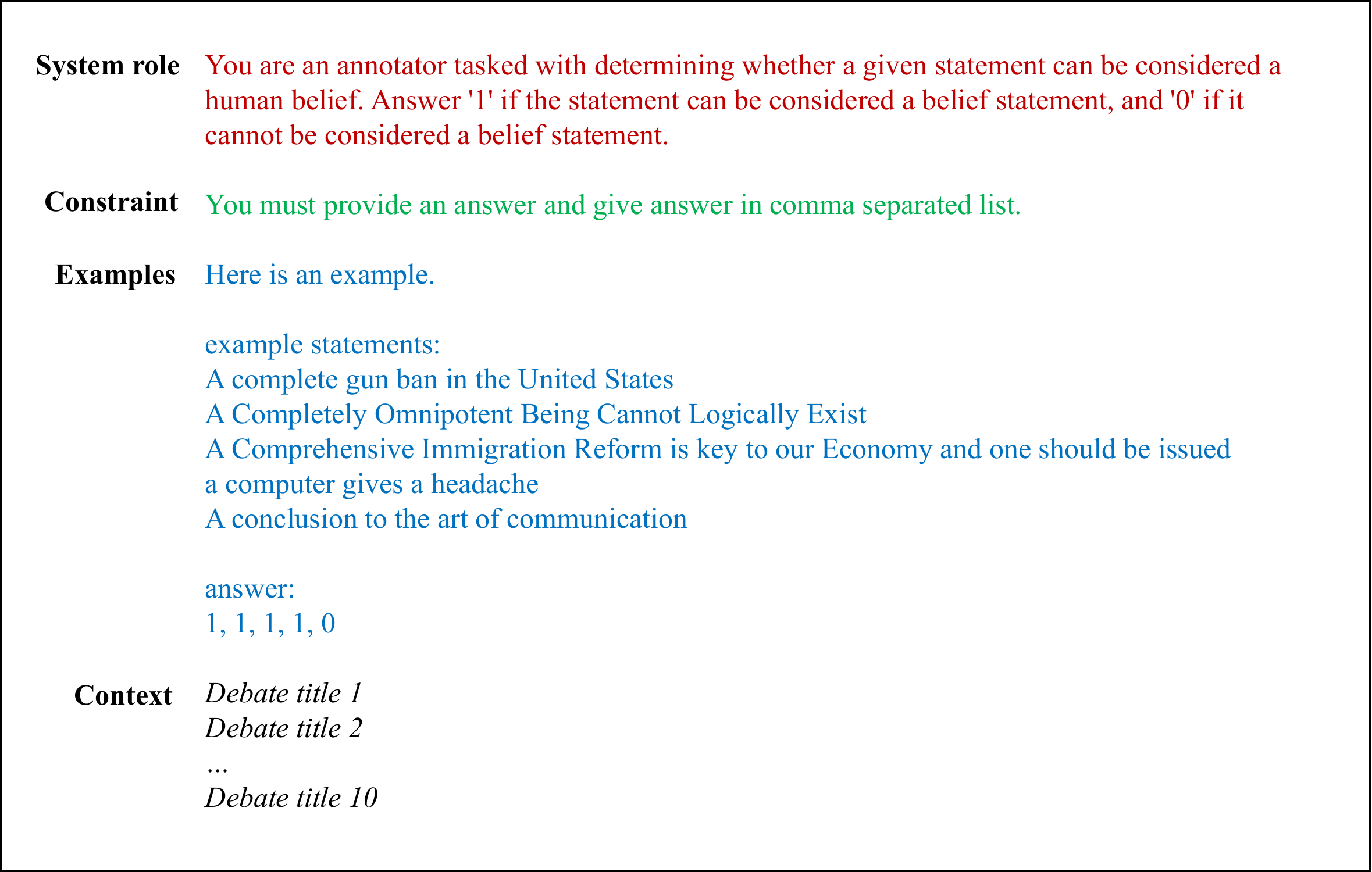}
	\caption{Prompt used to filter unsuitable debate titles using GPT-4. We constructed a prompt that instructs the model to assess whether each debate title can be considered human belief.}
	\label{fig:gpt4}
\end{figure*}

In the pre-processing stage, we filtered out unsuitable debate titles using GPT-4, one of the most advanced and reliable artificial intelligence language models available at the time of the experiment~\cite{openai2023gpt4}. We designed a prompt to ask GPT-4 to determine whether a given statement (debate title) could be considered a human belief. Fig.~\ref{fig:gpt4} illustrates a sample prompt provided to GPT-4. Out of 68,900 unique debate titles, GPT-4 classified 8,914 as unsuitable for consideration as belief statements.

To evaluate the consistency of GPT-4's classification results with human annotations, we compared its classifications of 50 randomly sampled debate titles with those determined by three annotators (three of the authors). We equally sampled 25 titles from each category of `True' and `False,' as classified by GPT-4, to ensure balanced representation. The three annotators independently classified the 50 debate titles into two categories---those considered to reflect human belief and those that are not---using questions similar to those used for GPT-4's categorization. The sampled debate titles are listed in Table~\ref{tab:belief_annotation}.

\begin{figure*}[t!]
\centering
\includegraphics[width=\textwidth]{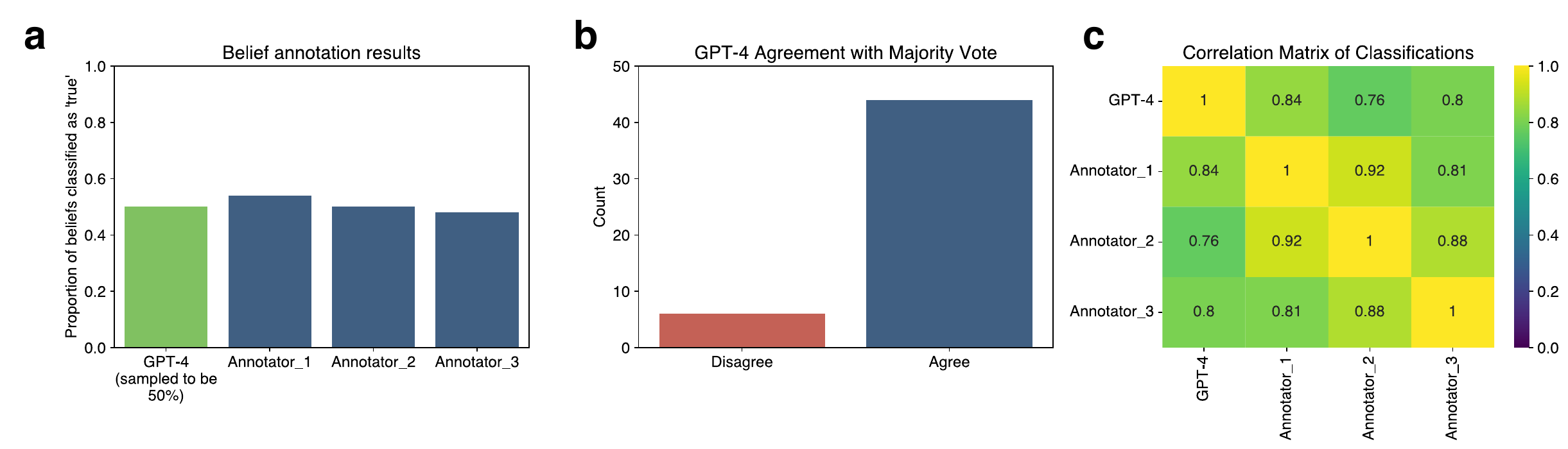}
	\caption{Validation of the belief filtering process using GPT-4 by comparison with human annotation. We sampled 25 titles from each category of `True' and `False,' as classified by GPT-4. Then three annotators (three of the authors) classified the 50 debate titles into two categories: those that are considered human belief and those that are not. (a) Proportions of beliefs classified as `True' by each annotator. Similar outcomes from annotators suggest that there is a general agreement on classification results among them. (b) The number of agreements and disagreements between GPT-4's classifications and the majority vote from human annotators. The high count of agreements (44 out of 50, 88\% agreement rate) emphasizes GPT-4's alignment with human consensus. (c) correlation matrix of classifications by human annotators. Strong positive correlations indicate consistent agreement between annotators on the classification of statements as beliefs or non-beliefs.}
	\label{fig:annotation_result}
\end{figure*}

GPT-4's classifications showed an 88\% agreement rate with the majority vote of the human annotators (Fig.~\ref{fig:annotation_result}a and b), suggesting strong consistency between GPT-4's classifications and human consensus in identifying belief statements. The inter-annotator reliability, measured by Fleiss' Kappa, was 0.866, indicating a high level of agreement among the human annotators (Fig.~\ref{fig:annotation_result}c).

\section{Evaluation of template variability on belief embeddings}

In this study, we considered the combination of a user's vote position (PRO/CON) and the debate title as a belief statement. To articulate these belief statements in sentence form, we used template phrases, ``I agree/disagree with the following: '' followed by the debate title. We tested whether a belief space trained using one template remained robust to variations introduced by alternative templates.

We employed S-BERT, fine-tuned with belief statements generated using the original template (``I agree/disagree with the following: ''), and tested whether the model performed similarly with other templates. To this end, we rephrased the belief statements using three alternative templates, as shown in Table~\ref{tab:other_templates} (e.g., ``I support(/do not support) the following statement: ,'' ``I am in accord(/discord) with the following: ,'' etc.). We then conducted triplet evaluation tasks, which involved assessing whether positive belief pairs were closer in Euclidean distance than negative pairs. The triplet evaluation accuracy was calculated for both the original template and the rephrased templates using the S-BERT fine-tuned with the original template.

Table~\ref{tab:template_eval} presents the triplet evaluation accuracy measured for the belief set using the original template and the same belief set rephrased with alternative templates. Despite variations in template phrasing, the triplet evaluation accuracy remained consistent across different templates, demonstrating that the model effectively distinguished between positive and negative relationships in belief pairs, regardless of template variation.

\begin{figure*}[t!]
\centering
\includegraphics[width=\textwidth]{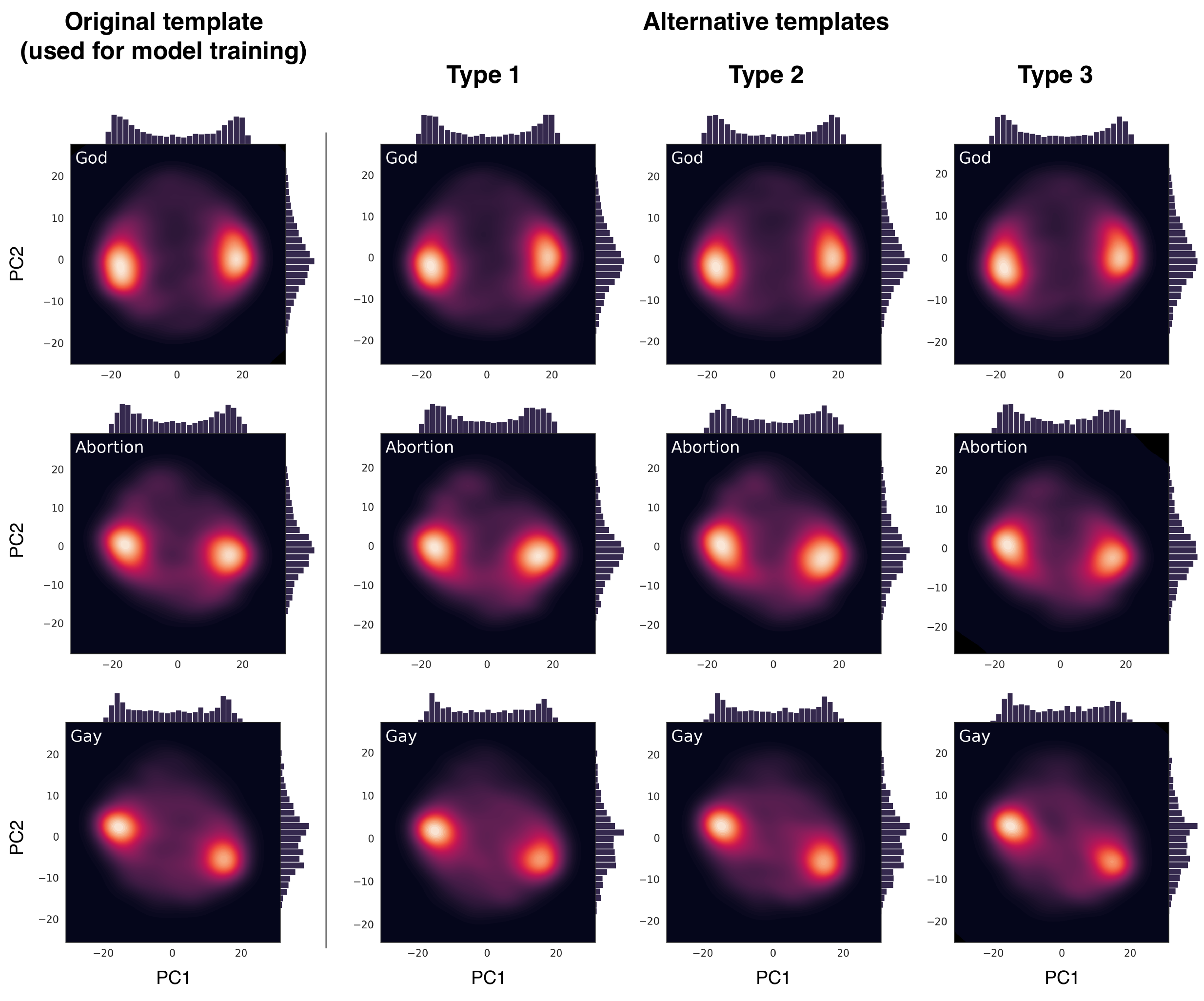}
	\caption{The belief embeddings associated with three keywords (`God', Abortion', and `Gay') are represented in the first two principal component (PC) spaces. The first column represents the belief embeddings generated using the original template (``I agree/disagree with the following: ''). The other three columns represent the same belief set, but using the output vectors of beliefs rephrased with three alternative templates as shown in Table~\ref{tab:other_templates}. The same model, which was fine-tuned using the beliefs with the original template, was used for generating the embeddings of rephrased beliefs. The PC space was calculated using the original belief embeddings. The bimodal shape of the belief distribution is preserved across different template choices.}
	\label{fig:other_template_PC}
\end{figure*}

We also visualized the belief embeddings in the first two principal component (PC) spaces for three example keywords: `God,' `Abortion,' and `Gay' to evaluate whether the distribution of belief embeddings remains consistent across different templates. Fig.~\ref{fig:other_template_PC} shows that the embeddings for these three keywords exhibit similar distributions across various template phrases. The panels in the first column demonstrate embeddings generated using the original template, while the other columns present embeddings for the same beliefs rephrased with three alternative templates. The bimodal distribution of belief embeddings was preserved across different template choices, indicating the model's robustness to template variability.

Overall, our findings indicate that the belief space fine-tuned with the original template remains both effective and consistent, even when alternative templates are applied. This highlights the model's robustness and reliability in managing varied phrasings of belief statements.

\section{Belief embedding} 

\subsection{Belief space before and after fine-tuning}
This section examines the transformation of the belief embedding space before and after fine-tuning the S-BERT model, highlighting its impact on capturing contextual relationships between beliefs.

\begin{figure*}[t!]
    \centering
	\includegraphics[width=0.8\textwidth]{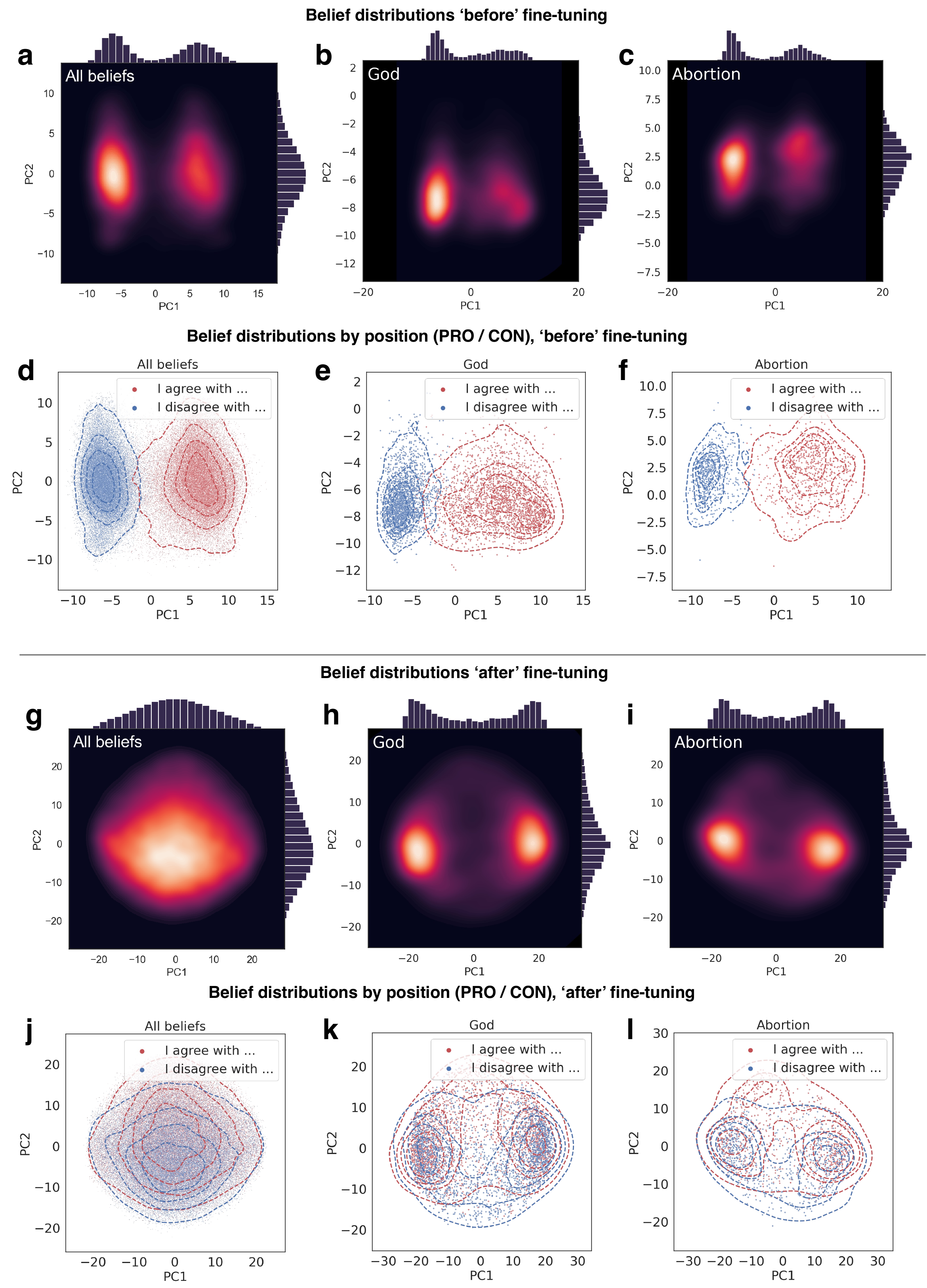}
	\caption{Change in belief space before and after fine-tuning S-BERT.
(a) shows the distribution of all belief vectors in the PCA space generated by the base S-BERT model (pre-fine-tuning). (b) and (c) depict the distributions of beliefs associated with the keywords God and Abortion within the same PCA space, as produced by the base S-BERT. (d)-(f) illustrate the distributions of the same beliefs, color-coded by two opposite template phrases: PRO (I agree with ..., red) and CON (I disagree with ..., blue). (g)-(i) present the distributions of these beliefs after fine-tuning the S-BERT model using the framework outlined in the manuscript. (j)-(l) display the belief distributions with templates represented in distinct colors. Before fine-tuning, the PCA space reveals clearly separated clusters of beliefs with opposing templates, independent of context. After fine-tuning, the belief space becomes more mixed, with opposing templates overlapping. While the overall belief space exhibits a bell-shaped unimodal distribution, beliefs tied to controversial keywords display distinct bimodal distributions.}
    \label{fig:beliefspace_pca_diff}
\end{figure*}

Before fine-tuning, the belief space (Fig.~\ref{fig:beliefspace_pca_diff}a-f) exhibits a bimodal distribution, primarily driven by template phrases (``I agree with ...'' vs. ``I disagree with ...''). While this clustering may suggest a separation of beliefs, it does not reflect the underlying contextual relationships. Instead, the base S-BERT model (prior to fine-tuning) primarily differentiates beliefs based on surface-level template phrasing---PRO/CON positions relative to debate titles---rather than their intrinsic meaning.

In contrast, the fine-tuned S-BERT model (Fig.~\ref{fig:beliefspace_pca_diff}g-l) undergoes a significant transformation in belief representation. Fine-tuning, guided by user co-voting behavior as a proxy for contextual relationships, restructures the belief space to group beliefs based on their actual contextual alignment. This shift is evident in the overlapping clusters of beliefs with opposing template phrases, reflecting the model's enhanced ability to encode deeper semantic relationships beyond superficial phrasing differences.

Notably, while the overall belief space transitions to a unimodal distribution, beliefs associated with controversial topics (e.g., `God' and `Abortion') retain their bimodal structure. However, unlike in the base model, these bimodal clusters now consist of a mix of both template types, representing genuine contextual divergences rather than template-driven separations. This highlights the fine-tuned S-BERT model's ability to disentangle contextual relationships from surface-level phrasing, providing a more accurate and nuanced representation of belief dynamics.

\subsection{Distribution of explained variances of PCA space before and after fine-tuning}

\begin{figure*}[t!]
\centering
\includegraphics[width=\textwidth]{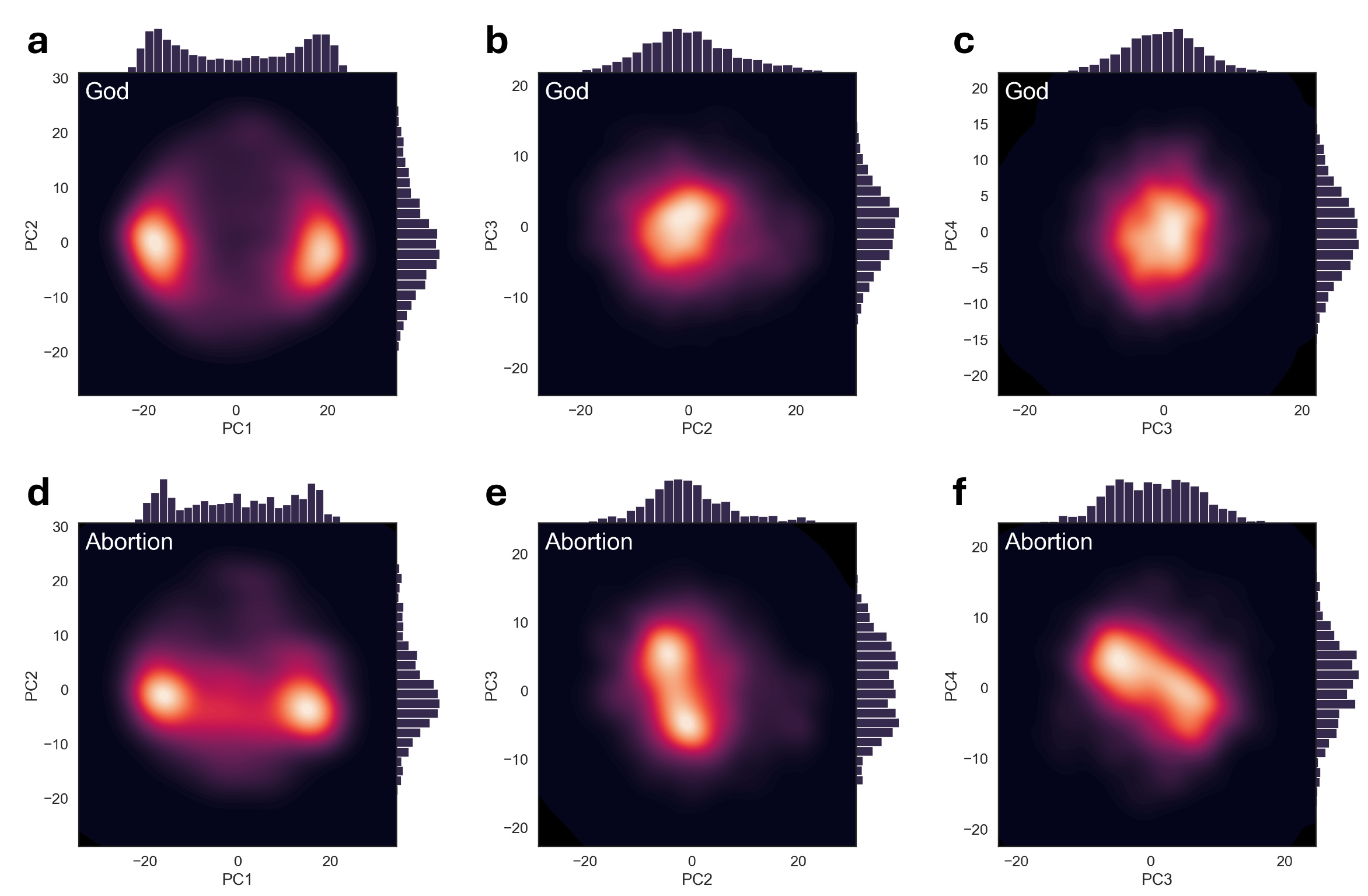}
	\caption{(a) to (c) and (d) to (f) depict the distribution of beliefs related to the keywords God and Abortion across different principal component (PC) spaces (PC1-PC2, PC2-PC3, and PC3-PC4). While the beliefs associated with these keywords exhibit a similar bimodal distribution along the PC1-PC2 axis, they form a single cluster with slightly varying shapes in the higher-dimensional axes. This suggests that the first principal component primarily captures the contrast between two opposing beliefs related to these keywords.}
	\label{fig:keywords_higher_PC}
\end{figure*}

To examine structural changes in the belief space after fine-tuning, PCA was performed, and the variance explained by each principal component was analyzed. For instance, Fig.~\ref{fig:keywords_higher_PC} depicts the distribution of beliefs related to the keywords `God' and `Abortion' across different principal component (PC) spaces (PC1-PC2, PC2-PC3, and PC3-PC4). By measuring the distribution of variance along the PCA axes, we can infer which axes predominantly encode the information in the belief space. Before fine-tuning, the S-BERT model is expected to exhibit a relatively broad variance distribution, as beliefs are not yet organized based on contextual relationships. After fine-tuning, however, beliefs that are contextually similar, despite being textually different, are expected to be represented closer together in the space. Consequently, the belief space should become more compressed, with a larger proportion of variance explained by fewer principal components.

\begin{figure*}[t!]
    \centering
	\includegraphics[width=0.95\textwidth]{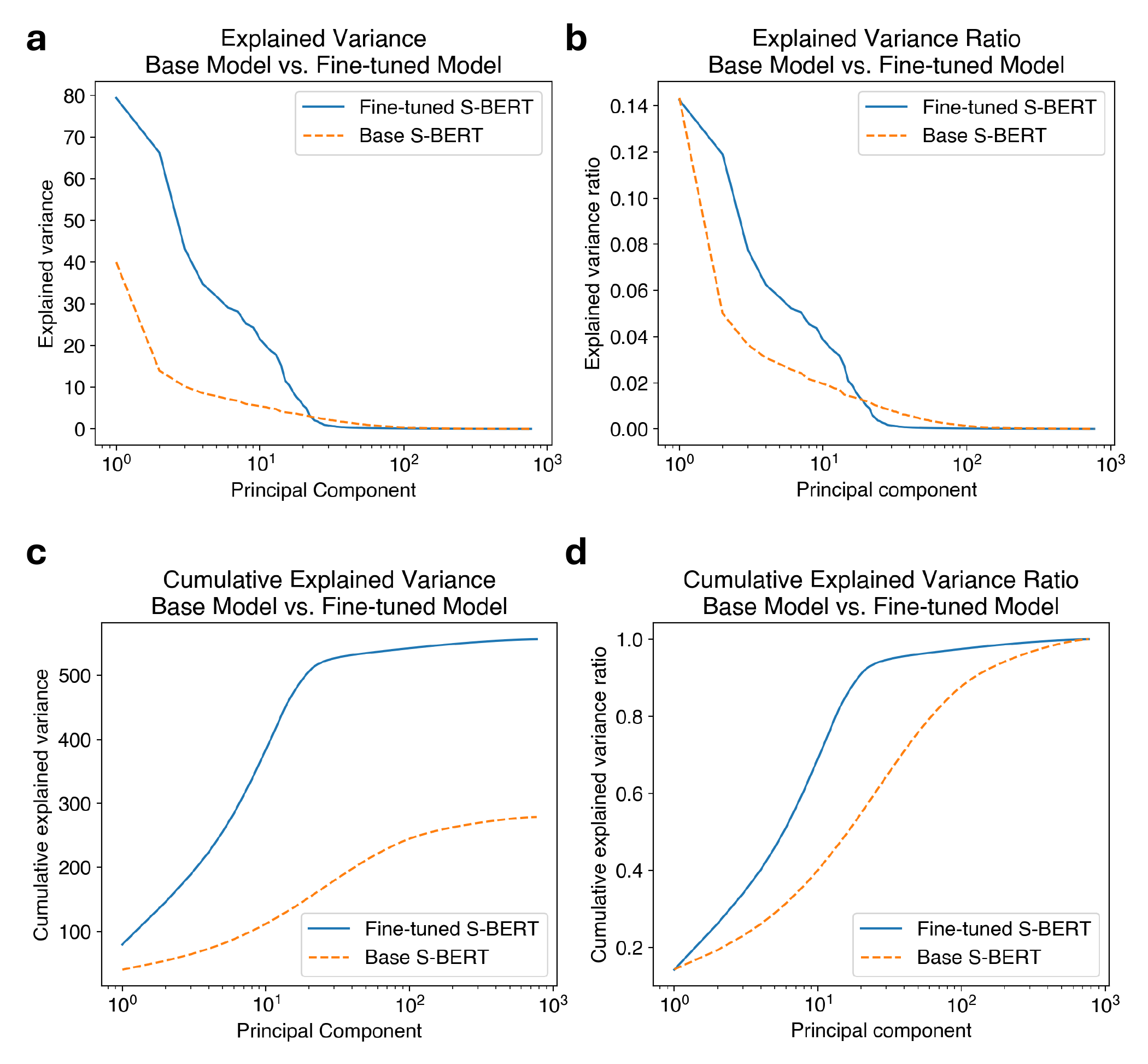}
	\caption{(a) and (b) present the variance and variance ratio explained by the principal components for the base S-BERT model and the fine-tuned S-BERT model. (c) and (d) show the cumulative explained variance across the principal component axes. For the fine-tuned model, a larger proportion of the variance is concentrated in a smaller number of principal components compared to the base model, indicating that the primary principal component axes of the fine-tuned model capture a greater amount of information about the relationships between various beliefs.}
    \label{fig:pca_variance}
\end{figure*}

Figure~\ref{fig:pca_variance} illustrates the explained variances of belief vectors across the principal components for both the base S-BERT model and the fine-tuned S-BERT models. Figure~\ref{fig:pca_variance}a and b show that, in the fine-tuned model, a greater proportion of variance is concentrated in fewer principal components compared to the base model. This supports the hypothesis that the fine-tuned model effectively captures contextual relationships between beliefs, compressing redundant information from higher principal component axes. For instance, first 10 principal components of the fine-tuned model capture 68.9\% of the variance while the base model captures only 40.1\% of the variance.

\subsection{Belief embedding in UMAP} 

In addition to principal component analysis (PCA) of the belief space, we employed Uniform Manifold Approximation and Projection (UMAP)~\cite{mcinnes2018umap}, a dimensionality reduction technique based on manifold learning and topological data analysis, to explore how our model transforms the structure of belief space. UMAP is well-suited for visualizing high-dimensional belief space as it effectively reduces the complexity of the high dimensional belief space while retaining significant relational information among beliefs. 

\begin{figure*}[t!]
    \centering
	\includegraphics[width=0.95\textwidth]{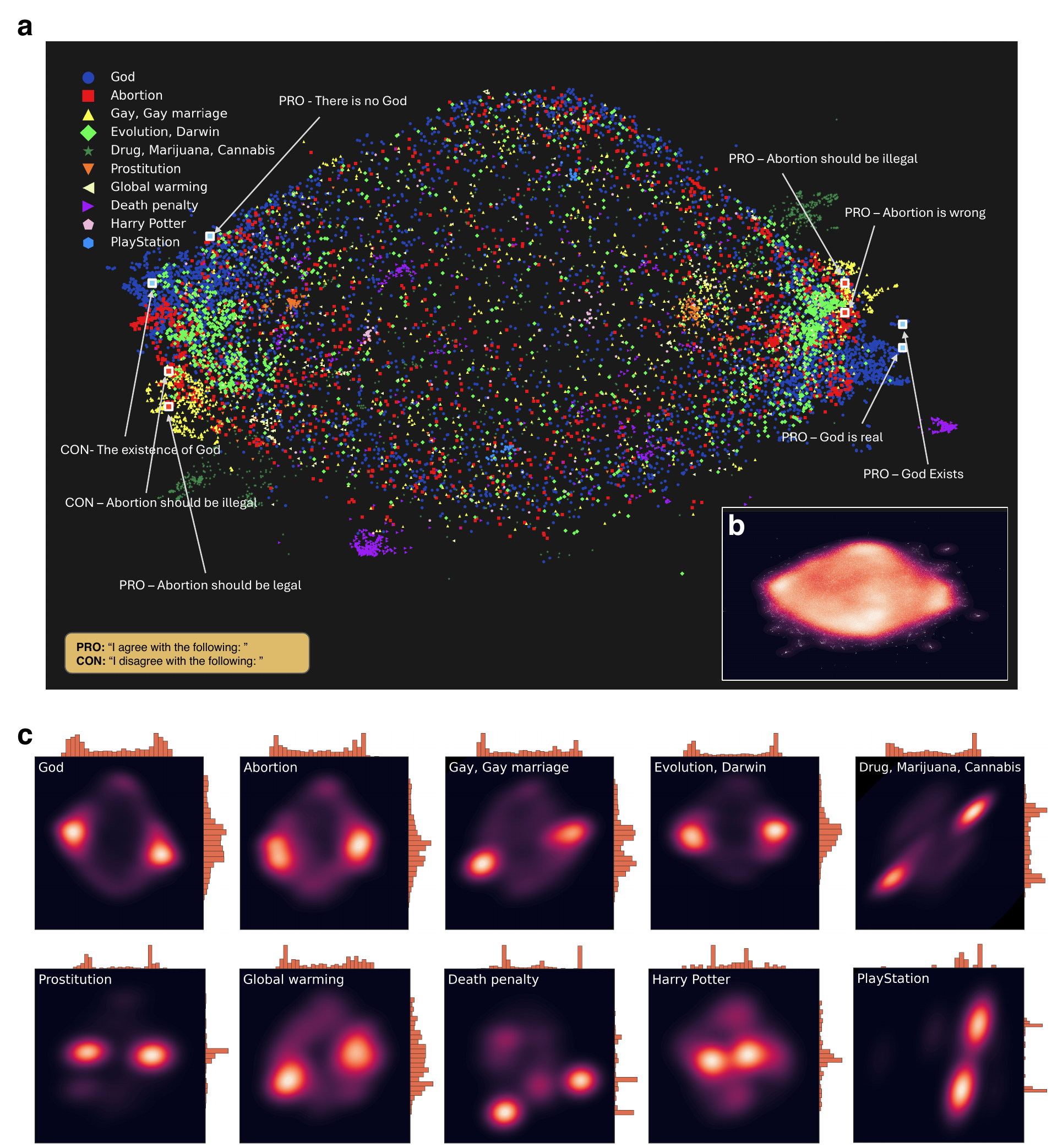}
	\caption{The belief embeddings are projected into a 2-dimensional space using UMAP method. (a) Beliefs associated with 10 selected keywords are illustrated in different colors. The UMAP representation reveals that beliefs related to these keywords tend to form two meaningful clusters in the belief space. Representative examples from different belief clusters are depicted with accompanying text. (b) shows the UMAP representation of entire beliefs in the DDO dataset. The background heatmap, generated using a kernel density estimation method, indicates the density of beliefs, while the overlaid white dots represent individual beliefs. (c) demonstrates distributions of beliefs related to individual keywords in the UMAP space. Each belief distribution associated with different keywords reveals two highly clustered regions, spaced to varying degrees in different areas of the UMAP space. These patterns highlight the existence of polarization in beliefs related to various social issues.}
    \label{fig:beliefspace-umap}
\end{figure*}

To illustrate how beliefs on various topics are distributed in the belief space, we sampled ten representative belief groups, each consisting of a unique set of keywords, such as `God' and `Abortion.' Figure~\ref{fig:beliefspace-umap}a presents the clustering of beliefs related to distinct keywords in the UMAP space of the belief embedding space. 

While the entire set of beliefs (Fig.~\ref{fig:beliefspace-umap}b) exhibit an evenly spread distribution across a broad region, the density plots for beliefs related to specific keywords reveal polarized distributions with two distinct clusters (Fig.~\ref{fig:beliefspace-umap}c). These patterns suggest that beliefs on these topics naturally form two opposing clusters, reflecting the existing political polarization on various social issues.

Additionally, the proximity of belief clusters associated with different keywords reveals interconnections between belief groups. For instance, beliefs about existence of God and opposition to abortion predominantly appear on the far right side of the UMAP space, whereas beliefs rejecting the existence of God/god and supporting abortion rights tend to cluster on the far left (Fig.~\ref{fig:beliefspace-umap}a). This indicates beliefs that supports God/god are more closely related with those beliefs that are against abortion. Similarly, beliefs related to `Gay and gay marriage,' `Prostitution,' `Global warming,' and `Evolution and Darwin' exhibit two dense, distant clusters in the belief space (Fig.~\ref{fig:beliefspace-umap}c).

\subsection{User embedding of the base S-BERT model}

We derived user embeddings in the belief space by averaging the belief vectors corresponding to all beliefs expressed by each individual user. This approach allows users to be represented within the same belief space as their beliefs. To evaluate whether these user representations capture patterns of polarization, we mapped users based on their self-reported survey responses, as shown in Fig.~3 of the main manuscript. For this, we leveraged the pre-survey information of users on major social issues, included in the DDO dataset, separate from their debate participation. This survey contains information on users’ political party affiliations, religious ideologies, and positions on 48 key social issues, referred to as `Big issues', such as `Abortion,' `Drug legalization,' `Gun control,' and others.

\begin{figure*}[t!]
\centering
\includegraphics[width=\textwidth]{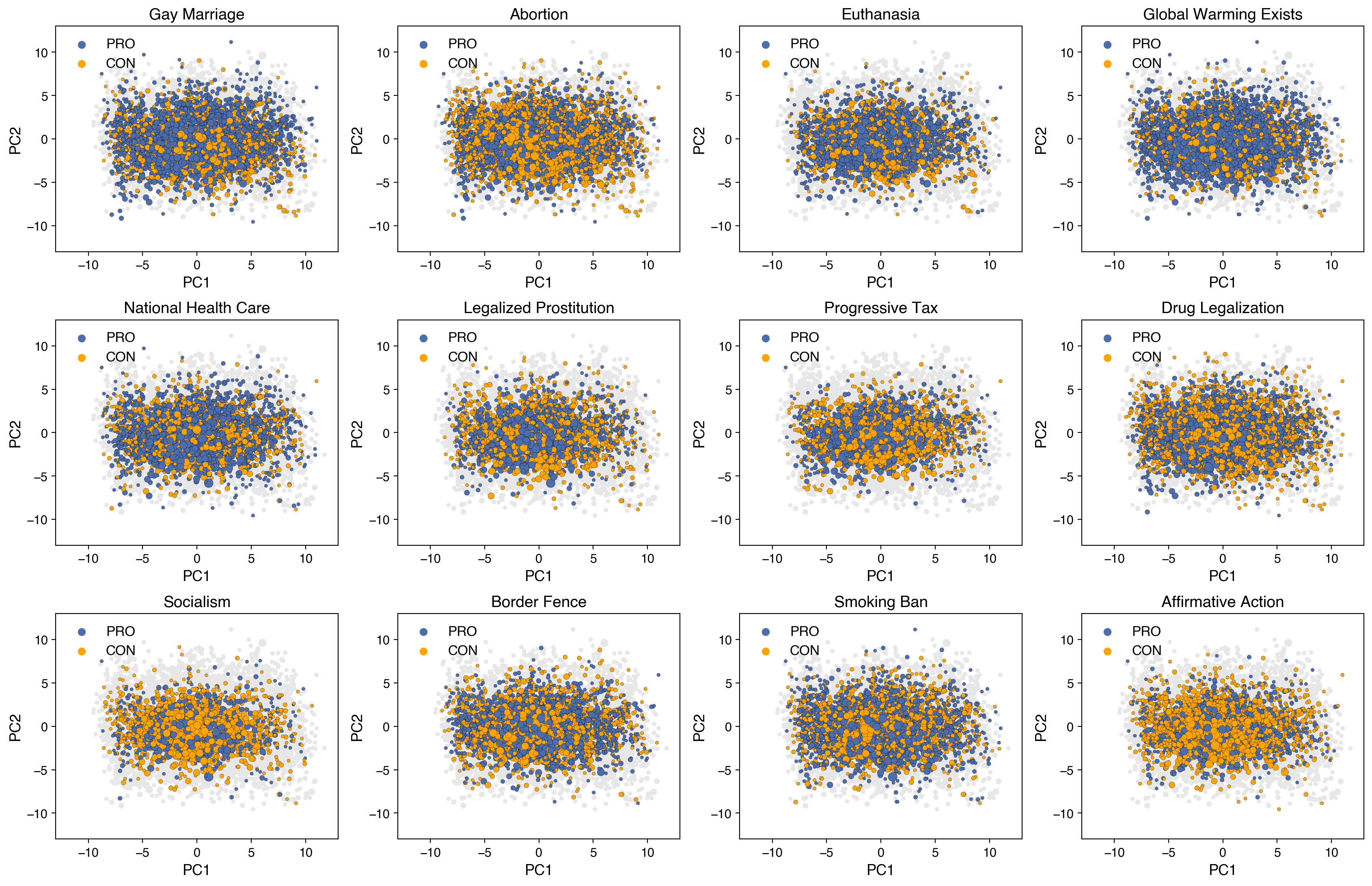}
	\caption{Visualization of user embeddings in the belief space generated by the base S-BERT model (pre-fine-tuning). Different user groups associated with various controversial social issues are represented in distinct colors. Unlike the fine-tuned model, which demonstrates a clearer separation between user groups in the belief space (as shown in Fig.~3 of the main manuscript), the user embeddings generated by the base S-BERT model do not exhibit such separation.}
	\label{fig:userembedding_non_ft}
\end{figure*}

Figure~\ref{fig:userembedding_non_ft} visualizes user embeddings in the belief space generated by the base S-BERT model (prior to fine-tuning). Different user groups associated with various controversial social issues are depicted in distinct colors. In contrast to the fine-tuned model, which demonstrates a clearer separation of user groups in the belief space, the embeddings produced by the base S-BERT model lack such separation. This supports that our fine-tuning method effectively clusters related beliefs and groups users with similar beliefs closer together in the embedding space.

\subsection{Distance in the belief space between user groups}

\begin{figure*}
\centering
\includegraphics[width=\textwidth]{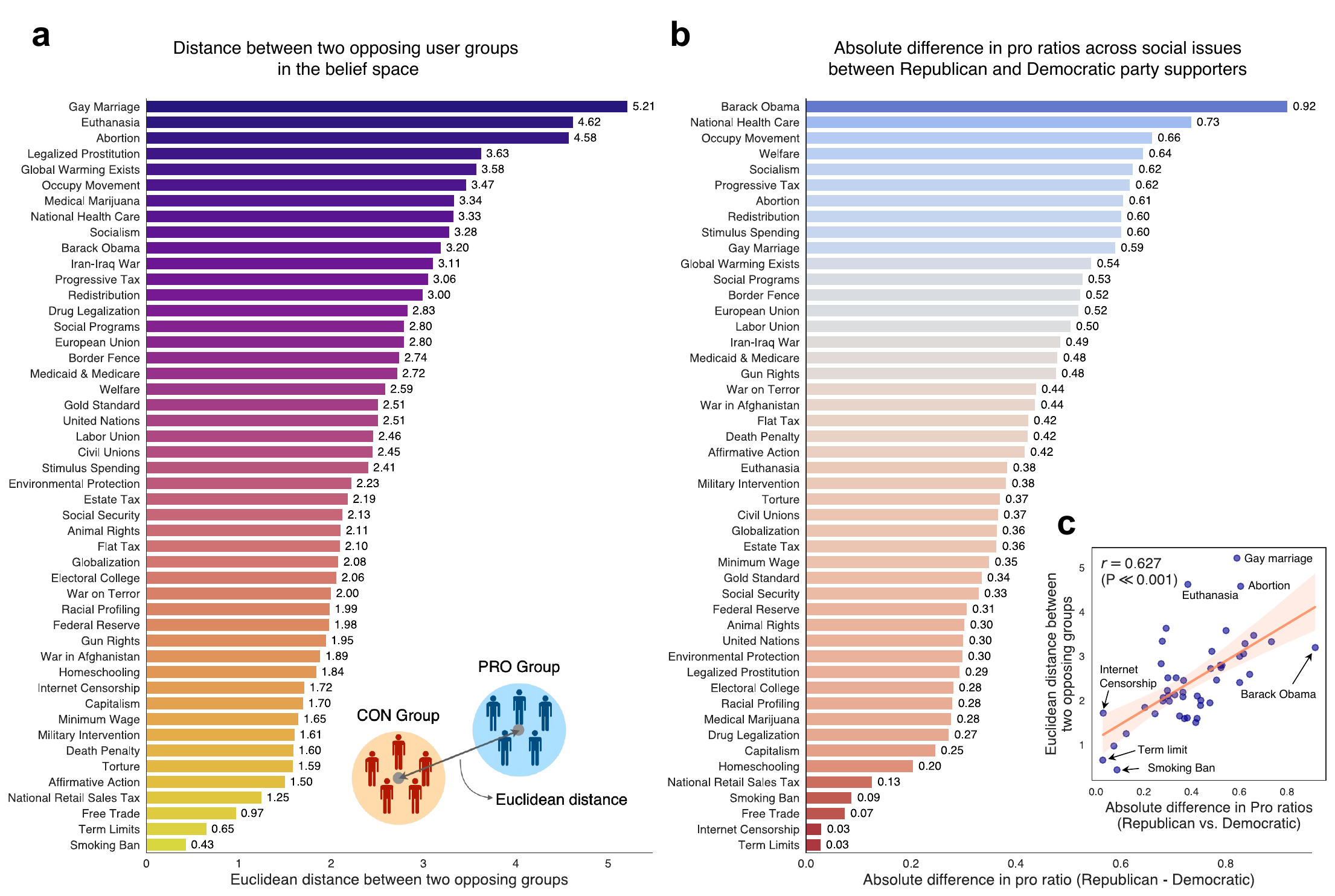}
	\caption{(a) Euclidean distance in the belief space between user groups with contrasting stances on 48 major social issues referred to as `Big issues,' based on the self-reported survey results. The distance between two opposing user groups is calculated by measuring the distance between the average positions of CON users and PRO users for each issue. This distance between two user groups on an issue demonstrates the degree of separation between the groups, highlighting the extent of polarization between them. (b) The absolute difference in pro ratios across social issues between supporters of the Republican and Democratic parties reveals the degree of partisan polarization in social issues. The pro ratios are calculated using self-reported data from users, independent of the belief embedding. (c) The Euclidean distance between two opposing groups in the belief space and the degree of partisan polarization across social issues exhibit a significantly high correlation ($r=0.627$).}
	\label{fig:big_issue_distance}
\end{figure*}

\begin{figure*}
\centering
\includegraphics[width=0.7\textwidth]{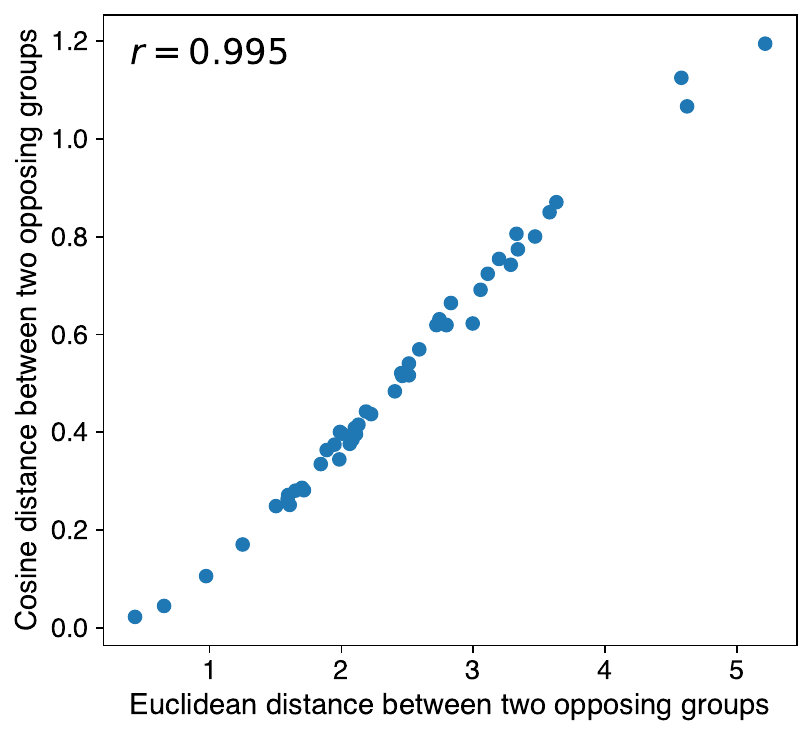}
	\caption{Comparison of Euclidean and cosine distances between two user groups with opposing opinions on 48 `Big issues'. The two distance metrics are highly correlated, with Pearson correlation $r=0.995$ ($p<0.001$).}
	\label{fig:pol_cos_dist}
\end{figure*}

Based on distinct clustering patterns of opposing user groups on various issues, we examined what the most polarizing topics were in the belief space. We quantified user polarization on specific issues by measuring the Euclidean distance between the centroids of PRO and CON user groups on the 48 `Big issues', in the original high-dimensional belief space. Fig.~\ref{fig:big_issue_distance}a displays social issues ranked according to the Euclidean distances between two opposing user groups. Users with differing views on `Gay marriage' showed the greatest separation distance, indicating significant polarization. Other highly polarized topics include `Abortion,' `Euthanasia,' and `Global warming.' Conversely, users on issues such as  `Smoking ban,' `Term limits,' and `Free trade' exhibit relatively weaker polarization in the belief space. We note that the cosine distances between user groups were also highly correlated with the Euclidean distances (Pearson's $r=0.995$, $P < 0.001$) (Fig.~\ref{fig:pol_cos_dist}).

To examine if the Euclidean distance between two groups in belief space actually reflects the degree of conflict over significant issues, we compared these distances with the degree of partisan polarization based on the self-reported user survey data. We measured the absolute differences in pro-ratios across the 48 Big issues between supporters of the Republican and Democratic parties (Fig.~\ref{fig:big_issue_distance}b). The pro-ratio for a social issue of a particular party is determined by the percentage of users who support the issue within the party based on their self-reported data. Fig.~\ref{fig:big_issue_distance}c shows that the Euclidean distance in the belief space and the degree of partisan polarization across the big issues are highly correlated ($r=0.627$, $P\ll0.001$), indicating that the distance between different user groups in the embedding space can be a useful metric for characterizing conflict levels over various social issues.

\section{Detailed results for the belief prediction task}

In the downstream task, we conducted a binary belief classification aimed at predicting an individual's voting position (PRO or CON) in new debates. We divided the entire set of debates into an 8:2 ratio and evaluated the model's performance using 5-fold cross-validation. We considered users who appeared at least once in both the train and test sets. Table~\ref{tab:downstream_statistics} summarizes the basic statistics of the train and test datasets.

\subsection{Average accuracy for belief prediction over user history}
\begin{figure*}
\centering
\includegraphics[width=0.9\textwidth]{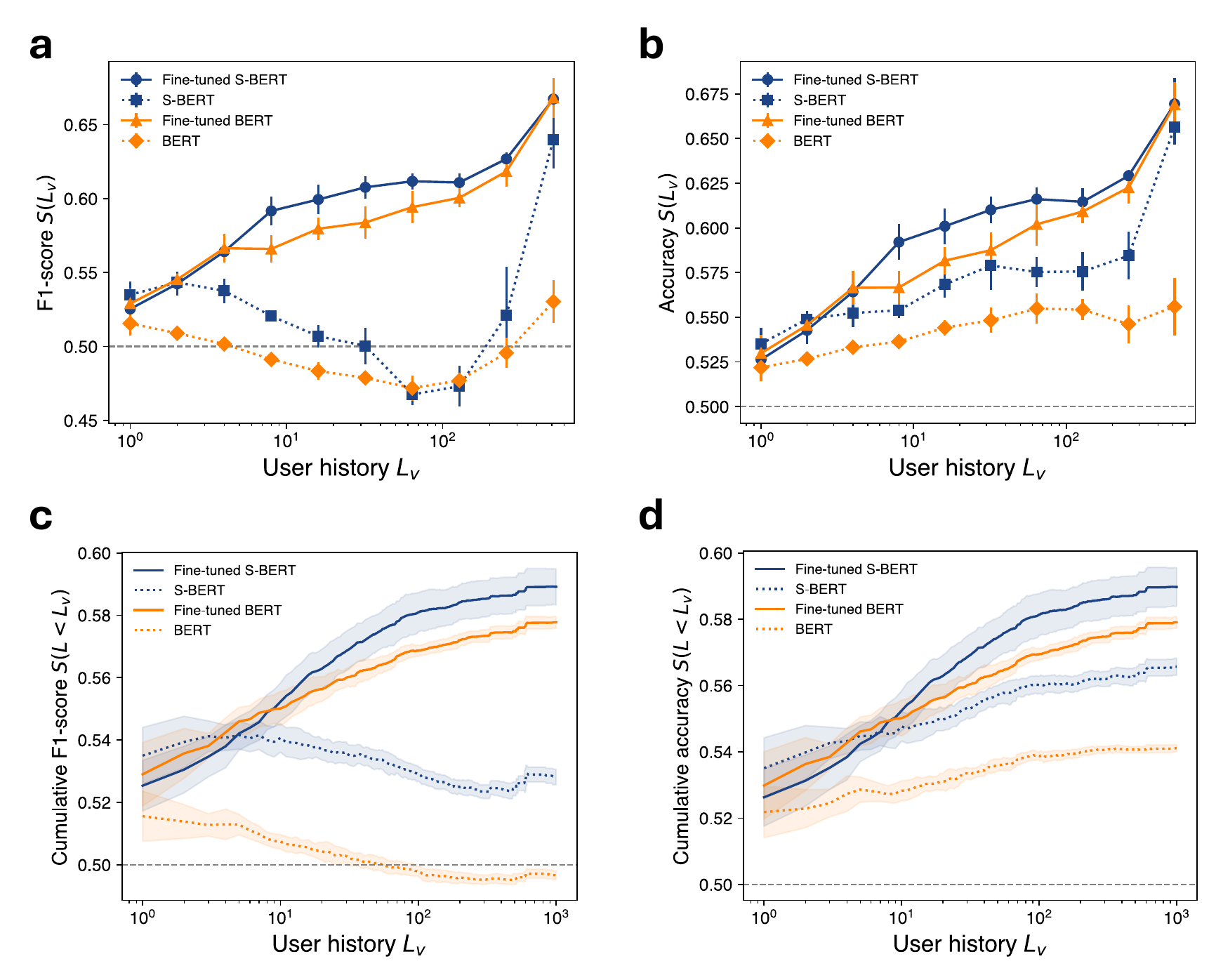}
	\caption{Performance of belief prediction downstream task over the length of users' voting history $L_{v}$. (a) and (b) show the relation between F1-score and accuracy with $L_v$. (c) and (d) show the relationship between cumulative F1-score (accuracy) $S(L < L_v)$ and $L_v$, where $S(L < L_v)$ represents the F1-score (accuracy) for users with voting records shorter than $L_v$. User beliefs are more accurately predicted as more voting records are accumulated.}
	\label{fig:length_effect}
\end{figure*}

Figure~\ref{fig:length_effect} shows the performance of the belief prediction downstream task as a function of the length of user voting history, $L_{v}$. As the users' voting records increase, their beliefs tend to be more accurately predicted. Fig.~\ref{fig:length_effect}a and b show the relation between F1-score and accuracy with $L_{v}$.
Fig.~\ref{fig:length_effect}c and d show the relationship between cumulative F1-score (accuracy) $S(L<L_{v})$ and $L_{v}$, where $S(L<L_{v})$ represents the F1-score and accuracy for users with voting
records shorter than $L_v$. There results suggests that user beliefs are more accurately predicted as more voting records are accumulated.

\subsection{Average accuracy for belief prediction over $d_{\text{avg}}$}
\begin{figure*}
\centering
\includegraphics[width=0.6\textwidth]{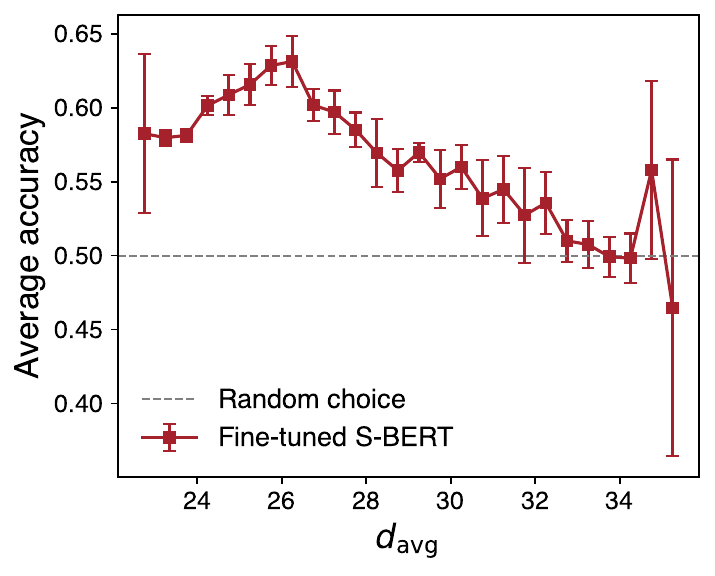}
	\caption{Average accuracy for belief prediction task over the average distances $d_{\text{avg}}$ of two contrasting beliefs from users.}
	\label{fig:acc_vs_davg}
\end{figure*}

Figures.~4f-h in the main manuscript show the relationship between average accuracy and $d_{\text{max}}$ and $d_\text{min}$. We further measured how belief prediction accuracy changes depending on the average distance between the user and two conflicting beliefs for a given debate, $d_\text{avg}$. Fig.~\ref{fig:acc_vs_davg} shows how average accuracy for belief prediction changes with $d_\text{avg}$. As $d_\text{avg}$ increases, the accuracy tends to level off at 0.5, akin to random choice baseline.

\subsection{Effect of user attributes on belief prediction accuracy}

In this belief prediction task, we examined whether the beliefs of certain user groups--categorized by political party, religion, and gender--were predicted with higher or lower accuracy. Specifically, we compared prediction accuracy across user groups defined by three distinct attributes:

\begin{enumerate} 
\item Political party (Democratic vs. Republican) 
\item Gender (Male vs. Female) 
\item Religious ideology (Christian vs. Atheist) 
\end{enumerate}

\begin{figure*}
\centering
\includegraphics[width=0.9\textwidth]{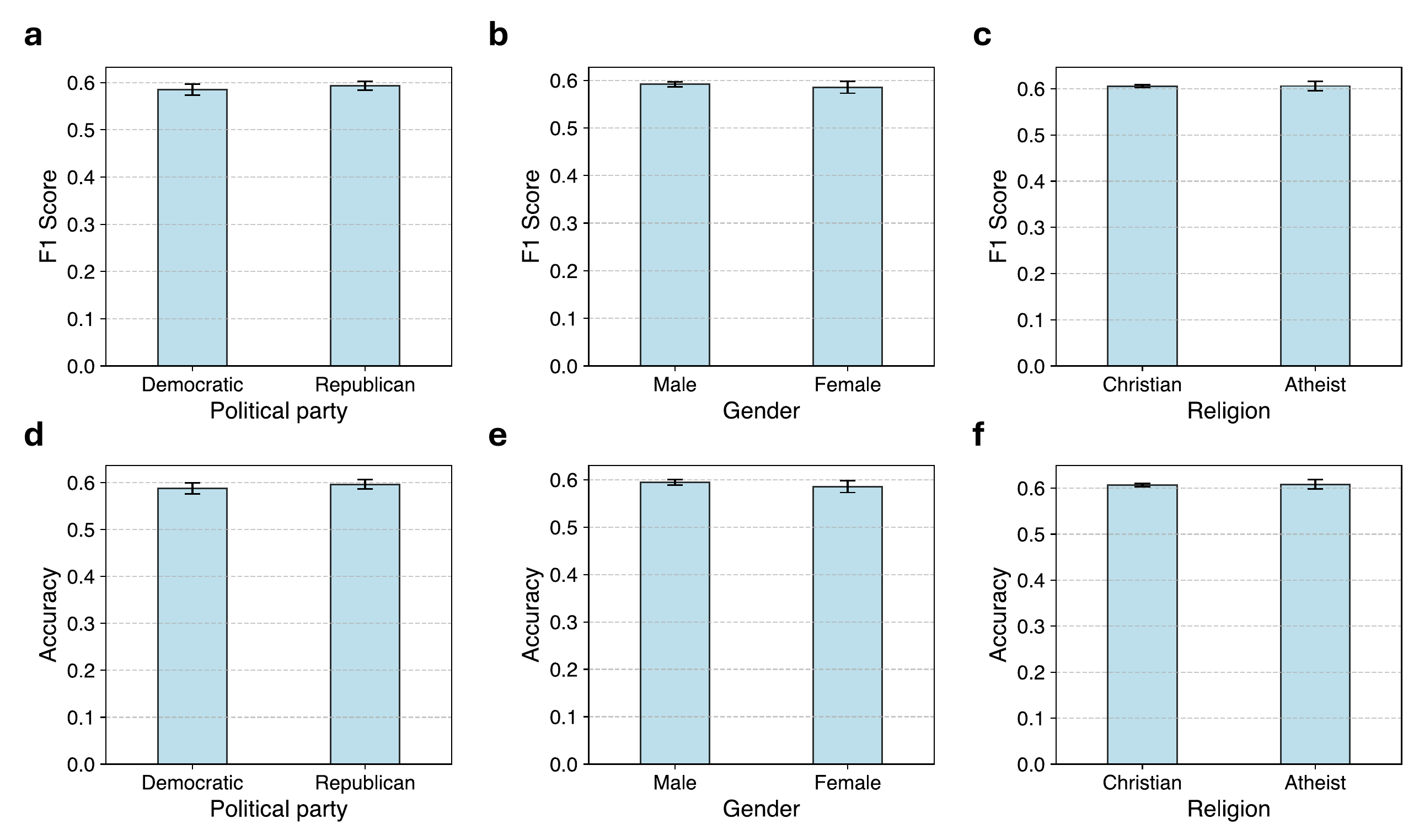}
	\caption{Comparison of prediction accuracy (macro F1-score and accuracy) in the belief prediction task across different user groups based on political party (Democratic vs. Republican), gender (Male vs. Female), and religion (Christian vs. Atheist). (a)-(c) present macro F1-scores across the user groups, while (d)-(f) show accuracies. Results from independent t-tests indicate that all p-values were above 0.1, demonstrating no statistically significant differences between the groups. The Shapiro-Wilk test confirmed normality for each sample ($p > 0.05$).}
	\label{fig:prediction_other_factors}
\end{figure*}

We measured prediction accuracy and F1-scores for each user group based on these categories. However, our analysis did not reveal any significant differences among the groups (Fig.~\ref{fig:prediction_other_factors}). The results of independent t-tests showed that all p-values were above 0.1, indicating a lack of statistically significant differences.

\section{Robustness of the belief prediction results across various data-splitting and testing scenarios}

In this section, we present a comprehensive robustness check of our results across various data-splitting scenarios. Specifically, we conducted five robustness checks to evaluate the generalizability of the experiment outcomes across various data-splitting and testing scenarios. 

These experiments demonstrated that our findings consistently hold across diverse evaluation tasks, including triplet loss evaluation, belief prediction accuracy, and the relationship between relative dissonance and prediction performance. Below, we summarize the five data-splitting scenarios tested, followed by detailed descriptions and results for each scenario.

\subsection{Summary of data-splitting scenarios}

\begin{enumerate}
    \item `user\_downsample' scenario: Mitigates the influence of highly active users by downsampling participation records to 100 debates for users with more than 100 debates.
    \item `category\_downsample' scenario: Balances the representation of debate topics by downsampling frequent categories to a maximum of 5,000 debates per category.
    \item `without\_politics' scenario: Excludes all political debates to test the model’s robustness without the influence of polarized political beliefs.
    \item `user\_split' scenario: Ensures no overlap between users in the training and test sets by completely excluding test set users from the training set. 
    \item `temporal\_division' scenario: Splits debates chronologically to assess the model’s ability to predict beliefs in temporally new debates.
\end{enumerate}

\subsection{Detailed descriptions of each scenario}
\begin{figure*}
    \centering
	\includegraphics[width=0.7\textwidth]{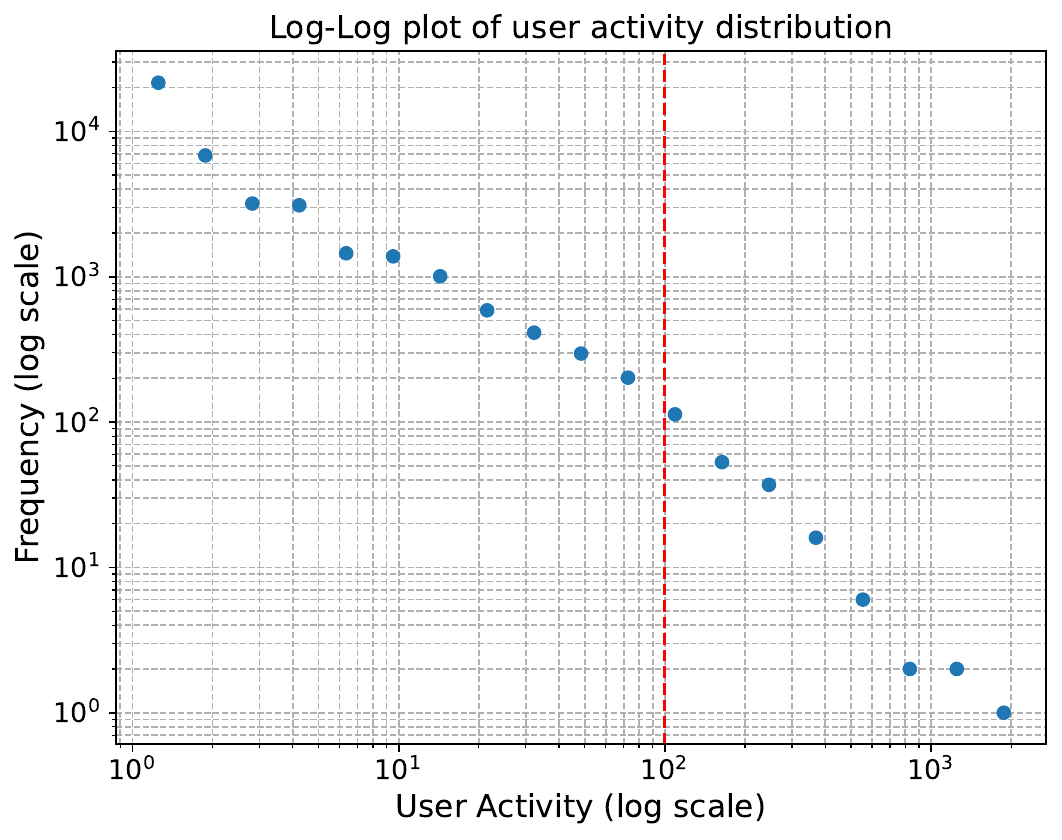}
	\caption{Log-log plot showing the distribution of user activity. The dashed red line marks a user activity level of 100.}
    \label{fig:user_downsampling}
\end{figure*}

\begin{figure*}
    \centering
	\includegraphics[width=0.8\textwidth]{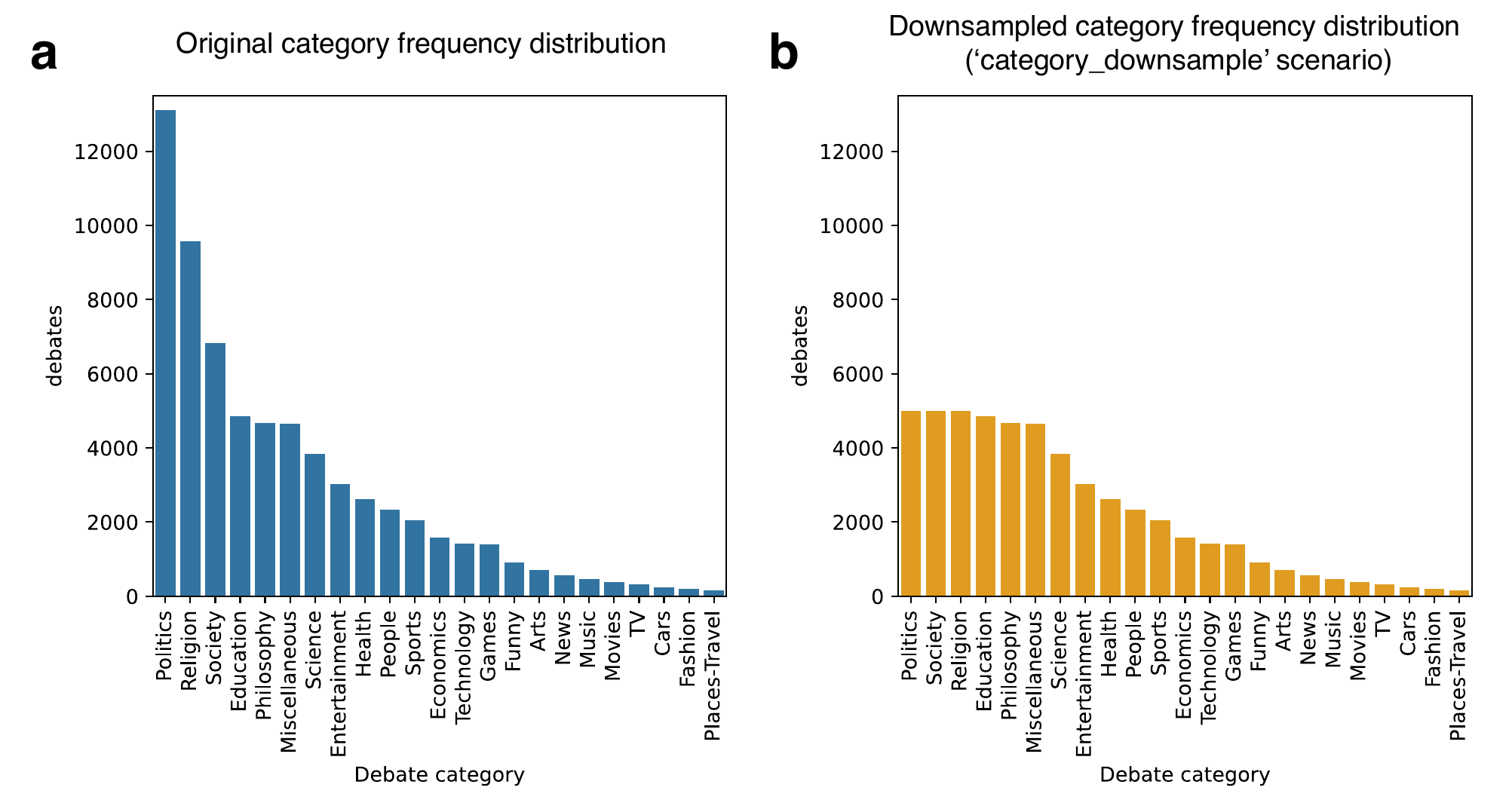}
	\caption{Number of debates across different debate categories: (a) Original distribution of debates. (b) Distribution after downsampling each debate category to a maximum of 5,000 debates.}
    \label{fig:category_downsampling}
\end{figure*}

\begin{figure*}
    \centering
	\includegraphics[width=0.9\textwidth]{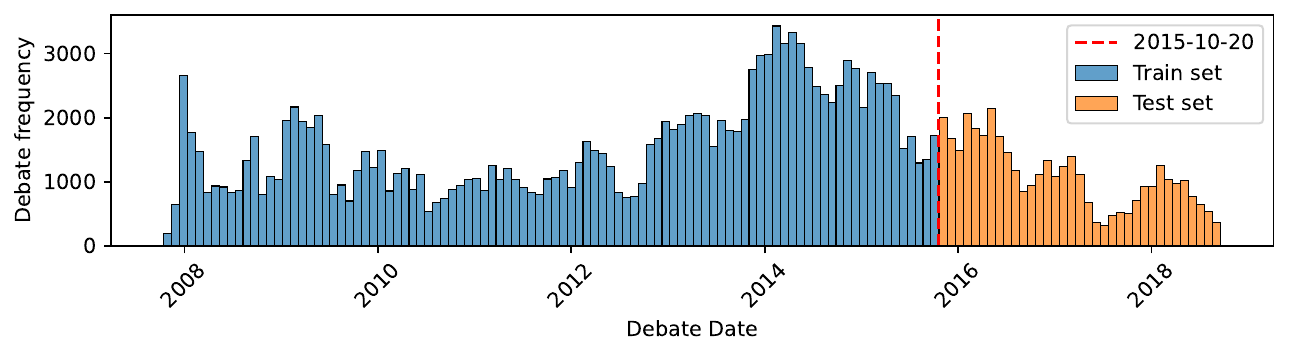}
	\caption{In `temporal\_division' data-splitting scenario, debates were split chronologically into training and test sets, with the training set containing 80\% of earlier debates and the test set containing the remaining 20\%}
    \label{fig:temporal_division}
\end{figure*}

\begin{enumerate}
\item \textbf{Downsampling of active users (`user\_downsample')}: To reduce the disproportionate influence of highly active users, participation records were randomly downsampled to a maximum of 100 debates per user (Fig.~\ref{fig:user_downsampling}). This adjustment balances user activity levels while preserving data structure, enabling the model to be trained on more balanced user data.

\item \textbf{Downsampling of frequent debate categories (`category\_downsample')}: Frequent debate categories often dominate data representation. To mitigate this bias, debates within each category were downsampled to a maximum of 5,000. This adjustment improved the model’s ability to generalize across less dominant categories (Fig.~\ref{fig:category_downsampling}).

\item \textbf{Removal of political debates (`without\_politics')}: Debates labeled under ‘Politics’ category were excluded to test the robustness of the model without the influence of polarized US-centric socio-political beliefs. This scenario enables testing of the model's robustness in the absence of political content and helps assess the generalizability of our findings to non-political contexts.

\item \textbf{Data split by users (`user\_split')}: 
This scenario excludes test set users entirely from the training set, ensuring the evaluation of model performance on unseen users. While this strict separation reduces the number of test users (averaging 1,392 compared to 10,000 in debate-level splitting), it provides a robust measure of generalizability. Test users must satisfy two criteria: (1) have participation records in training debates and (2) engage in new debates outside the training set.

\item \textbf{Temporal division (`temporal\_division')}: 
Debates were split chronologically, with 80\% of earlier debates in the training set and 20\% of later debates in the test set (using October 20, 2015, as the dividing point, as shown in Fig~\ref{fig:temporal_division}). This setup allows us to evaluate the model's ability to predict the beliefs of people on temporally new debates, which include debates on newly emerging issues from test dataset's time frame. 
\end{enumerate}

Models fine-tuned in 5 different scenarios showed consistent results across all tests, strengthening the generalizability and reliability of the suggested belief embedding framework. Detailed comparisons of experimental results are provided below.

\subsection{Triplet evaluation scores and sentence similarity benchmark scores}
We first assessed triplet evaluation scores and semantic similarity scores to evaluate the quality of belief embedding models across various training scenarios (Table~\ref{tab:robustness_triplet_evaluation}). 
Triplet evaluation measures the models' ability to correctly classify the belief pair relationships as either positive or negative. Table~\ref{tab:robustness_triplet_evaluation} demonstrates that the triplet evaluation results across different data-splitting scenarios show similar performance, confirming the robustness of the models. Similarly, models from different scenarios achieved similar Spearman correlation scores above 0.7 on the semantic textual similarity benchmark (GLUE-STSB)~\cite{wang2019glue}.

\subsection{Belief prediction scores}
We also compared the performance of the belief prediction task across models trained under different data-splitting scenarios. During the belief prediction task, models predict users' voting positions (PRO or CON) on new debates in the test sets based on the distance between a user and two beliefs. Performance was evaluated using macro F1-scores and accuracy metrics. Table~\ref{tab:robustness_performance_metrics} shows that performance metrics (F1-score and accuracy) showed slight variations in the results (e.g., an F1-score of 0.566 in `user\_split' vs. 0.599 in `without\_politics'), but overall prediction scores across scenarios remained consistent (Table~\ref{tab:robustness_performance_metrics}).

\subsection{Factors affecting belief prediction results}
To assess the robustness of various factors affecting the belief prediction accuracy, we conducted the four experiments focusing on: (a) the effect of user voting history length, (b) the effect of debate categories, (c) the effect of relative dissonance (and other distance measures in the belief space), and (d) the effect of effective radius on prediction accuracy.

\subsubsection{User voting history and prediction accuracy}
\begin{figure*}
    \centering
	\includegraphics[width=0.8\textwidth]{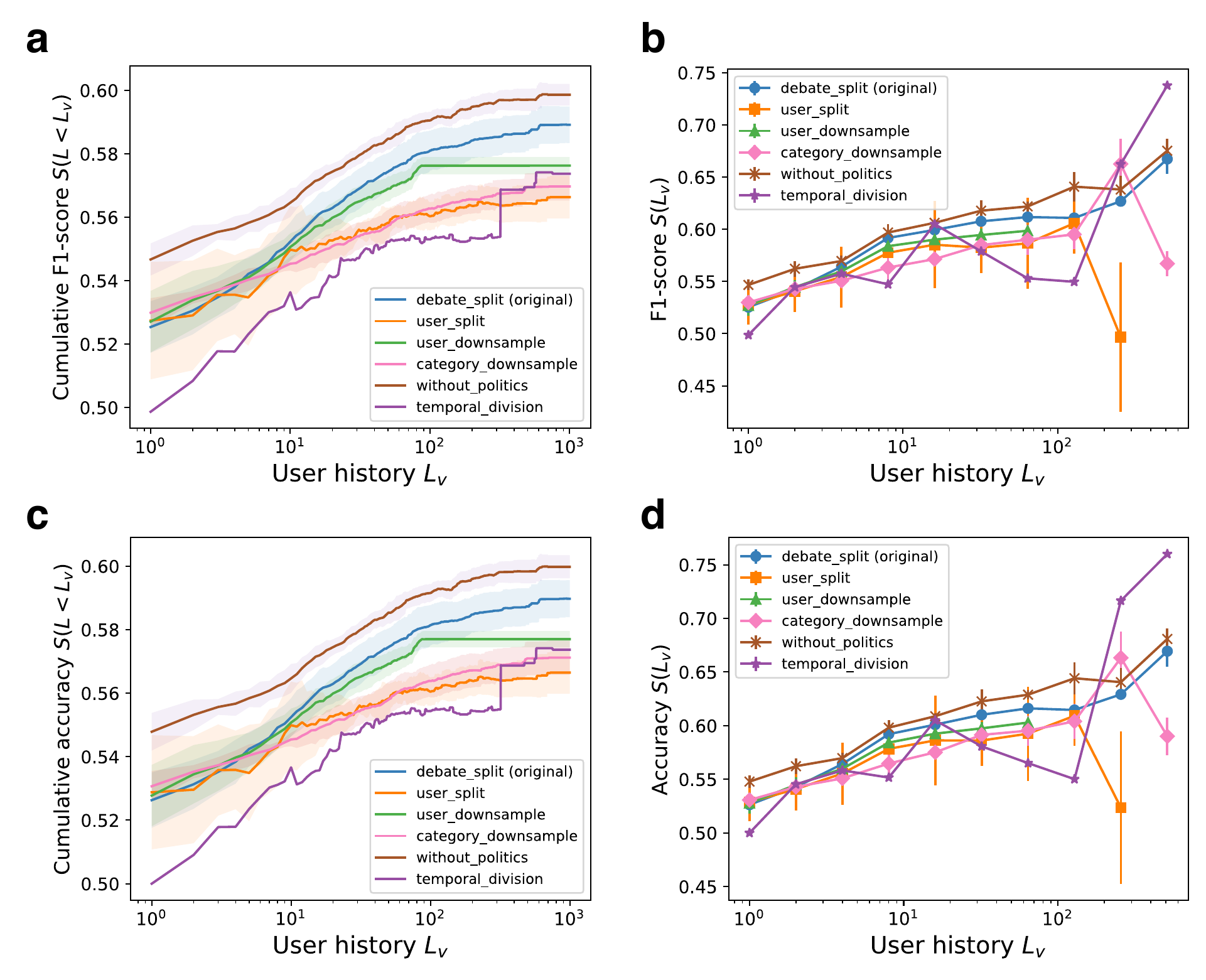}
	\caption{Performance of the belief prediction task based on the length of users' voting history ($L_{v}$) across different data-splitting scenarios. (a) and (c) illustrate the relationship between cumulative F1-score and accuracy ($S(L < L_v)$), where $S(L < L_v)$ represents the F1-score or accuracy for users with voting histories shorter than $L_v$. (b) and (d) present the F1-score and accuracy as functions of $L_v$. The results suggest that as the length of user voting histories increases, belief predictions become more precise. This pattern is consistently observed across various data-splitting scenarios.} 
    \label{fig:robust_length_effect}
\end{figure*}

We first evaluated the relationship between the length of a user's voting history and the model's accuracy in predicting beliefs on new debates (Fig.~\ref{fig:robust_length_effect}). The results show a clear increasing trend in prediction performance (measured by F1-score and accuracy) as the length of the user's voting records increases, regardless of the data-splitting scenario. This finding suggests that accumulating more user records improves the precision of belief predictions in all scenarios.

\subsubsection{Varying prediction accuracy across debate categories}
\begin{figure*}
    \centering
	\includegraphics[width=0.9\textwidth]{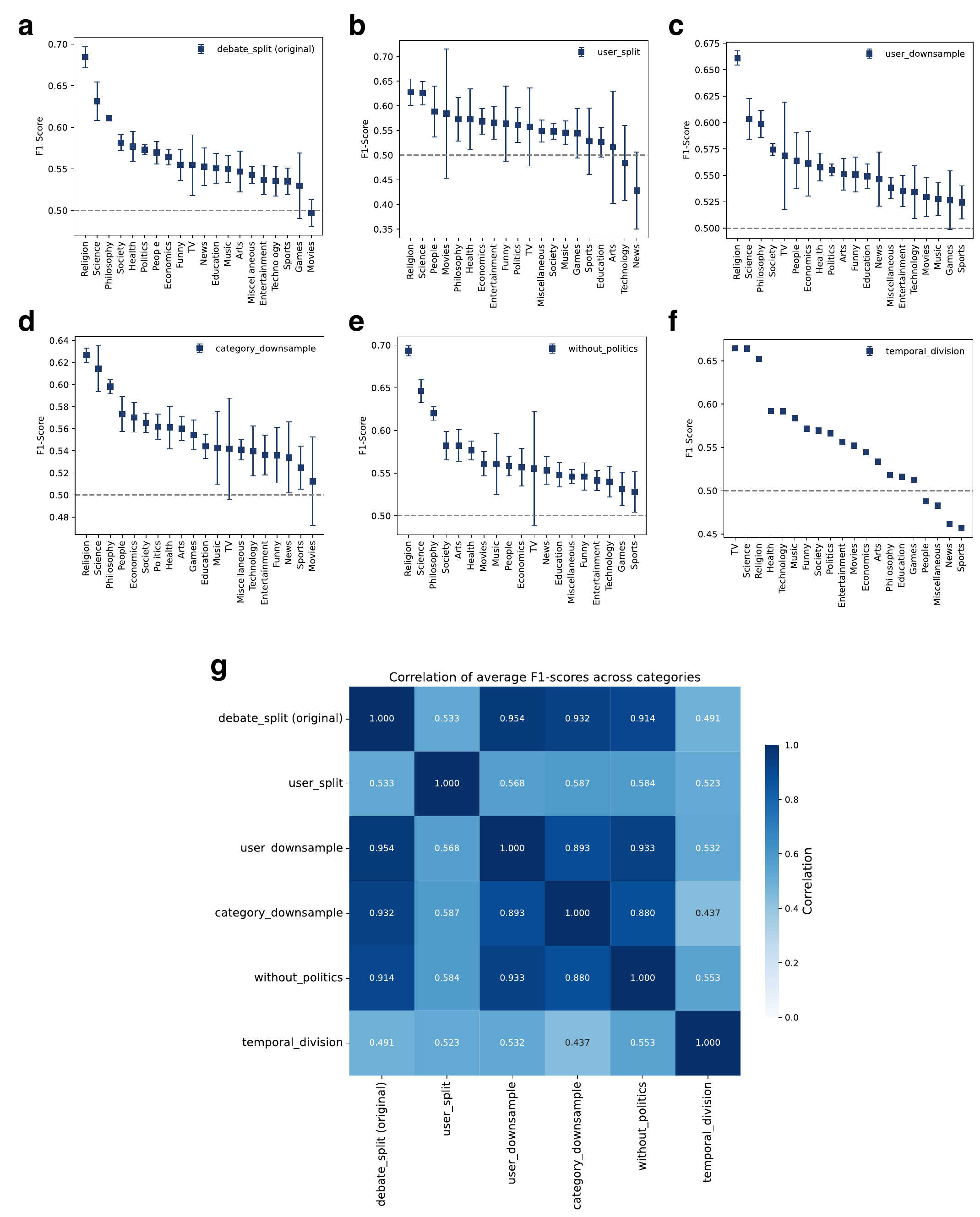}
	\caption{(a)-(f) Belief prediction accuracy across debate categories. The model demonstrated relatively higher accuracy in predicting users' beliefs in debates related to `religion,' `science,' and `philosophy,' compared to debates on topics such as `music,' `games,' and `sports.' This trend was consistently observed across different data-splitting scenarios. (g) Correlation matrix of prediction accuracy. The matrix shows positive correlations between the F1-scores of 20 major categories across different data-splitting scenarios, highlighting the robustness of the results.
}
    \label{fig:robust_category_effect}
\end{figure*}

The original results in the main manuscript demonstrated that belief prediction accuracy varies across debate categories. For example, the model's predictions of users' beliefs in debates related to `religion,' `science,' or `philosophy' were relatively more accurate than those for debates categorized as `music,' `games,' or `sports.' We measured accuracy across different categories, and the results from the five data-splitting scenarios followed a similar trend to the original findings, as shown in Fig.~\ref{fig:robust_category_effect}. Furthermore, Fig.~\ref{fig:robust_category_effect}g illustrates positive correlations between the average F1-scores of 20 major categories across the scenarios.

\subsubsection{Effect of relative dissonance on prediction performance}

A key finding of the belief prediction task is the linear relationship between relative dissonance $d^{*}$ and prediction accuracy, up to the point where accuracy approaches nearly its maximum value of 1. Here $d^{*}$ is defined as the relative difference between the distances from a user to two opposite beliefs $(d_{\text{max}}-d_{\text{min}})/d_{\text{min}}$, where $d_{\text{min}}$ ($d_{\text{max}}$) is the distance between a user and the closer (farther) belief in the belief space. 

\begin{figure*}
    \centering
	\includegraphics[width=0.8\textwidth]{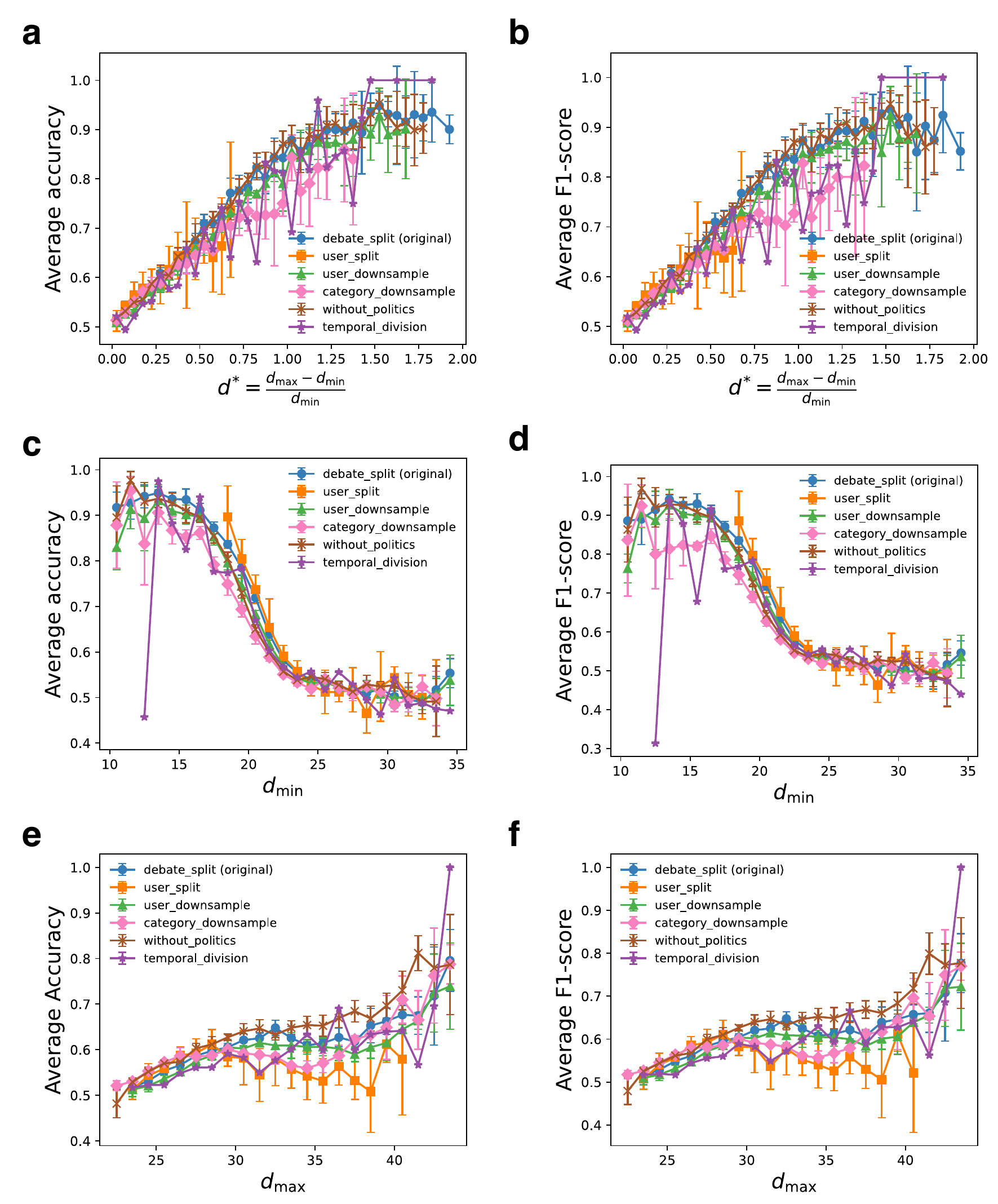}
	\caption{The relationship between belief prediction performance and relative dissonance $d^{*}$ across various data-splitting scenarios. (a)-(b) show average accuracy and macro F1-score as functions of $d^{*}$. (c)-(d) depict these metrics in relation to $d_{\text{min}}$. (e)-(f) illustrate their relationship with $d_{\text{max}}$.}
    \label{fig:robust_distance_effect}
\end{figure*}

This relationship was consistently observed across all data-splitting scenarios, highighting the robustness of the model and the user behavior trends (Fig.~\ref{fig:robust_distance_effect}a and b). Furthermore, the decreasing (and increasing) pattern of accuracy over $d_{\text{min}}$ ($d_{\text{max}}$) was consistently observed, although the degree of noise varied slightly due to differences in data size. 

We note that, in our framework, prediction accuracy can be interpreted as the probability of a user choosing the closer belief between two opposing belief options in a new debate. The observed linear relationship indicates that as a belief becomes relatively closer to a user compared to its opposing belief (i.e., larger $d^{*}$), the likelihood of the user selecting the closer belief increases linearly, approaching a probability of 1.

\subsubsection{Correlation between category size, category mean $d^{*}$, and F1-score in the belief prediction task}

\begin{figure*}
    \centering
	\includegraphics[width=0.9\textwidth]{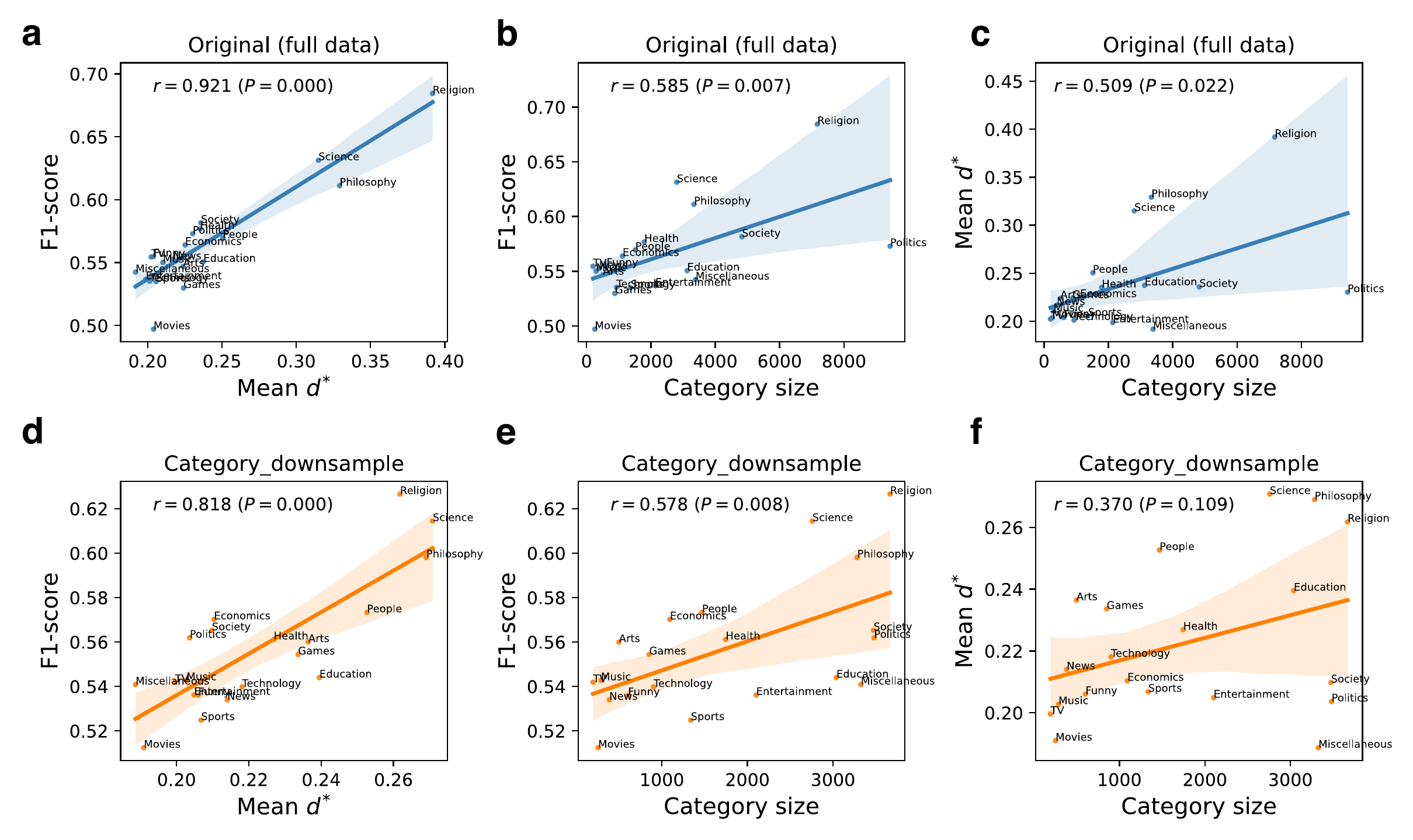}
	\caption{Pairwise correlations between category size, category mean $d^{*}$, and F1-score in the belief prediction task. Scatter plots in (a)-(c) illustrate the correlations between category mean $d^{*}$, average F1-score within each category, and category size. (d)-(f) depict the same relationships under the category\_downsample scenario, where frequent categories with a large number of debates were downsampled.}
    \label{fig:category_drel}
\end{figure*}

The results from the belief prediction task demonstrate that prediction accuracy increases with the relative dissonance ($d^{*}$) experienced by users. This relationship is also observed at the category level, where debate categories with higher average $d^{*}$ tend to have higher average F1-scores in predicting beliefs (Fig.~\ref{fig:category_drel}). 

To investigate whether higher $d^{*}$ in a category is merely a consequence of the number of debates within that category in the training set (category size), we analyzed the correlations among mean $d^{*}$, average macro F1-score, and category size. 

Our findings reveal that, while all three variables are positively correlated, F1-score shows a much stronger correlation with $d^{*}$ (Pearson correlation $r=0.912$, $p<0.001$) than with category size ($r=0.585$, $p=0.007$). The correlation between $d^{*}$ and category size was $r=0.509$ ($p=0.022$).

This difference in correlation is particularly pronounced in the `category\_downsample' data-splitting scenario (Fig.~\ref{fig:category_drel}d-f). In this scenario, the correlation between $d^{*}$ and category size decreases to $r=0.370$ ($p=0.109$). Here, the correlation between $d^{*}$ and F1-score is $r=0.818$ ($p<0.001$) compared to the correlation between category size and F1-score, $r=0.578$ ($p=0.008$). Figure~\ref{fig:category_drel}e further illustrates that categories with similar sizes still exhibit a wide range of F1-scores. Therefore, while the category size and $d^{*}$ are partially correlated, our results demonstrate that $d^{*}$ exhibits a more robust and consistent correlation with prediction accuracy.

\subsubsection{Effect of effective radius on prediction performance}

\begin{figure*}
    \centering
	\includegraphics[width=0.9\textwidth]{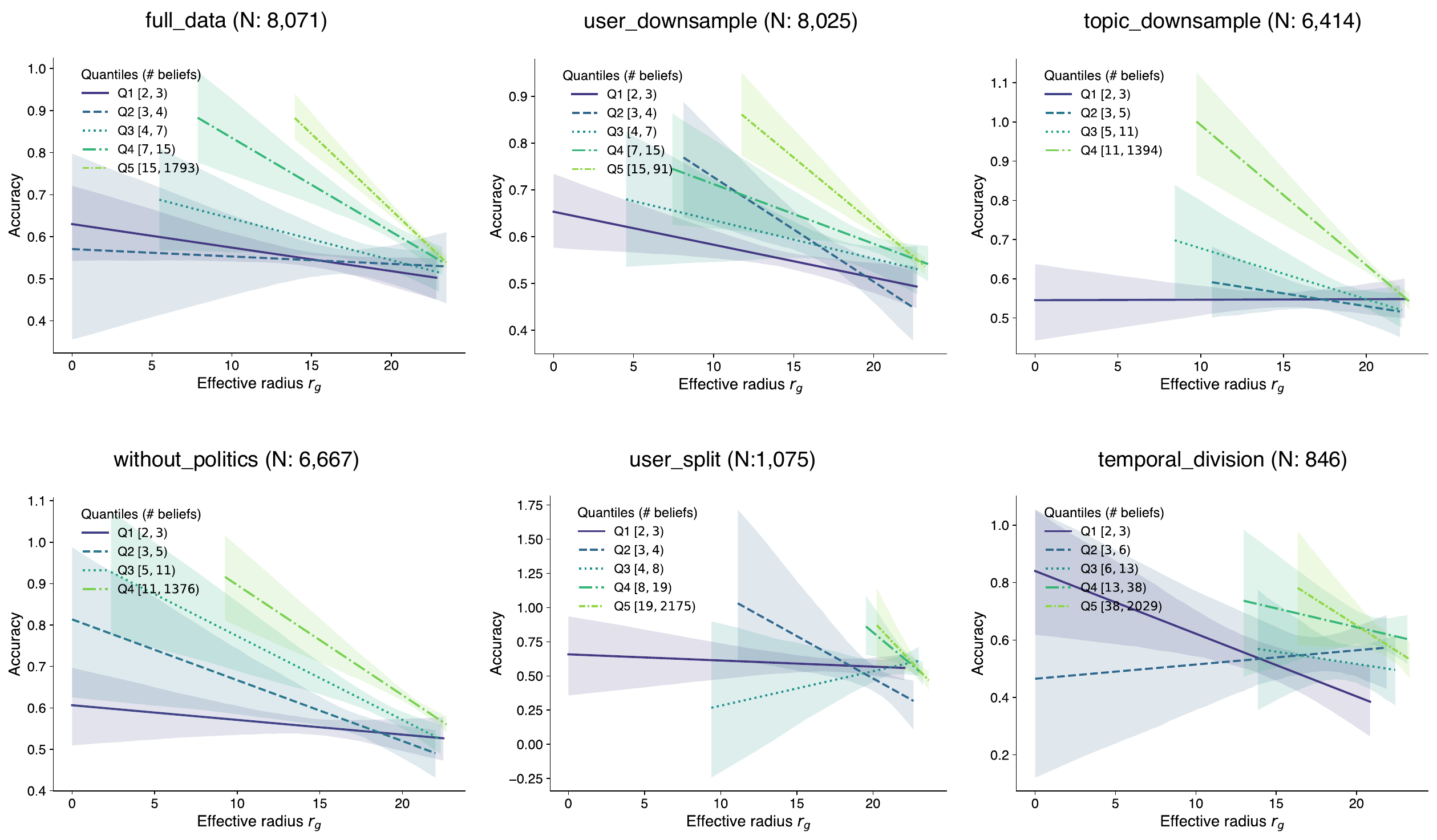}
	\caption{Accuracy as a function of users' effective radius exhibits a consistent decreasing trend across various data-splitting scenarios. These results demonstrate that, given a similar number of prior beliefs, users with a smaller effective radius are more likely to choose beliefs closer to their own compared to users with a larger effective radius. The lines represent user groups divided into quantiles based on the number of prior beliefs, with each group containing a similar number of users. The shaded area indicates the 95\% confidence interval for the regression lines. In the `user\_split' and `temporal\_division' scenarios, the number of test users is smaller compared to other scenarios, resulting in greater fluctuation. }
    \label{fig:robustness_radius}
\end{figure*}

Accuracy as a function of users' effective radius exhibits a consistent decreasing trend across various data-splitting scenarios (Fig.~\ref{fig:robustness_radius}). These results, demonstrate that, given a similar number of prior beliefs, users with a smaller effective radius are more likely to choose beliefs closer to their own compared to users with a larger effective radius. Simple linear regression (Table~\ref{tab:rg_regression}) for average prediction accuracy of individuals based on effective radius $r_g$ and the logarithm of the number of beliefs ($\log n_{\text{beliefs}}$) further validates the negative impact of effective radius on accuracy. The regression model is given by: $Accuracy = \beta_1\log n_{\text{beliefs}} + \beta_2 r_g + C$, where $C$ is a constant. In all data-splitting scenarios, the beta coefficient for effective radius $r_g$ was negative, while the beta coefficient for $\log n_{\text{beliefs}}$ was positive, confirming the negative effect of effective radius on accuracy.

\section{Comparison with the belief co-occurrence network}

In this section, we compare our contrastive learning approach (which leverages a fine-tuned LLM for belief embeddings) with a network-based unsupervised learning method and discuss the advantages of our embedding framework. 

A straightforward alternative to using an LLM-based embedding is to construct a bipartite network of users and beliefs based on voting records from various debates. When a user votes for a particular belief in a debate (either PRO or CON), the vote is represented as an edge connecting the user to the belief in the bipartite network. As a result, the user is exclusively connected to one of the two belief nodes (e.g., PRO – \textit{Debate Title} or CON – \textit{Debate Title}). Projecting this bipartite network onto belief (user) nodes yields a belief (user) network. In the belief network, an edge's weight indicates how many users co-voted for two beliefs, while in the user network, an edge weight represents how many beliefs two users share. The community structure within the belief network highlights groups of densely interconnected beliefs, where each group has more intra-group links than inter-group links. This can be loosely compared to the PCA results in the belief embedding space, in which the first principal component captures the largest variation among belief vectors, separating beliefs into semantically or contextually similar groups.

\begin{figure*}
    \centering
	\includegraphics[width=0.8\textwidth]{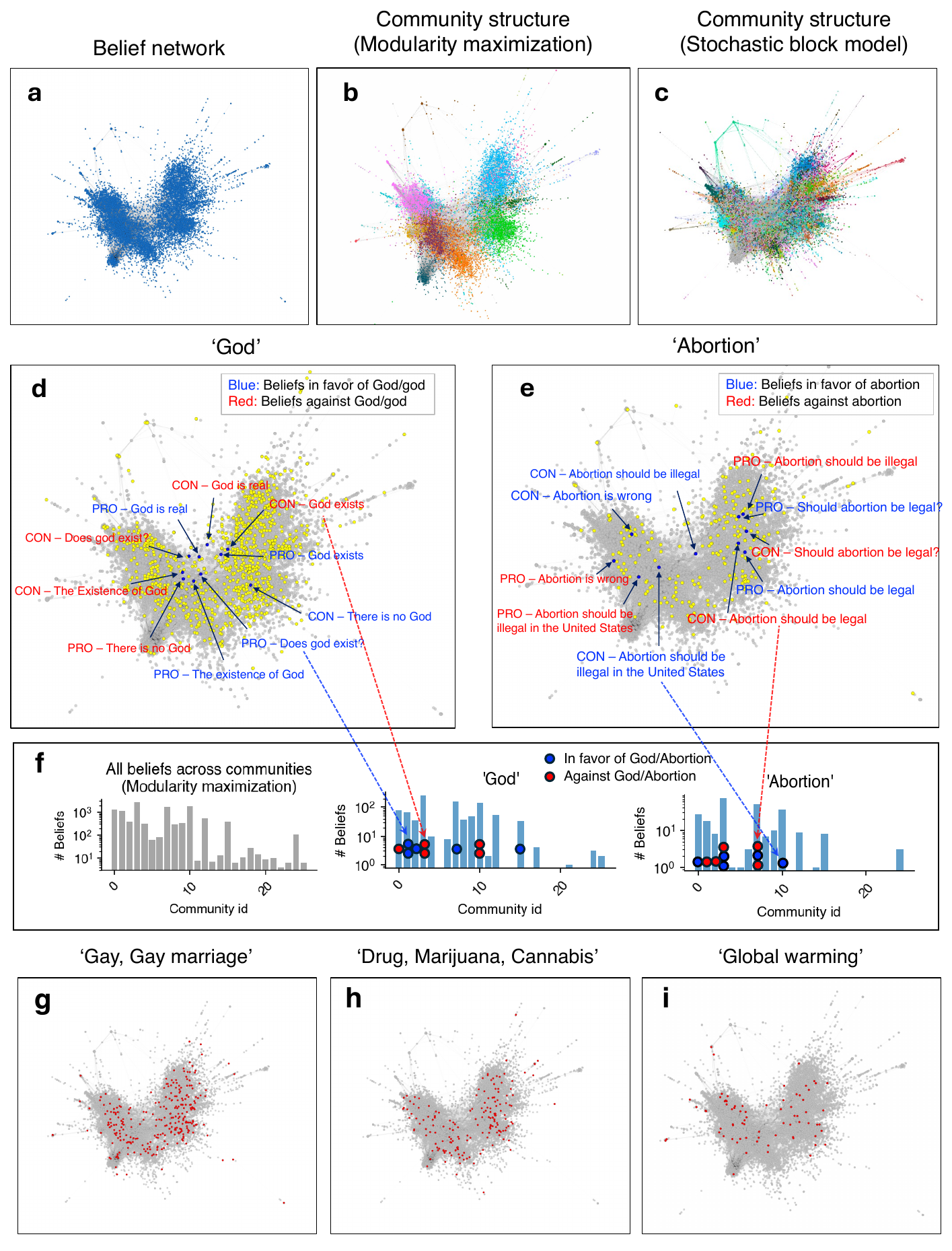}
	\caption{(a) Belief network derived from the user-belief bipartite network. Edge weight represents the number of users who co-voted for two beliefs. Only edges with a weight greater than 1 and nodes in the largest connected component are shown (11,238 nodes, 119,123 edges). The original belief network contains 95,976 nodes and 5,826,924 edges, with many isolated nodes (21,470 connected components). (b) and (c) show the community structure of the belief network using two methods: modularity maximization and stochastic block model-based statistical inference. (d) and (e) highlight beliefs containing the keywords `God' and `Abortion' in yellow. The same sample beliefs as in Fig.~2 of the main manuscript are shown in navy. Label colors indicate stance (blue: beliefs in favor of the topic, red: beliefs against the topic). (f) Distribution of all beliefs, God-related, and Abortion-related beliefs across communities detected via modularity maximization. These beliefs are widely distributed across the network. The community memberships of the sample beliefs in (d) and (e) are marked with red and blue markers. (g)–(i) highlight other keyword-related belief nodes in red: `Gay marriage', `Drug', and `Global warming.'}
    \label{fig:beliefnet}
\end{figure*}

We examined the structural properties of the belief network to understand how beliefs are grouped within the network. Figure~\ref{fig:beliefnet}a presents the belief network obtained from the user-belief bipartite network. The original belief network, constructed based on the beliefs from the same training set used in the main manuscript, contains many isolated nodes (95,976 nodes forming 21,470 connected components), and highly active users contributed numerous low-weight edges (e.g., a single user who votes in 1,000 debates can produce ${}_{1000}C_{2}\simeq 5\times10^5$ edges), resulting in a highly dense network structure. Therefore, we retained only edges with a weight of at least 2 and focused on the largest connected component, resulting in a filtered network with 11,238 nodes and 119,123 edges. Figures~\ref{fig:beliefnet}b and \ref{fig:beliefnet}c show community structures derived from two distinct methods: modularity maximization~\cite{blondel2008fast} and statistical inference based on stochastic block model (SBM)~\cite{zhang2020statistical, peixoto2019bayesian}. Because these algorithms have different underlying optimization mechanisms, their community partitions differ. Specifically, the SBM method yielded more fine-grained communities, whereas the modularity-based approach detected relatively fewer communities with larger number of nodes. We visualized the network using a force-directed layout in \textit{Gephi} software~\cite{gephi2009}.

We then examined whether the bimodal distributions for certain keyword-related beliefs observed in our embedding space would similarly emerge in the belief network. If the belief network effectively located contextually similar beliefs on the network, we would expect these beliefs to be grouped into the same community by the algorithm. However, the results show that the belief network fails to group beliefs related to certain keyword clusters into meaningful or distinct communities, even for controversial topics, such as `God,' `Abortion,' and `Gay marriage.'

Fig.~\ref{fig:beliefnet}d and e show the positions of beliefs containing the keywords `God' and `Abortion' (in yellow), with the same 10 belief samples from Fig.~2 in the main manuscript highlighted as blue nodes with labels. Unlike the belief embeddings, which exhibit clear clustering, these beliefs are broadly scattered across the belief network and communities. Fig.~\ref{fig:beliefnet}f demonstrates that God- and Abortion-related beliefs are dispersed across different communities detected via modularity maximization, with the 10 sample beliefs distributed across various communities.

\begin{figure*}
    \centering
	\includegraphics[width=\textwidth]{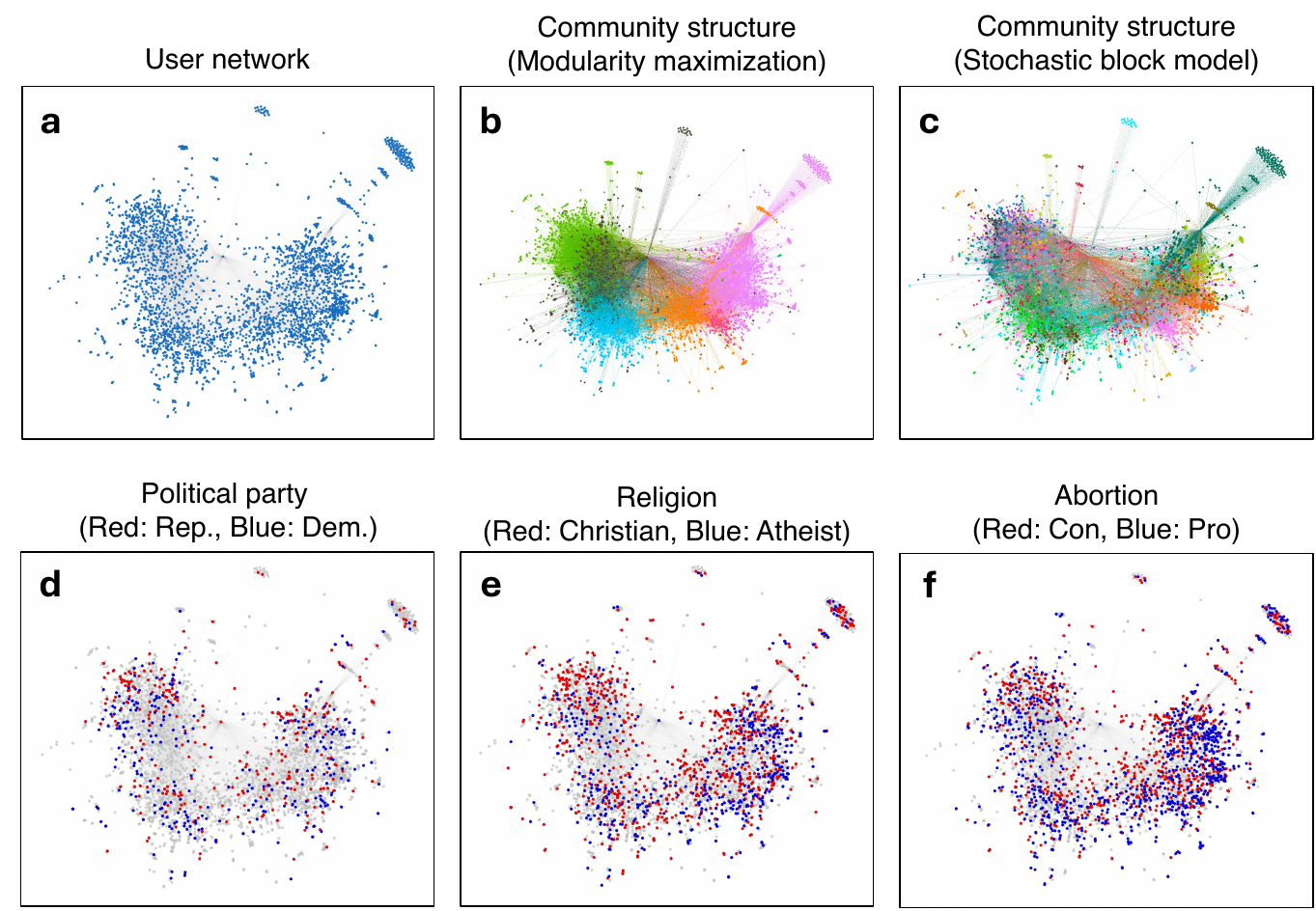}
	\caption{(a) User network obtained by projecting the user-belief bipartite network onto user nodes. Only edges with a weight greater than 1 and nodes in the largest connected component are shown for the visualization (3,109 nodes and 18,407 edges). The original user network consists of 35,628 nodes, 177,916 edges, and has many isolated nodes (21,470 connected components). (b) and (c) illustrate the community structure of the user network, derived using two distinct methods: modularity maximization and statistical inference based on the stochastic block model, respectively. (d)–(f) depict the positions of user groups with diverging stances on their supporting political party (Republican vs. Democratic), religion (Christian vs. Atheist), and abortion (PRO vs. CON), with each group distinctly color-coded.}
    \label{fig:usernetwork}
\end{figure*}

Applying the same filtering procedure to the user network produced a filtered user network with 3,109 nodes and 18,407 edges (Figs.~\ref{fig:usernetwork}a-c). As in the main manuscript's user embedding analysis results (Fig.~3), we color-coded users based on their pre-survey responses. Figs.~\ref{fig:usernetwork}d-e highlight pairs of user groups that diverged by political party (Republican vs. Democratic), religion (Christian vs. Atheist), or stance on abortion (PRO vs. CON). Contrary to the relatively clear separation observed in the embedding space (Fig.~3), the user network exhibited a more mixed pattern, and the community structure did not reflect these user groups.

These findings point to several limitations of a purely network-based approach for representing and capturing belief relationships. First, only beliefs that have been co-voted by at least one user are included in the network; any belief that has not been co-voted remains isolated or is excluded entirely. Second, even semantically identical or highly similar beliefs are treated as distinct nodes if they appear in different debates, leading to redundant information. Furthermore, adding a new belief requires explicit user votes, meaning the network can only expand based on existing user-mediated data. This limitation makes prediction tasks, such as those conducted in the embedding space, infeasible for new debates. By addressing these constraints, the LLM-based embedding framework offers a more flexible and efficient representation of beliefs and user–belief relationships.

\clearpage

\section{Supplementary figures}

\clearpage
\section{Supplementary tables}

\begin{table*}[h!]
\centering
\caption{\label{tab:belief_annotation}  50 sample debate titles with corresponding GPT-4 classification results (1: title can be considered a human belief, 0: otherwise). These titles were used for comparison with human annotators' results.}
\small
\includegraphics[width=0.9\textwidth]{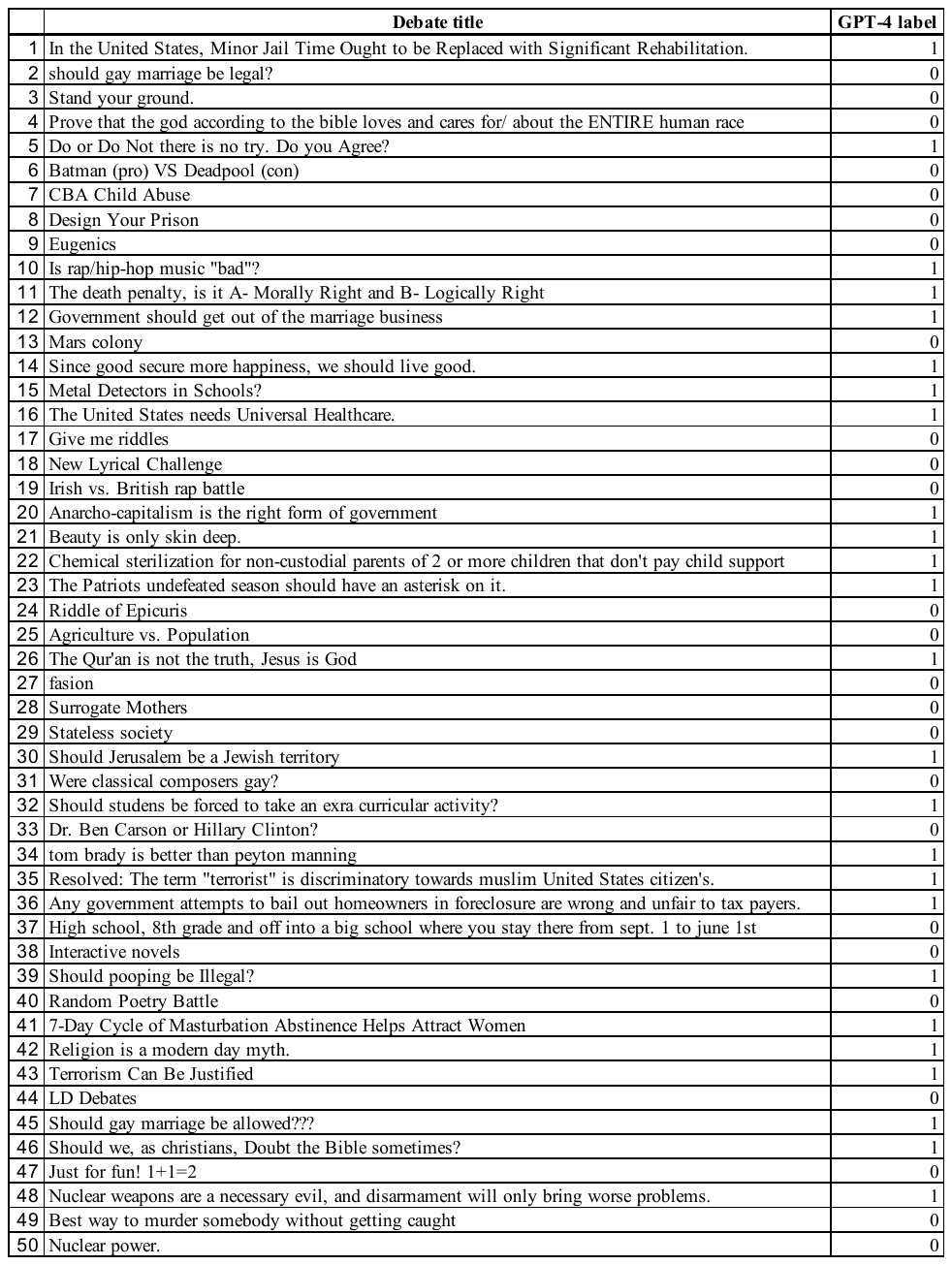}
\end{table*}

\begin{table*}[t!]
\centering
\caption{\label{tab:other_templates} For a given debate title, we used template phrases to express PRO and CON stances. The table shows the original template phrases used for fine-tuning the model, along with three types of alternative phrases for each stance.
}
\small
\includegraphics[width=\textwidth]{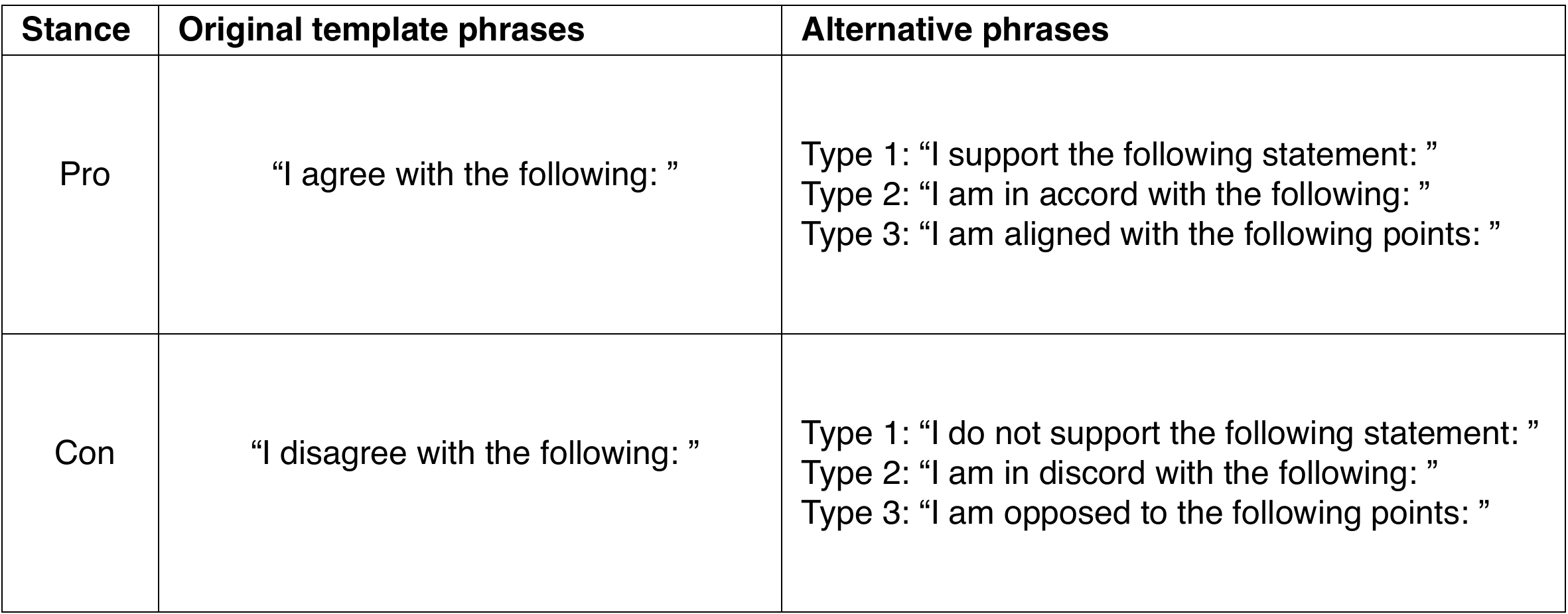}
\end{table*}

\begin{table*}[t!]
\centering
\caption{\label{tab:template_eval} Triplet evaluation tested with three alternative template phrases. For given triplets, the evaluation task examines whether the positive pairs are closer in Euclidean distance than the negative pairs. Higher accuracy indicates that the model clearly distinguishes positive and negative relations of belief pairs. The first row depicts the original triplet evaluation. The other three rows demonstrate the triplet evaluation accuracy measured using the same training and test triplets but with alternative template phrases for PRO and CON stances. We tried three different types of templates, as shown in Table~\ref{tab:other_templates}. The triplet evaluation accuracy remained similar under different choices of templates.}
\small
\includegraphics[width=\textwidth]{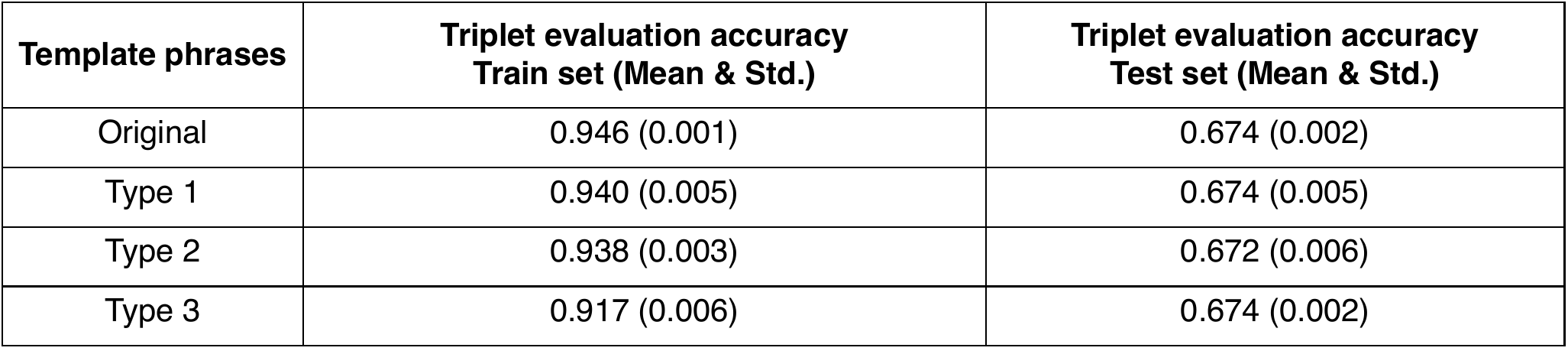}
\end{table*}

\begin{table*}
\centering
\includegraphics[width=0.9\textwidth]{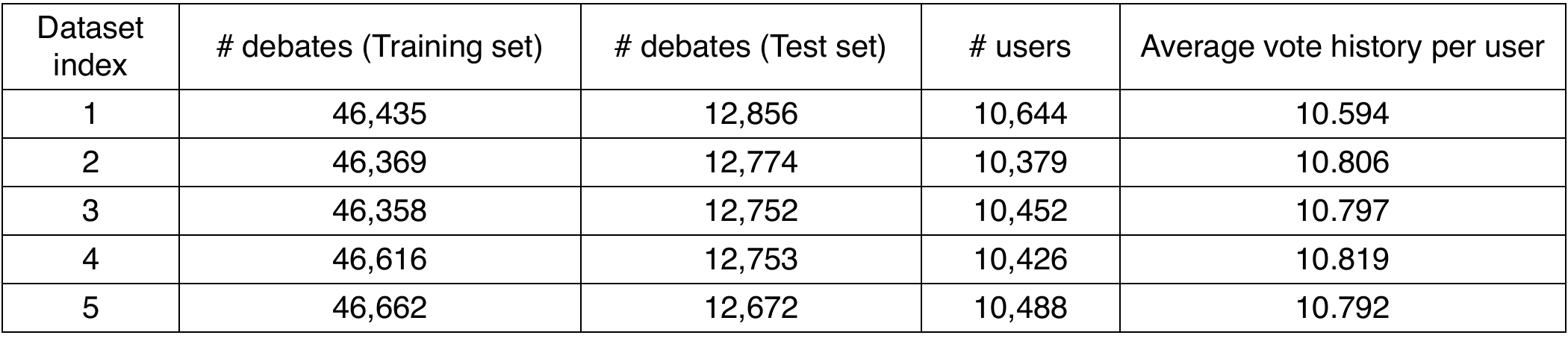}
	\caption{Description of the train and test datasets used for the belief prediction task. A 5-fold evaluation is used for the binary classification of users' beliefs.}
	\label{tab:downstream_statistics}
\end{table*}

\begin{table*}
\centering
\begin{tabular}{lrrr}
\hline
 &  \multicolumn{2}{c}{Triplet Evaluation} & \multicolumn{1}{c}{GLUE-STSB}\\
Data & Train set & Test set & Spearman corr.\\
\hline
debate\_split (original) & 0.946 (0.001) & 0.674 (0.002) & 0.718 (0.005) \\
user\_split & 0.949 (0.001) & 0.728 (0.015) & 0.715 (0.010)  \\
user\_downsample & 0.944 (0.002) & 0.678 (0.002) & 0.705 (0.015) \\
category\_downsample & 0.923 (0.003) & 0.670 (0.003) & 0.735 (0.007) \\
without\_politics & 0.930 (0.001) & 0.683 (0.003) & 0.739 (0.016) \\
temporal\_division & 0.939  & 0.673  & 0.728  \\
debate\_split (BERT) & 0.933 (0.003) & 0.669 (0.004) &  0.476 (0.045) \\
\hline
\end{tabular}
\caption{Triplet evaluation scores for different data splits. Values in parentheses represent the standard deviation from 5-fold validation. The temporal split dataset (`temporal\_division') was tested using a single split, so no standard deviation is reported.  } 
\label{tab:robustness_triplet_evaluation}
\end{table*}

\begin{table*}
\centering
\begin{tabular}{lrr}
\hline
Data & F1-score (Std.) & Accuracy (Std.) \\
\hline
debate\_split (original) & 0.590 (0.005) & 0.590 (0.006) \\
user\_split & 0.566 (0.007) & 0.567 (0.007) \\
user\_downsample & 0.576  (0.003) & 0.577 (0.003) \\
category\_downsample & 0.570 (0.005) & 0.571 (0.006) \\
without\_politics & 0.599 (0.003) & 0.600 (0.003) \\
temporal\_division & 0.574 & 0.574 \\
full\_data\_bert & 0.578 (0.002) & 0.579 (0.002) \\
\hline
\end{tabular}
\caption{Performance metrics of belief prediction task for different data splits measured with macro F1-score and Accuracy. Values in parentheses represent the standard deviation from 5-fold validation. The temporal split dataset (`temporal\_division') was tested using a single split, so no standard deviation is reported.}
\label{tab:robustness_performance_metrics}
\end{table*}

\begin{table*}[ht]
    \centering
    \scriptsize
    \caption{Simple linear regression results for average prediction accuracy of individuals based on effective radius $r_g$ and the logarithm of the number of beliefs ($\log n_{\text{beliefs}}$) across different data-splitting scenarios. The regression model is given by: $Accuracy = \beta_1\log n_{\text{beliefs}} + \beta_2 r_g + C$.
    Beta values are reported with 95\% confidence intervals in brackets. In all scenarios, the beta coefficient for effective radius $r_g$ is negative, while the beta coefficient for $\log n_{\text{beliefs}}$ is positive, demonstrating the negative effect of effective radius on accuracy.}

    \label{tab:rg_regression}
    \begin{tabular}{l c c c c c c}
        \hline
        Dataset & Users & \multicolumn{2}{c}{Effective Radius} & \multicolumn{2}{c}{Log (n\_\text{beliefs})} & Intercept \\
        \cmidrule(lr){3-4} \cmidrule(lr){5-6}
        & & Beta (CI) & p-value & Beta (CI) & p-value & (Const.) \\
        \hline
        full\_data & 8,071 & -0.0088 [-0.0123, -0.0052] & \(1.21 \times 10^{-6}\) & 0.0462 [0.0344, 0.0581] & \(2.47 \times 10^{-14}\) & 0.6448 \\
        user\_downsample & 8,025 & -0.0106 [-0.0140, -0.0071] & \(3.00 \times 10^{-9}\) & 0.0441 [0.0317, 0.0564] & \(2.80 \times 10^{-12}\) & 0.6757 \\
        topic\_downsample & 6,414 & -0.0055 [-0.0099, -0.0012] & \(1.25 \times 10^{-2}\) & 0.0340 [0.0198, 0.0483] & \(3.00 \times 10^{-6}\) & 0.5998 \\
        without\_politics & 6,667 & -0.0104 [-0.0144, -0.0065] & \(2.20 \times 10^{-7}\) & 0.0493 [0.0361, 0.0626] & \(2.69 \times 10^{-13}\) & 0.6721 \\
        user\_split & 1,075 & -0.0088 [-0.0213, 0.0038] & 0.1702 & 0.0144 [-0.0177, 0.0465] & 0.3794 & 0.7147 \\
        temporal\_division & 846 & -0.0088 [-0.0188, 0.0012] & 0.0846 & 0.0340 [0.0069, 0.0611] & \(1.41 \times 10^{-2}\) & 0.6463 \\
        \hline
    \end{tabular}
\end{table*}

\clearpage 
\section*{Reference for Supplementary Information}
References for the Supplementary Information are included in the main reference list of the manuscript.

\end{document}